\documentclass[11pt]{article}

\usepackage[margin=1in]{geometry}
\usepackage{graphicx}
\usepackage{amsmath, amssymb, amsfonts, bbm}
\usepackage{xr-hyper} 
\usepackage{hyperref}
\usepackage{url}
\usepackage{soul}
\usepackage{color}
\usepackage{cancel}
\usepackage{natbib}
\usepackage{float}
\usepackage[dvipsnames]{xcolor}

\usepackage[normalem]{ulem}
\newcommand{\stkout}[1]{\ifmmode\text{\sout{\ensuremath{#1}}}\else\sout{#1}\fi}

\usepackage{multirow}
\usepackage{arydshln}

\usepackage{tikz}
\usepackage{tikz-qtree}
\usetikzlibrary{decorations.pathmorphing}
\usetikzlibrary{trees}
\usetikzlibrary{automata,positioning}

\allowdisplaybreaks

\DeclareMathOperator{\pa}{pa}

\DeclareMathOperator*{\argmin}{\arg\!\min}

\newcommand{\hilbert}{\hbox{${\rm I\kern-.2em H}$}}

\newcommand{\openr}{\mathbb{R}}
\newcommand{\openn}{\hbox{${\rm I\kern-.2em N}$}}
\newcommand{\opend}{\hbox{${\rm I\kern-.2em D}$}}

\newtheorem{theorem}{Theorem}

\newtheorem{definition}[theorem]{Definition}

\newtheorem{lemma}[theorem]{Lemma}

\newtheorem{remark}[theorem]{Remark}

\usepackage{mathtools}
\usepackage{enumerate}
\usepackage{parskip}
\newcommand{\E}{\mathbb{E}} 
\newcommand{\G}{\mathcal{G}}

\newcommand{\I}{\mathbb{I}}

\newcommand{\razi}[1]{\textcolor{olive}}
\newcommand{\anna}[1]{\textcolor{Salmon}}

\title{Fair Risk Minimization under \\ Causal Path-Specific Effect Constraints \\ \vspace{0.5cm}}

\date{}

\usepackage{authblk}

\author[]{Razieh Nabi} 
\author[]{David Benkeser} 
\affil[]{Department of Biostatistics and Bioinformatics\protect\\ Emory University, Atlanta, GA, USA}
\begin{document}

\maketitle

\begin{abstract}
  % This work introduces a novel framework for deriving and estimating fair optimal predictions, focusing on causal and counterfactual path-specific constraints within mean squared error and cross-entropy risk frameworks. We offer closed-form solutions for these constrained optimization problems, detailing the theoretical foundations and practical implications of our approach. Additionally, we propose flexible semiparametric estimation strategies tailored to diverse model specifications, emphasizing the estimation of key nuisance components essential for implementing fairness adjustments in various data-generating contexts. Our methodology not only advances the discourse on fair machine learning but also provides actionable insights for practitioners seeking to balance fairness with predictive performance.

 This paper introduces a framework for estimating fair optimal predictions using machine learning where the notion of fairness can be quantified using path-specific causal effects. We use a recently developed approach based on Lagrange multipliers for infinite-dimensional functional estimation to derive closed-form solutions for constrained optimization based on mean squared error and cross-entropy risk criteria. The theoretical forms of the solutions are analyzed in detail and described as nuanced adjustments to the unconstrained minimizer. This analysis highlights important trade-offs between risk minimization and achieving fairnes. The theoretical solutions are also used as the basis for construction of flexible semiparametric estimation strategies for these nuisance components. We describe the robustness properties of our estimators in terms of achieving the optimal constrained risk, as well as in terms of controlling the value of the constraint. We study via simulation the impact of using robust estimators of pathway-specific effects to validate our theory. This work advances the discourse on algorithmic fairness by integrating complex causal considerations into model training, thus providing strategies for implementing fair models in real-world applications.
    
\end{abstract}

%######################################
\section{Introduction}
\label{sec:intro}
%######################################

% Fairness in algorithmic decision-making
In the realm of machine learning and artificial intelligence, ensuring fairness in algorithmic decision-making has become an imperative challenge. This concern is driven by the growing awareness of the ethical and societal consequences stemming from automated systems. Unlike the traditional focuses that solely enhance statistical and machine learning algorithms across various performance metrics, the emphasis on fairness injects a nuanced layer of complexity into predictive modeling. It mandates a conscientious evaluation of how algorithmic predictions may perpetuate or mitigate existing biases, thereby influencing real-world outcomes across diverse societal sectors.

% Diverse definitions of fairness
The notion of fairness within the machine learning community includes a diverse array of definitions for what it means for an algorithm to be \emph{fair}, with each definition informed by distinct ethical considerations and operational implications \citep{mitchell2018prediction, plecko2022causal, barocas2023fairness}. This diversity often leads to a challenging paradox for practitioners and theorists alike, as adhering to one notion of fairness can sometimes contradict the requirements or objectives of another \citep{kleinberg2016inherent, corbett2018measure, friedler2021possibility}. The task of ensuring algorithmic fairness is made more complex by its inherently  context-specific nature, which demands careful consideration in selecting fairness criteria that are not only theoretically sound but also practically applicable and aligned with prevailing societal values. 

Counterfactual and causal reasoning has emerged as an important framework for quantifying the notion of algorithmic fairness \citep{zhang2017causal, kusner2017counterfactual, zhang2018fairness, chiappa2019path, makhlouf2020survey, nabi2022optimal, nilforoshan2022causal}. Our work centers specifically around building prediction functions that satisfy constraints based on the notion of path-specific causal effects \citep{nabi2018fair}. 
% For example, in building a salary recommendation system for workers involved in a field that requires extensive manual labor, the causal pathway from workers' biological sex to physical strength to work productivity to salary may be deemed a ``fair pathway'' for recommending lower salary. However, any other pathway from sex to salary may be deemed discriminatory and we may wish to build recommendation systems that respect these notions of fairness. 
This approach generally aims to limit the influence of impermissible causal path-specific effects of sensitive attributes on the specified outcomes by extending the concept of direct and indirect effects as notions of (un)fairness that was suggested by \citet{pearl2009causality}. The selection of impermissible pathways must be tailored to each unique context, enhancing the model's relevance by necessitating a thorough examination of the mechanisms linking sensitive attributes with outcomes, as well as the influence of other covariates and mediators. This approach not only makes the model more contextually sensitive but also underscores the importance of understanding the intricate relationships at play.

% Constrained optimization   
By leveraging methods from causal inference, constrained optimization, and semiparametric statistics, we aim to develop an optimal predictive model by nullifying a specified path-specific effect. We frame the task of fair optimal predictions as the outcomes of a penalized risk function, integrating fairness principles seamlessly into the optimization process of the model, a stance that resonates with the insights of previous research \citep{donini2018empirical, zafar2019fairness, chamon2022constrained, nabi2024statistical}. Central to this paper is the theoretical development of \citet{nabi2024statistical}, who characterized the constrained functional parameter as the minimizer of a penalized risk criterion using a Lagrange multiplier formulation. For a given Lagrange multiplier value, the minimizer over the unconstrained parameter space was shown to satisfy a differential equation involving gradients of the risk function and the constraint functional. Across a range of Lagrange multiplier values, this solution defines a path through the unconstrained parameter space, referred to as the constraint-specific path. The Lagrange multiplier value that satisfies the desired constraint then defines the optimal constrained functional parameter. 
% Notably, many machine learning challenges, beyond algorithmic fairness, are suited to constrained optimization, including Neyman-Pearson classification \citep{cannon2002learning}, churn reduction \citep{goh2016satisfying}, robust and adversarial learning \citep{sinha2018certifying, madry2018towards, zhang2019theoretically}, and safe reinforcement learning \citep{garcia2015comprehensive, paternain2019constrained}. Other methods for ensuring fairness include \textit{pre-processing} techniques, which adjust input data to reduce dependencies between sensitive attributes and class labels \citep{kamiran2012data, zemel2013learning, calmon2017optimized}, and \textit{post-processing} approaches, which alter model outputs to achieve fairness \citep{hardt2016equality, woodworth2017learning}. 

% Contributions
This work builds on the theoretical framework developed by \citet{nabi2024statistical} to make several contributions to the field of fair machine learning: \textit{(i) Path-specific effects integration:} We derive the necessary theory to apply the framework of \citet{nabi2024statistical} to  constraints based on path-specific effects. We show closed-form solutions exist for deriving optimal predictions subject to causal path-specific effect constraints under two common risk criteria---mean squared error and cross-entropy. Our closed-form solutions enable us to quantify the reduction in risk and analyze the trade-offs between risk and fairness constraints, thereby providing a clear understanding of the cost of achieving fairness in predictive models.
\textit{(ii) Mechanisms of optimal fair prediction:} We derive insight into the mechanisms of optimal fair prediction by scrutinizing the closed-form solutions. This examination elucidates how fairness constraints alter the optimal unconstrained prediction function and highlights the trade-offs involved in ensuring fairness via counterfactual criteria.  
Further, we reveal that the fairness adjustment process depends not only on the magnitude of the constraint but also significantly on its gradient and the variance of the gradient, thereby detailing how these elements systematically influence the transformation of an unconstrained optimal risk minimizer into a constrained (or fair-optimal) risk minimizer. 
%Specifically, we show how the modifications to the risk minimizer involve the constraint, its gradient, and the variance of the gradient, thereby offering a detailed view of the fairness adjustment process. 
% These solutions illustrate how to transform an unconstrained optimal risk minimizer into a constrained (or optimal-fair) risk minimizer by incorporating modifications that involve the constraint, its gradient, and the variance of the gradient. This process not only lays a solid theoretical foundation for our approach but also clarifies how fairness considerations can be systematically integrated into predictive models. 
\textit{(iii) Robust and flexible estimation:} Given the closed-form solutions of the fair optimal functional minimizer, we address the challenge of estimating the nuisance functionals that are required for the fairness adjustments. \citet{nabi2024statistical} considered only simple plug-in estimators in their examples. Here we study the extent to which estimators may be improved via consideration of more robust estimation frameworks. In sum, our work provides a flexible strategy for learning prediction functions that satisfy constraints defined by pathway-specific causal effects, in addition to enriching our understanding of the implications of fair prediction in this context, both from a statistical, as well as societal point of view.

% Organization 
The paper is organized as follows: Section~\ref{sec:constrained-learning} provides background on statistical learning under causal path-specific effect constraints. Section~\ref{sec:closed-forms} delves into deriving closed-form solutions for fair optimal predictions under both mean squared error and cross-entropy risks. Section~\ref{sec:intuition}, dives into conceptual facets of the closed-form solutions. Section~\ref{sec:est}, introduces a flexible semiparametric estimation framework, which is essential for implementing the proposed fairness adjustments within varied data-generating contexts. Section~\ref{sec:sims} presents simulations to demonstrate the practical application of our theoretical findings. The discussion in Section~\ref{sec:conc} reflects on the broader implications of our work, the challenges encountered, and potential future research directions. All proofs are deferred to the appendix. 

%######################################
\section{Statistical learning under causal fairness constraints}
\label{sec:constrained-learning}
%######################################

Consider the observed datum $O = (S, W, Y)$, with $S$ indicating a sensitive attribute, $W$ denoting other covariates, and $Y$ being the outcome variable. We assume $O$ is distributed according to some distribution $P_0$ within a statistical model $\mathcal{M}$. The model $\mathcal{M}$ could be nonparametric, representing the unrestricted set of all possible probability distributions compatible with $O$, or it could be other infinite-dimensional semiparametric models. The statistical learning problem that we consider can broadly be defined as predicting the value of $Y$ from inputs $(S,W)$. Accordingly, for ease of notation, we sometimes use $Z$ to collectively refer to $(S, W)$. For simplicity, we assume access to a collection of $n$ independent copies of the data unit $O$.

To facilitate our discourse, we adopt the framework of directed acyclic graphs (DAGs) to describe causal relationships among variables. The DAG framework enables the factorization of any distribution $P \in \mathcal{M}$ as $P(O) = \prod_{O_j \in O} P(O_j \mid \pa_\G(O_j))$, where $\pa_\G(O_j) \subset \{S, W, Y\}$ are the parents of variable $O_j \in \{S, W, Y\}$ in DAG $\mathcal{G}$. The causal model implied by a DAG is often described by a set of nonparametric structural equation models with independent error terms (NPSEM-IE) \citep{pearl2009causality}. In this causal model, each variable $O_j \in O$ is generated via a structural equation $f_{O_j}(\pa_\G(O_j), \epsilon_{O_j})$, where $f_{O_j}$ is an unrestricted structural model and the errors $(\epsilon_{O_j}: j)$ are assumed to be mutually independent. Throughout the paper, we use lowercase letters to indicate the values of observed variables.

Let $\psi_0$ be a function mapping from a support of $P_0$ to the real numbers (if $Y$ is continuous) or to the unit interval (if $Y$ is binary). We denote the space of all such functional mappings by $\Psi$. We define $\psi_0$ as the unconstrained minimizer of a relevant risk function $R_{P_0}(\psi_0)$, that is $\psi_0(z) = \argmin_{\psi \in {\Psi}} \  R_{P_0}(\psi)$. Such risks are often formulated as the expectation of a loss function $L(\psi)$, $R_{P_0}(\psi) = \int L(\psi)(o) dP_0(o)$. Many relevant statistical learning problems can be formulated in this way. For example, consider the loss function $L(\psi)(o) = \{y - \psi(z)\}^2$, in which case $\psi_0$ is simply the conditional mean of $Y$ given $Z$.

We consider learning $\psi_0$ while adhering to a pre-defined fairness constraint, which requires that a real-valued functional parameter of $\psi_0$ is set to zero or is otherwise bounded. We formally define the constrained functional parameter in the following. 

\subsection{Constrained functional parameter}
\label{subsec:constrained_func}

For a given $\psi \in \Psi$, let $\Theta_{P_0}(\psi)$ denote a user-selected constraint. The objective is learning a functional parameter, say $\psi^*_0$, where $\Theta_{P_0}(\psi^*_0) = 0$, $\psi^*_0 = \argmin_{\psi \in \Psi, \Theta_{P_0}(\psi) = 0} R_{P_0}(\psi).$ A common approach to solving this constrained optimization is via Lagrangian methods \citep{sundaram1996first,boltyanski1998geometric, chamon2022constrained, nabi2024statistical}. The problem can also be stated in terms of the inequality constraint that $\Theta_{P_0}(\psi^*_0) \le t$ for some threshold $t$. Here, we focus on equality constraints, though our methods easily apply to inequality constraints as well.

\citet{nabi2024statistical} characterized the constrained minimizer above using the \textit{canonical gradient} of the risk, denoted by $D_{R, P_0}(\psi)$, and of the constraint functional, denoted by $D_{\Theta, P_0}(\psi)$. Briefly, these gradients are defined via the notion of pathwise derivations along paths through $\psi$ of the form $\{\psi_{\delta, h}: \delta \in \mathbb{R}\}$, where $\delta$ is a univariate parameter indexing the path, $h$ is the direction of the path defined as $\frac{d}{d\delta}\psi_{\delta, h}|_{\delta = 0}$, with $\psi_{\delta, h}|_{\delta = 0} = \psi$. Let $L^2(P_Z)$ denote the space of all bounded functions $h$ of $Z$ with finite variance under $P_Z$. For any $h(Z) \in L^2(P_Z)$,  we can view the pathwise derivatives of the risk and constraint functionals as mappings $h \rightarrow\frac{d}{d\delta}R_{P_0}(\psi_{\delta, h})|_{\delta = 0}$ and $h \rightarrow \frac{d}{d\delta}\Theta_{P_0}(\psi_{\delta, h})|_{\delta = 0}$, respectively. Assuming these mappings are bounded linear functionals on $ L^2(P_Z)$, by the Reisz representation theorem, these pathwise derivatives will have an inner-product representation using the unique elements $D_{R, P_0}(\psi)$ and $D_{\Theta, P_0}(\psi)$. 

\citet{nabi2024statistical} proposed to construct a \textit{constraint-specific path}, indexed by the Lagrangian multiplier $\lambda \in \openr$, through the unconstrained parameter that would yield a solution to estimation of the constrained functional parameter. Any given point on this path, denoted by $\psi_{0, \lambda}$, is the minimizer to the Lagrangian problem: $\psi_{0, \lambda} = \argmin_{\psi \in \Psi} R_{P_0}(\psi) + \lambda \Theta_{P_0}(\lambda)$. The authors proved that, for any given datum $o$, the constraint-specific path satisfies: 
\begin{align}
    D_{R,P_0}(\psi_{0, \lambda})(o) + \lambda D_{\Theta, P_0}(\psi_{0, \lambda})(o) = 0 \ , \ \ \forall \ \lambda \in \openr \tag{C1} \ .
    \label{eq:pathcondition}
\end{align}%
The constrained parameter $\psi^*_0(z)$ is then equivalent to $\psi_{0, \lambda_0}$, where $\lambda_0$ is the value that satisfies the constraint $\Theta_{P_0}(\psi_{0, \lambda_0}) = 0$. For additional details regarding Condition \eqref{eq:pathcondition}, refer to Appendix~\ref{app:constraint-specific-path}. 

We now consider applying this framework to constraints defined by pathway-specific causal effects.

\subsection{A class of causal and counterfactual constraints} 
\label{subsec:class_constraint}

In this paper, we adopt the causal perspective on fairness as described by \citet{nabi2018fair}, which analyzes fairness through the mechanisms by which a sensitive attribute influences the outcome. Depending on ethical standards and the specific context, certain causal pathways from the sensitive attribute to the outcome are deemed impermissible. The determination of which pathways are considered impermissible hinges on the specific domain and the interpretative significance of the involved variables. 

\begin{figure}[t]
    \centering
    \scalebox{0.85}{
	\begin{tikzpicture}[
		bullet/.style={circle,fill,inner sep=2pt}, 
		long dash/.style={dash pattern=on 10pt off 2pt}, 
		decoration = {snake, pre length=0pt,post length=0pt,}
		]

        % \begin{scope}[xshift=0cm, yshift=1.75cm]
        \begin{scope}[xshift=0cm, yshift=0cm]
    		\path[->, thick]
    		
    		node[] (s) {$S$}
    		node[above of=s, xshift=1.2cm] (x) {$X$}
    		node[right of=s, xshift=1.4cm] (y) {$Y$}
    	
    		% (x) edge[black, -- , dashed] (s) 
            (x) edge[black] (s)
    		(x) edge[black] (y) 
    		(s) edge[black] (y)
    
            % node[below of=s, xshift=1.2cm, yshift=0.1cm] (t1) {(a)} ;
            node[below of=s, xshift=1.2cm, yshift=-0.1cm] (t1) {(a)} ;
    		
	   \end{scope}
        % \begin{scope}[xshift=4.cm, yshift=1.75cm]
        \begin{scope}[xshift=5.cm, yshift=0cm]
    		\path[->, thick]
    		
    		node[] (s) {$S$}
    		node[right of=s, xshift=0.75cm] (m) {$M$}
    		node[above of=m,  xshift=0cm, yshift=0.0cm] (x) {$X$}
    		node[above left of=m, yshift=0.cm] (u) {}
    		node[right of=m, xshift=0.75cm] (y) {$Y$}
    		
    		% (x) edge[black, -, dashed, bend left=0] (s) 
            (x) edge[black] (s)
    		(x) edge[black] (m) 
    		(x) edge[black, bend right=0] (y) 
    		(s) edge[black] (m) 
    		(m) edge[black] (y) 
            (s) edge[black, bend right=20] (y)
            
            % node[below of=m, xshift=0cm, yshift=0.1cm] (t2) {(b)} ;
            node[below of=m, xshift=0cm, yshift=-0.1cm] (t2) {(b)} ;
	   \end{scope}
        % \begin{scope}[xshift=1.cm, yshift=-1.5cm]
        \begin{scope}[xshift=11.cm, yshift=0cm]
    		\path[->, thick]
    		
    		node[] (s) {$S$}
    		node[right of=s, xshift=0.75cm] (m) {$M$}
    		node[above right of=m,  xshift=0cm, yshift=.5cm] (x) {$X$}
    		node[right of=m, xshift=0.75cm] (l) {$L$}
    		node[right of=l, xshift=0.75cm] (y) {$Y$}
    		
    		% (x) edge[black, -, dashed, bend left=0] (s) 
            (x) edge[black] (s)
    		(x) edge[black] (m) 
            (x) edge[black] (l) 
    		(x) edge[black, bend right=0] (y) 
    		(s) edge[black] (m) 
    		(m) edge[black] (l) 
            (l) edge[black] (y) 
            (s) edge[black, bend right=20] (l)
            (m) edge[black, bend right=20] (y)
            (s) edge[black, bend right=25] (y)
            
            node[below of=m, xshift=0.9cm, yshift=-0.2cm] (t3) {(c)} ;
	   \end{scope}
  
	\end{tikzpicture}
    }
    % \vspace{-0.5cm}
    \caption{Depicted are example DAGs demonstrating scenarios where (a) the total effect of sensitive attribute $S$ on outcome $Y$ is considered unacceptable, (b) the direct effect of $S$ on $Y$ is viewed as impermissible, and (c) any set of causal paths from $S$ to $Y$ could be identified as unacceptable. 
    }
    \label{fig:dag_examples}
\end{figure}
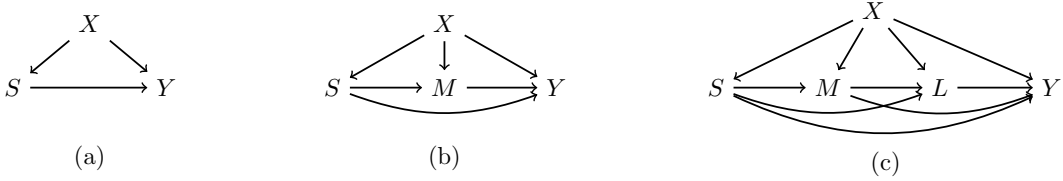

For instance, should \emph{any} influence of a sensitive attribute $S$ on an outcome $Y$ be considered inappropriate, then we may consider setting the constraint equal to the \textit{average treatment effect} (ATE) of $S$ on $Y$, represented as the effect along the $S \rightarrow Y$ pathway in Figure~\ref{fig:dag_examples}(a) with $Z = (S, X)$. For a binary $S$, the ATE is defined as $\E[Y(1) - Y(0)]$, where $Y(s) = f_Y(s,X,\epsilon_Y)$ denotes the potential outcome if $S=s$; $f_Y(.)$ denotes the structural equation for outcome $Y$. In other settings, there may be certain pathways whereby $S$ impacts $Y$ that are considered fair, while any impact on $S$ on $Y$ outside of these pathways would be considered unfair. In this case, we may wish to set our constraint equal to a direct effect, e.g., the $S \rightarrow Y$ pathway in Figure~\ref{fig:dag_examples}(b) with $Z = (S, X, M)$. Such direct effects are often defined via potential outcomes of the form $Y(s, M(s')) = f_Y(s, X, f_M(s', X, \epsilon_M), \epsilon_Y)$. The variable $Y(s, M(s'))$ denotes the outcome $Y$ would assume if $S=s$, while $M$ behaves as if $S=s'$. A \textit{natural direct effect} (NDE) can then be formulated as $\E[Y(1, M(0)) - Y(0)]$.

This idea of certain, but not all, pathways being adjudicated as fair vs. unfair extends to more complex relationships between sensitive and mediating variables. In particular, we can formulate constraints as general \textit{path-specific effects} (PSEs) along unfair pathways. For instance, in Figure~\ref{fig:dag_examples}(c) with $Z = (S, X, M, L)$, we see several pathways from $S$ to $Y$ involving either $M$, $L$ or both. Each pathway could be scrutinized and adjudicated as fair vs. unfair  depending on the context. To formalize this concept within the NPSEM-IE framework, let $\rho$ denote a set of impermissible causal pathways from $S$ to $Y$. We define the potential outcome $Y(s, s'; \rho)$ to reflect scenarios where variables along the pathways specified by $\rho$ behave as if $S = s$, while those along pathways not included in $\rho$ behave as if $S = s'$. 

% This potential outcome is formalized as follows. First, consider $Z_j \in Z \setminus S.$ 
% We define $Z_j(s')$ recursively using the structural equation 
% \begin{align*}
% Z_j(s') = f_{Z_j}\big( s' \ \text{if} \ S \in \pa_\G(Z_j), \{Z_k(s') \ \text{if} \ Z_k \in \pa_\G(Z_j) \setminus \{ S, X\} \}), X, \epsilon_{Z_j}  \big) \ , 
% % \label{eqn:pse_zj_a}
% \end{align*}
% where $X$ denotes all parents of $Z_j$ that are not on a causal pathway from $S$ to $Z_j$. 
This potential outcome $Y(s, s'; \rho)$ is formalized as follows. First, consider $Z_j \in Z \setminus S.$ 
We define $Z_j(s')$ as the counterfactual variable had $S=s'$. It is formalized by adjusting the structural equation of $S$ to $S=s'$. This adjustment causes $S$ to evaluate to $s'$ across the structural equations of all downstream variables. Consequently, the structural equation for $Z_j$ is adapted so that the evaluations of the parents of $Z_j$ reflect $S$ set to $s'$. Furthermore, we define $Z_j(s, s'; \rho)$ by partitioning all mediators between $S$ and $Z_j$ into two disjoint sets: those falling on any edge in $\rho$, denoted $\pa^{\rho}_\G(Z_j)$, and those falling on no edges in $\rho$, denoted $\pa^{\overline{\rho}}_\G(Z_j)$. The structural equation for $Z_j(s, s'; \rho)$ is then recursively determined by:
\begin{align*}
	% \label{eqn:pse_zj_b}
	Z_j(s, s'; \rho) = f_{Z_j}\Big( \big\{ Z_k(s, s'; \rho) \ \text{if} \ Z_k \in \pa^{\rho}_\G(Z_j) \big\}, \big\{ Z_k(s') \ \text{if} \ Z_k \in \pa^{\overline{\rho}}_\G(Z_j) \big\}, X_j, \epsilon_{Z_j} \Big) \ .
\end{align*}%
where $X_j$ denotes all parents of $Z_j$ that are not on a causal pathway from $S$ to $Z_j$. 
If $Z_j = S$, we let $Z_j(s, s'; \rho) = s$ and $Z_j(s')=s'$. Given these definitions, we now define $Y(s, s'; \rho)$ recursively via the following structural equation: 
\begin{align}
	\label{eqn:pse}
	Y(s, s'; \rho) = f_Y\Big( \big\{ Z_j(s, s'; \rho) \ \text{if} \ Z_j \in \pa^{\rho}_\G(Y) \big\}, \big\{ Z_j(s') \ \text{if} \ Z_j \in \pa^{\overline{\rho}}_\G(Y) \big\}, X, \epsilon_Y \Big) \ .
\end{align}
Here, $\pa^{\rho}_\G(Y)$ and $\pa^{\overline{\rho}}_\G(Y)$ again partition the set of all mediators between $S$ and $Y$ into two sets: those along edges in $\rho$ and all others, while $X$ represents the set of parents of $Y$ that are not on any causal pathway from $S$ to $Y$. 

For illustration, consider the DAG in Figure~\ref{fig:dag_examples}(c). We choose  $\rho_1 = \{S \rightarrow L \rightarrow Y\}$ and $\rho_2 = \{S \rightarrow M \rightarrow Y, \ S \rightarrow M \rightarrow L \rightarrow Y\}$ as two different collection of pathways.  The $\rho_1$-specific and $\rho_2$-specific effects can be described by considering $Y(s, s'; \rho_1) = f_Y(s', M(s'), L(s, M(s')), X, \epsilon_Y)$ and $Y(s, s'; \rho_2) = f_Y(s', M(s), L(s', M(s)), X, \epsilon_Y)$, respectively. Here, $L(s, M(s')) = f_{L}(s, M(s'), X, \epsilon_{L})$, $L(s', M(s)) = f_{L}(s', M(s), X, \epsilon_{L})$, $M(s) = f_{M}(s, X, \epsilon_M)$ and $M(s') = f_{M}(s', X, \epsilon_M)$. 

We now formally define the notion of unfairness that is the subject of this paper. 
\begin{definition}[Unfair PSE effect]
     Given a collection of unfair paths $\rho$, the unfair effect of $S$ on $Y$ is defined as the $\rho$-specific effect, denoted by $\Delta^\rho$ and expressed as $\Delta^\rho = \E\left[ Y(1, 0; \rho) - Y(0) \right]$. 
    \label{def:unfair_eff}
\end{definition}

In addition to the common assumptions of positivity and consistency, the identification of PSEs is feasible under specific ignorability assumptions via the \textit{edge g-formula} \citep{shpitser2016causal}. Identification of PSEs in an NPSEM-IE model is grounded in the concept of \textit{edge inconsistency}. A counterfactual outcome $Y(s, s'; \rho)$ is said to be edge inconsistent when any arguments of $f_Y$ depend on both $Z_j(s, \ldots)$ and $Z_j(s', \ldots)$ simultaneously. In such cases, $Z_j$ acts as a ``recanting witness'' that obstructs the identification of the $\rho$-specific effect. For instance in DAG of Figure~\ref{fig:dag_examples}(c), if we examine the effect of $S$ on $Y$ exclusively through the pathway $S \rightarrow M \rightarrow Y$ and disregarding the pathway $S \rightarrow M \rightarrow L \rightarrow Y$, then $M$ acts as a recanting witness, since $M(s)$ and $M(s')$ both appear in the structural equation of $Y$, i.e.,  $f_Y(s', M(s), L(s', M(s')), X, \epsilon_Y)$. This hinders the identification of the effect specific to $\rho$. 

Given an edge-consistent counterfactual $Y(s, s'; \rho)$, we categorize observed variables $O$ into four distinct sets: (i) $X$, encompassing variables not situated on any causal path from $S$ to $Y$; (ii) $S$, the sensitive attribute in question; 
% (iii) $\mathbb{M}_\rho$, composed of variables for which the assignment to $S$ in their counterfactual definitions is 0; and (iv) $\mathbb{L}_\rho$, which holds the remaining variables, i.e., the set of variables for which the assignment to $S$ in their counterfactual definitions is 1. Specifically, if a counterfactual is defined as $Y(1, M(0))$, we have $\mathbb{M}_\rho=\{M\}, \mathbb{L}_\rho = \{Y\}$, and if it is defined as $Y(0, M(1))$, we have $\mathbb{M}_\rho=\{Y\}, \mathbb{L}_\rho = \{M\}$. 
(iii) $\mathbb{M}_\rho$, composed of variables for which the assignment to $S$ in their counterfactual definitions diverges from the assignment to $S$ in the outcome's counterfactual definition; and (iv) $\mathbb{L}_\rho$, which holds the remaining variables, including the outcome variable $Y$ itself. Specifically, $\mathbb{M}_\rho$ includes a variable like $M$ if the counterfactuals are defined as $Y(s, M(s'))$ or $Y(s', M(s))$, indicating that the value assignment to $S$ in $M$'s counterfactual does not match the value assignment to $S$ in $Y$'s counterfactual definition. Conversely, if the counterfactual is specified as $Y(s, M(s))$, implying agreement in the assignment to $S$ between $M$ and $Y$, then $\mathbb{M}_\rho$ is the empty set and $\mathbb{L}_\rho = \{M, Y\}$. 

Following this, we let $s_y \in \{0, 1\}$ represent the assignment of $S$ in the counterfactual outcome, with the alternative assignment denoted by $s'_y$. Specifically, if the direct causal path $S \rightarrow Y$ is included in $\rho$, then $s_y$ is set to $1$; otherwise, it is set to $0$. We assume throughout the main draft that this direct causal path is indeed part of $\rho$, highlighting that any direct influence of the sensitive feature $S$ on the outcome $Y$ is considered unfair and, thus, not permissible. This assumption implies that $\mathbb{L}_\rho$ contains mediators with $S=1$ assignment, and $\mathbb{M}_\rho$ includes variables with $S=0$ assignment. In Appendix~\ref{app:in_ex_direct_eff}, we provide potential variations in our results when the direct effect is excluded from $\rho$.

\begin{lemma}[Identification of causal constraints]\label{lem:id_constraints}
    Given an NPSEM-IE associated with DAG $\mathcal{G}$, the $\rho$-specific effect $\Delta^\rho$, as detailed in Definition~\ref{def:unfair_eff}, is identifiable if and only if there are no recanting witnesses. Assuming $\rho$ includes the direct effect $(s_y=1)$, the identification functional, denoted by $\Theta_{\Delta, P_0}(\psi_0)$, is given by:
    % {\small 
    % \begin{equation}\label{eq:pse_id}
    % \begin{aligned}
    %     \Theta_{\Delta, P_0}(\psi_0) 
    %     &= \I(s_y=1) \bigg\{ \displaystyle \int  \Big\{ \psi(1, w) \prod_{L_i \in \mathbb{L}_\rho \setminus Y} dP_0(\ell_i \mid \pa_\G(\ell_i)) \Big|_{S=1} - \psi(0, w) \prod_{L_i \in \mathbb{L}_\rho \setminus Y} dP_0(\ell_i \mid \pa_\G(\ell_i)) \Big|_{S=0} \Big\} \\ 
    %     &\hspace{3cm} \times \prod_{M_i \in \mathbb{M}_\rho} dP_0(m_i \mid \pa_\G(m_i)) \Big|_{S=0} \  dP_0(x) \bigg\} \\
    %     &\hspace{0.25cm} + \I(s_y = 0) \bigg\{ \displaystyle \int  \Big\{ \prod_{M_i \in \mathbb{M}_\rho} dP_0(m_i \mid \pa_\G(m_i)) \Big|_{S=1} -  \prod_{M_i \in \mathbb{M}_\rho} dP_0(m_i \mid \pa_\G(m_i)) \Big|_{S=0} \Big\} \\ 
    %     &\hspace{3cm} \times \psi(0, w) \prod_{L_i \in \mathbb{L}_\rho} dP_0(\ell_i \mid \pa_\G(\ell_i)) \Big|_{S=0} \  dP_0(x)  \bigg\} \ 
    % \end{aligned}
    % \end{equation}
    % }%
    {\small 
    \begin{align}
        \Theta_{\Delta, P_0}(\psi_0) 
        &= \displaystyle \int \Big\{ \psi_0(1, w) \prod_{L_i \in \mathbb{L}_\rho \setminus Y} dP_0(\ell_i \mid \pa_\G(\ell_i)) \Big|_{S=1} - \psi_0(0, w) \prod_{L_i \in \mathbb{L}_\rho \setminus Y} dP_0(\ell_i \mid \pa_\G(\ell_i)) \Big|_{S=0} \Big\} \notag \\ 
        &\hspace{4cm} \times \prod_{M_i \in \mathbb{M}_\rho} dP_0(m_i \mid \pa_\G(m_i)) \Big|_{S=0} \  dP_0(x) \ . 
        \label{eq:pse_id}
    \end{align}
    }
\end{lemma}
In this lemma, $\psi_0(s, w)$ corresponds to $\psi_0(z)$ where the sensitive attribute $S$ is fixed at the value $s$, and $W$ includes the sets $X$, $\mathbb{M}\rho$, and $\mathbb{L}\rho$, excluding the outcome variable $Y$. 
The notation $f(.)\big|_{S=s}$ is used as a shorthand to denote the evaluation of the function $f$ when $S$ is set to $s$. 

As an example, consider the DAG in Figure~\ref{fig:dag_examples}(c), and assume $\rho = \{S \rightarrow Y, S \rightarrow L \rightarrow Y\}$. Then $Y(1, 0; \rho)$ is defined as $Y(1, M(0), L(1, M(0)))$ with $s_y = 1$, $\mathbb{M}_\rho = \{M\}$, and $\mathbb{L}_\rho = \{L, Y\}$. According to Lemma~\ref{lem:id_constraints}, the $\rho$-specific effect is identified as 
{\small 
\begin{align*}
    \displaystyle \int \left\{ \psi_0(1, m, \ell, x)  dP_0(\ell \mid S=1, m, x) - \psi_0(0, m, \ell, x) dP_0(\ell \mid S=0, m, x)  \right\} dP_0(m \mid S=0, x)  dP_0(x) \ , 
\end{align*}
}%
where $\psi_0(s, m, \ell, x) = \int y dP_0(y \mid s, \ell, m, x)$. 
    
%######################################
\section{Closed-form solutions for fair optimal predictions}
\label{sec:closed-forms}
%######################################

In this section, we provide explicit solutions for the constrained parameter $\psi^*_0(z)$, focusing on two loss functions: (i)  L2 loss, commonly used in regression problems,  $L(\psi)(O) = (Y - \psi(Z))^2$, where its expected value defines the mean squared error (MSE) risk; and (ii) the log loss, suitable for classification tasks, represented as $L(\psi)(O) = - Y \log \psi(Z) - (1-Y) \log (1-\psi(Z))$, with its expectation known as the cross-entropy risk. 

Recalling Section \ref{subsec:constrained_func}, the key objects in the analysis of fair risk minimizers are the gradient of the risk function and the constraint fucntional. These are provided for our two selected risk functions and the $\rho$-specific constraint in the lemma below.

\begin{lemma}[Canonical gradients] 
    The canonical gradients for the MSE and cross-entropy risks are, respectively: 
    \begin{align}
        D_{R, P_0}(\psi)(Z) &= 2\{\psi(Z) - \psi_0(Z)\} \ , \qquad (\textit{\small mean squared error risk}) \label{eq:R_gradient_mse} \\
        D_{R, P_0}(\psi)(Z) &= \displaystyle \frac{\psi(Z) -\psi_0(Z)}{\psi(Z) \ (1- \psi(Z))} \ . \qquad (\textit{\small cross-entropy risk})\label{eq:R_gradient_cross}
    \end{align}

    In addition, the canonical gradient of the constraint, defined in Definition~\ref{def:unfair_eff} and identified via \eqref{eq:pse_id} (while assuming $\rho$ includes the direct effect), is expressed as: 
    \begin{align}
        D_{\Theta_\Delta, P_0}(Z) = \frac{2S - 1}{P_0(S \mid \pa_\G(S))} \times \prod_{M_i \in \mathbb{M}_\rho} \frac{P_0(M_i \mid \pa_\G(M_i) \setminus S, S = 0)}{P_0(M_i \mid \pa_\G(M_i))} \ .
        \label{eq:theta_gradient}
    \end{align}
    \label{lemma:gradients}
\end{lemma}%

Due to the independence of $D_{\Theta_\Delta, P_0}(\psi)$ from $\psi$, we drop $\psi$ from the notation. 

We have the following theorems establishing the optimal fairness-constrained minimizer under these two risk criteria.

\begin{theorem}[Mean Squared Error risk] 
    Assuming $\psi^*_0(z) = \argmin_{\psi \in \Psi, \Theta_{\Delta, P_0}(\psi) = 0} P_0 L(\psi)$, with $L(\psi)$ representing the L2 loss and $\Theta_{\Delta, P_0}(\psi)$ specified by \eqref{eq:pse_id}, the conjunction of condition~\eqref{eq:pathcondition} and $\Theta_{\Delta, P_0}(\psi^*_0) = 0$  necessitates 
    \begin{align}
        \psi_{0}^*(z) = \psi_0(z) -  \Theta_{\Delta, P_0}(\psi_0) \frac{D_{\Theta_\Delta, P_0}(z)}{\sigma^2(D_{\Theta_\Delta, P_0})} \ ,   
        \label{eq:mse_closed-form}
    \end{align}%
    where $D_{\Theta_\Delta, P_0}(z)$ is the constraint gradient, detailed in \eqref{eq:theta_gradient}, and $\sigma^2(D_{\Theta_\Delta, P_0}) = \int D_{\Theta_\Delta, P_0}^2(z) dP_0(z)$.  The mean squared difference between $\psi^*_{0}$ and $\psi_0$ is:
    \begin{align}
       \E\left[ (\psi^*_0(Z) - \psi_0(Z))^2\right] = \frac{\Theta^2_{\Delta, P_0}(\psi_0)}{\sigma^2(D_{\Theta_\Delta, P_0})} \ . 
       \label{eq:mse_fair-unfair}
    \end{align}
\label{thm:closed-mse}
\end{theorem}

The closed-form solution in \eqref{eq:mse_closed-form} along with the mean-squared difference between optimal fair minimizer $\psi^*_0$ and optimal unconstrained minimizer $\psi_0$ in \eqref{eq:mse_fair-unfair} allow us to quantify the reduction in risk and the trade-offs between risk and fairness constraints, offering a precise measure of how fairness adjustments impact the overall risk.

\begin{theorem}[Cross-entropy risk] 
    Assuming $\psi^*_0(z) = \argmin_{\psi \in \Psi, \Theta_{\Delta, P_0}(\psi) = 0} P_0 L(\psi)$, with $L(\psi)$ representing the negative log loss and $\Theta_{\Delta, P_0}(\psi)$ specified by \eqref{eq:pse_id}, condition~\eqref{eq:pathcondition} and $\Theta_{\Delta, P_0}(\psi^*_0) = 0$ together imply
    \begin{align}
        \psi^*_0(z) = \psi_0(z) - \Theta_{\Delta, P_0}(\psi_0) \frac{D_{\Theta_\Delta, P_0}(z) \ \sigma^2_{\psi^*_0}(z) }{ \E[ D^2_{\Theta_\Delta, P_0}(Z) \ \sigma^2_{\psi^*_0}(Z)]} \ , 
        \label{eq:cross_closed-form}
    \end{align}%
    where $\sigma^2_{\psi^*_0}(z) = \psi^*_0(z) (1-\psi^*_0(z))$. The mean squared difference between $\psi^*_{0}$ and $\psi_0$ is bounded by
    \begin{align}
       \E\left[ (\psi^*_0(Z) - \psi_0(Z))^2\right] \leq \frac{\Theta^2_{\Delta, P_0}(\psi_0)}{4 \times \E[D^2_{\Theta_\Delta, P_0}(Z) \sigma^2_{\psi^*_0}(Z)]} \ . 
       \label{eq:cross_fair-unfair}
    \end{align}
\label{thm:closed-cross}
\end{theorem}

While the formulation presented in \eqref{eq:cross_closed-form} is similar to the representation of $\psi_0^*$ under mean squared error, it is not ideally suited for estimation, owing to $\psi_0^*$ appearing on both sides of the equation. For estimation, we may instead rely on a characterization of the constraint-specific path. Under condition~\eqref{eq:pathcondition}, this path is characterized by a quadratic equation: $\lambda D_{\Theta_\Delta, P_0}(z) \psi^2_{0, \lambda}(z) - (1 + \lambda_0 D_{\Theta_\Delta, P_0}(z)) \psi_{0, \lambda}(z) + \psi_0(z) = 0$, for all $\lambda \in \openr$. Thus, we have the following lemma to provide a closed-form solution for any $\psi_{0, \lambda}(z)$ adhering to the constraint-specific path. 
\begin{lemma}
Under conditions of Theorem~\ref{thm:closed-cross}, any $\psi_{0, \lambda}(z)$ on the constraint-specific path has a unique solution in the unit interval as follows:
    \begin{align} 
       \psi_{0, \lambda}(s,w) = \frac{ 1 + \lambda D_{\Theta_\Delta, P_0}(s,w) - (2s-1)  \left[ \left\{ 1 + \lambda D_{\Theta_\Delta, P_0}(s,w) \right\}^2 - 4\lambda \psi_0(s,w)D_{\Theta_\Delta, P_0}(s,w) \right]^{1/2}  }{2\lambda D_{\Theta_\Delta, P_0}(s,w)} \ .  \label{eq:psi0lambda_cross}   
    \end{align}
    \label{lem:closed_quadratic_cross}
\end{lemma}
Given that $\psi^*_0$ equates to $\psi_{0, \lambda_0}$, this explicit formulation of $\psi_{0, \lambda}$ facilitates an iterative approach to estimation of $\psi^*_0$, discussed further in Section~\ref{sec:est}.

\begin{remark}\label{remark:def_odds}
    % In contexts where proportional changes are more informative than absolute differences, it may be relevant to redefine the unfair effect on a multiplicative scale. In Appendix~\ref{app:odds_ratio_scale}, we propose a definition of the $\rho$-specific effect on an odds ratio scale. There we show that the scale of the effect is particularly well suited for optimization of cross-entropy risk.
    In Appendix~\ref{app:odds_ratio_scale}, we study the case that the $\rho$-specific constraint is formulated on a log-odds ratio scale. In this case, the constrained optimizer $\psi_0^*$ under cross-entropy risk has a linear formulation identical to the mean squared error solution in \eqref{eq:mse_closed-form}. This suggests that there may be ``canonical'' formulations of the constrained optimization problem in terms of risk and constraint, such that the optimal constrained solution is a linear function of $\psi_0$. See Appendix~\ref{app:odds_ratio_scale} for further discussion.
\end{remark}

%Having derived solutions for $\psi^*_0$ as functions of various components of $P_0$---namely, the unconstrained parameter $\psi_0$, the constraint $\Theta_{P_0}$, its gradient $D_{\Theta, P_0}$, and the variance of the gradient $\sigma^2(D_{\Theta, P_0})$--we now turn our attention to the estimation of these crucial components. The effectiveness of our methodology in enforcing fairness while optimizing predictions hinges on accurately estimating these components. 

\section{Intuition underlying theoretical results}
\label{sec:intuition}

Equation \eqref{eq:mse_closed-form} implies that the fair risk minimizer $\psi^*_{0}$ can be viewed as an adjustment to the unconstrained risk minimizer $\psi_0$. This adjustment is characterized by three components: $\Theta_{\Delta, P_0}(\psi_0)$, $D_{\Theta_\Delta, P_0}$, and $\sigma^2(D_{\Theta_\Delta, P_0})$. 
Each component is interpretable in its own right:
(i) \textit{Magnitude of systematic disparities}: The parameter $\Theta_{\Delta, P_0}(\psi_0)$ represents the magnitude of systematic disparities linked to $S$ under sampling from $P_0$ -- the larger the underlying disparities, the larger the adjustment that must be made to $\psi_0$; 
(ii) \textit{Adjustment where it matters most}: the gradient of the constraint $D_{\Theta_\Delta, P_0}$ can be viewed as the direction in the model space for $\psi_0$ that leads to the largest change in the constraint. Thus, the constrained minimizer seeks to minimize changes made to $\psi_0$ by making the largest adjustments to $\psi_0$ in regions where these adjustments maximally impact the value of the constraint; (iii) \textit{Ability to impact fairness through adjustment}: the variance of the constraint gradient, $\sigma^2(D_{\Theta_\Delta, P_0})$, represents the extent to which adjustments in $\psi_0$ can impact the constraint. Large variance implies that there are regions of the covariate space, where the gradient of the constraint is much steeper than others, implying that minor changes to $\psi_0$ would result in comparably large changes in the value of the constraint. In this case, we only need make relatively minor adjustments to $\psi_0$ in order to satisfy the fairness constraint. On the other hand, if the variability in the gradient of the constraint is small, then adjustments to $\psi_0$ must be made approximately uniformly across the covariate space, as all values of $Z$ have approximately the same influence on the value of the constraint. 

At the population level, the discrepancy between fairness-adjusted functional parameter $\psi^*_0$ and unconstrained functional parameter $\psi_0$ averages to zero, $\E_{P_0}[\psi^*_0(Z)-\psi_0(Z)] = 0$. This indicates that, although fairness adjustments may alter individual predictions in both positive and negative directions, these adjustments, when aggregated across the population, do not systematically skew the predictions in any particular direction. This outcome arises from the strategy outlined in \eqref{eq:mse_closed-form}, which involves introducing adjustments that are mean-zero, as evidenced by $\mathbb{E}_{P_0}[ D_{\Theta, P_0}(Z) ] = 0$. The rationale for this adjustment strategy rests on the understanding that the MSE of any modified function $\tilde{\psi}(z) = \psi_0(z) + f(z)$, for which $\mathbb{E}_{P_0}[ f(Z) ] \neq 0$, would inherently exceed the MSE of $\psi_0(z) + f(z) - \mathbb{E}_{P_0}[ f(Z)]$. Consequently, while the fairness adjustments themselves do not introduce a net bias across the population due to their mean-zero characteristic, they nevertheless result in a redistribution of the model’s predictive accuracy among individuals. This redistribution affects the variance of the model's predictions and thereby its overall MSE, as quantified by \eqref{eq:mse_fair-unfair}. 

The geometrical interpretation of the individual-level adjustments offered by \eqref{eq:mse_closed-form} enriches our understanding of these adjustments, illustrating a targeted modification strategy that prioritizes areas with the most substantial potential for fairness correction as indicated by the gradient's value. This prioritization strategy inherently focuses on less represented data points, amplifying their role in the fairness adjustment process and underscoring a methodical approach to mitigating unfairness.

\begin{figure}[!t]
    \centering
    \includegraphics[scale=0.2]{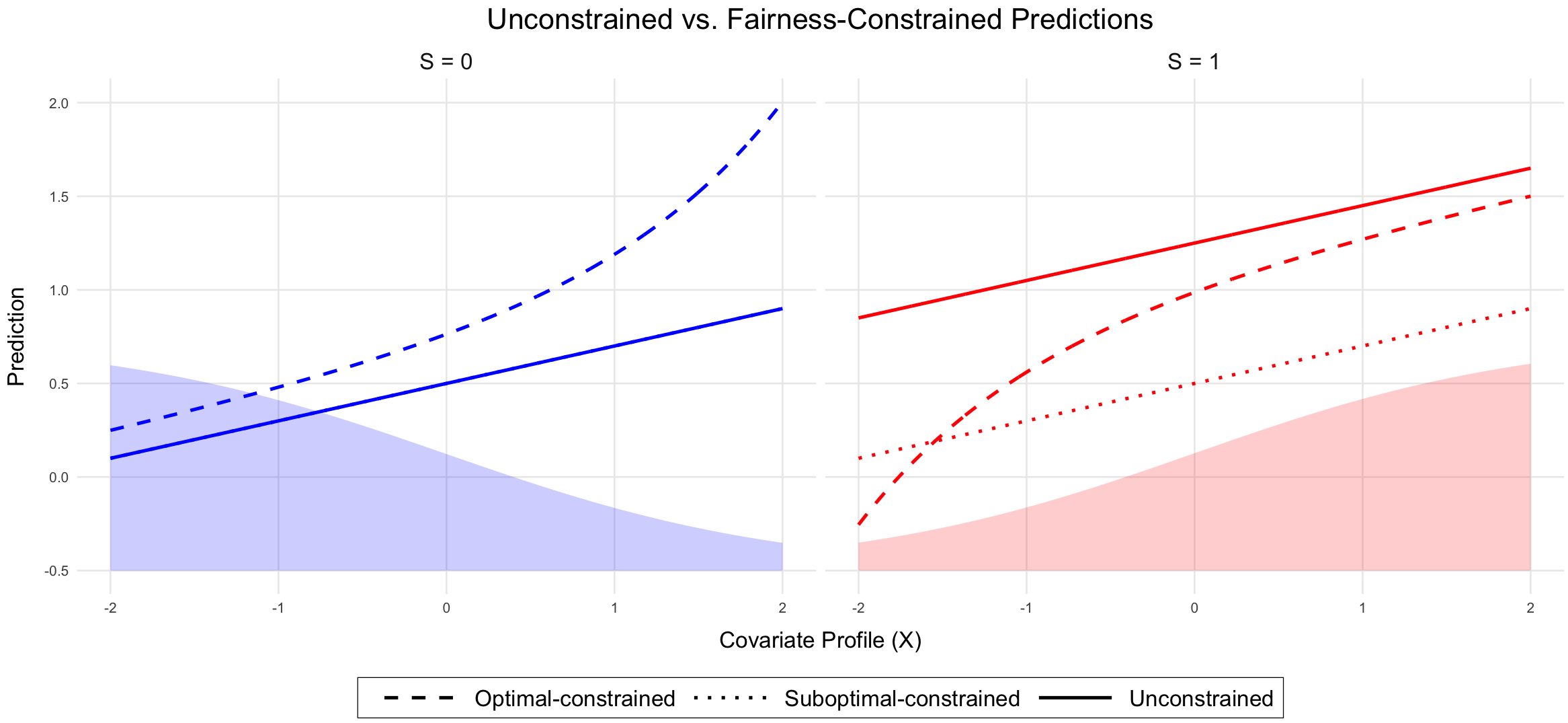}
    \caption{Predictions for the $S = 0$ group (left, blue) and $S = 1$ group (right, red). The optimal prediction function $\psi_0$ (solid line) confers a disadvantage to the $S = 0$ group as evidenced by lower predictions throughout the range of $X$. Using $\psi_0(0,X)$ to predict for both groups (dotted line) solves the constraint, but is suboptimal for prediction owning to large errors made in predictions for the $S = 1$ group. These errors are minimized through prediction using the optimal constrained prediction function $\psi_0^*$ (dashed line). The shaded regions, depicted in lighter tones at the bottom of each plot, represent the distribution of covariate profile $X$ within the sub-population stratified by $S$ values.} 
    \label{fig:fair_adj_ate}
\end{figure}

For instance, consider the DAG model in Figure~\ref{fig:dag_examples}(a) where nullifying the average treatment effect of $S$ on $Y$ is the constraint of interest. Suppose that $X$ is a univariate random variable uniformly distributed between -2 and 2, and $P_0(S=1 \mid X = x) = \text{expit}(x)$. Thus, higher values of $x$ are associated with the $S = 1$ class. Outcomes are generated according to a linear mean function, $\psi_0(s, x) = 0.5 + 0.2x + 0.75s$ (solid lines in Figure~\ref{fig:fair_adj_ate}), implying the ATE of $S = 1$ vs. $S = 0$ is $\Theta_{\Delta, P_0}(\psi_0) = 0.75$. If higher predicted outcomes confer an advantage, then the positive value of $\Theta_{\Delta, P_0}(\psi_0)$ implies that the $S = 0$ group would be on average disadvantaged by predicting from $\psi_0$. A simple solution to nullify the ATE in this example is to use $\psi_0(0, x)$ to predict for an individual with $X = x$, irrespective of their observed value of $S$ (dotted line in Figure~\ref{fig:fair_adj_ate}). While this approach indeed satisfies the constraint, it results in suboptimal predictive performance. Firstly, it introduces a population-level bias in the sense that $\E\{\psi_0(0,X) - \psi_0(S,X)\} \ne 0$, and secondly, individual-level predictions suffer due to the large differences between $\psi(1,X)$ and $\psi(0,X)$ for the relatively large number of observations with $S=1$ and higher $X$ values, evident in Figure~\ref{fig:fair_adj_ate}. According to our findings, optimal predictions must incorporate the constraint $\Theta_{\Delta, P_0}(\psi_0) = 0.75$, the constraint gradient $D_{\Theta_\Delta, P_0}(1, x) = 1/\text{expit}(x)$ and $D_{\Theta_\Delta, P_0}(0, x) = - 1/(1-\text{expit}(x))$, and the variance of constraint gradient $\sigma^2(D_{\Theta_\Delta, P_0}) = \E[1/\{\text{expit}(x) (1-\text{expit}(x))\}]$. The fair optimal predictions are shown as dashed lines in Figure~\ref{fig:fair_adj_ate}. Here, we see that in contrast to the na{\"i}ve approach to solving the constraint, the difference between the optimal constrained predictions and the unconstrained predictions is minimal for regions of the $X$ distribution with the greatest support. The largest differences in predictions arise only for values of $X$ that are seen relatively uncommonly, yet owing to the relative steepness of the constraint gradient are able to maximally change the constraint. The mean squared difference between the suboptimal and optimal solutions demonstrates that the former, $\E[(\psi_0(0, X) - \psi_0(S, X))^2]$, is indeed greater than the latter, $\E[(\psi^*_0(S, X) - \psi_0(S, X))^2]$. 

\begin{figure}[!t]
    \centering
    \includegraphics[scale=0.2325]{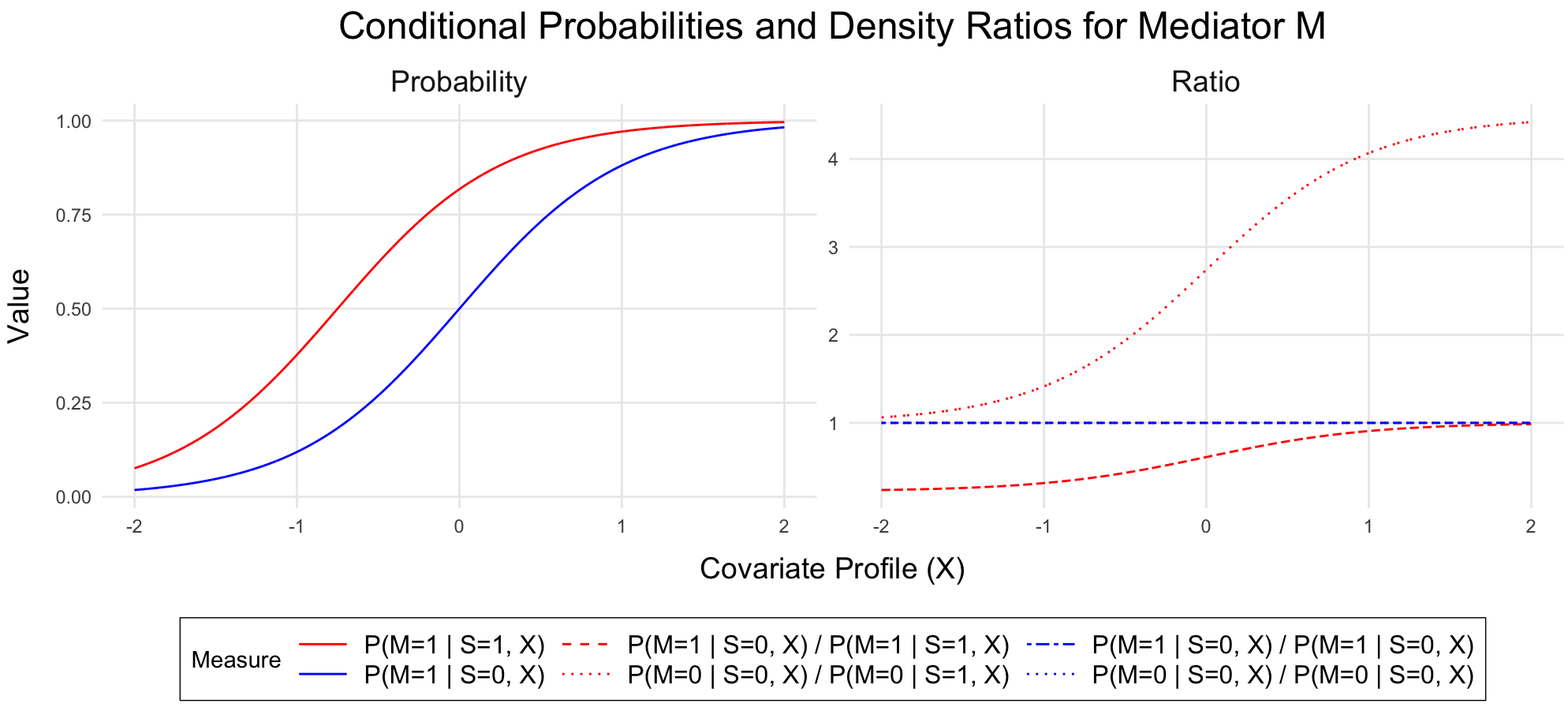} 

    \vspace{0.25cm}
    \includegraphics[scale=0.2325]{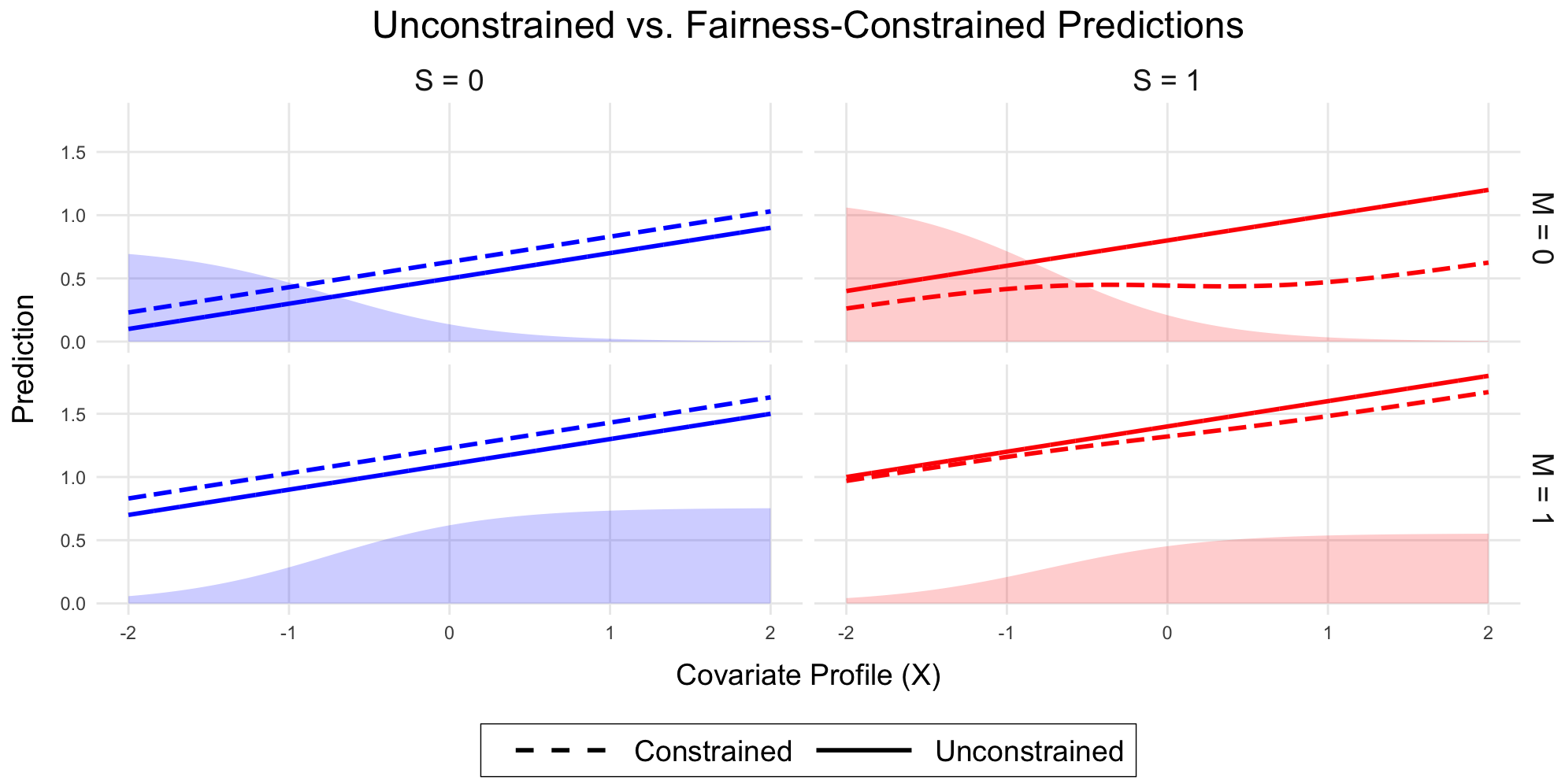}
    \caption{Illustrating the effects of covariate and mediator profiles on fairness adjustments in predictive modeling. 
    \textit{(Top)} This panel displays the probability and density ratios for mediator $M$ across different covariate profiles $X$, stratified by class membership $S$. Higher values of $X$ increase the likelihood of $M=1$ for both $S=0$ and $S=1$ groups. 
    \textit{(Bottom)} This plot contrasts unconstrained (solid lines) with fairness-adjusted (dashed lines) predictions, distinguished by class membership $S$ (color-coded) and mediator values (spatially arranged with predictions for $M=0$ at the top and $M=1$ at the bottom). The shaded regions, depicted in lighter tones at the bottom of each plot, represent the distribution of covariate profile $X$ within the sub-population stratified by values of $S$ and $M$. The visualization reveals that for the $S=1$ group, individuals with $M=0$ and higher $X$ values receive larger adjustments to promote fairness. In contrast, adjustments for the $S=0$ group remain consistent across both $X$ and $M$ values. }    \label{fig:fair_adj_de}
\end{figure}

Consider now the DAG model in Figure~\ref{fig:dag_examples}(b), where the constraint involves nullifying the direct effect of $S$ on $Y$. In this example, we are interested in understanding how values of mediators impact adjustment. Thus, for simplicity we assume that $P_0(S=1 \mid x)=0.5$ for all $x$. The mediator $M$ is a binary variable with $P_0(M = 1 \mid S = s, X = x) = \text{expit}(2x + 1.5s)$, implying a higher likelihood of $M=1$ among the $S = 1$ compared to the $S = 0$ class across all values of $X$ (top left panel of Figure \ref{fig:fair_adj_de}). According to Lemma~\ref{lemma:gradients}, the gradient of the constraint is $D_{\Theta_\Delta, P_0}(s, x, m) = (4s-2) \{P_0(M = m \mid S=0, X = x)/P_0(M = m \mid S=s, X = x)\}$. Thus, the gradient is steepest for values ($s,x,m$) such that the ratio $P_0(M = m \mid S=0, X = x)/P_0(M = m \mid S=s, X = x)$ is large. First, we note that if $s = 0$ then the ratio is constant for all $(x,m)$ (top right panel of Figure \ref{fig:fair_adj_de}, blue line) indicating that modifying predictions made for the $S = 0$ class has a uniform impact on the constraint across values of $(x,m)$. On the other hand, for the $S = 1$ class, modifying predictions with $m = 0$ and large $x$ values will have a relatively larger impact on the value of the direct effect constraint. Therefore, it is these values that are adjusted the most relative to the optimal prediction function (lower panel of Figure \ref{fig:fair_adj_de}). 

Additional data generating processes are similarly considered in Appendix~\ref{app:sims}. There, we investigate scenarios where the propensity score $P_0(S=1 \mid X)$ varies with $X$, affecting adjustments across different distributions of $X$ and the interplay between $X$ and $M$ for the $S=1$ class. The impact of the variance of the constraint gradient $\sigma^2(D_{\Theta_\Delta, P_0})$ on fairness adjustments is also detailed, illustrating how variability in these gradients influences the adjustment mechanisms.

Regarding general $\rho$-specific constraints, the constraint gradient $D_{\Theta_\Delta, P_0}(Z)$ outlined in \eqref{eq:theta_gradient} becomes more complex, encompassing mediator densities within $\mathbb{M}_\rho$ and the propensity of the sensitive characteristic.  Nevertheless, the underlying logic remains consistent: the magnitude of adjustment in the $S=1$ class is governed by the inverse of the propensity score and the product $\prod_{M_j \in \mathbb{M}_\rho} {P_0(M_j \mid S=1, \pa_\G(M_j))}/{P_0(M_j \mid S=0, \pa_\G(M_j))}$, which quantifies shifts in some mediator distributions under the counterfactual scenario of transitioning individuals from the $S=1$ to $S=0$ class. This mechanism seeks to neutralize the benefits that accrue through some downstream effects of $S$, addressing disparities in a targeted manner. 

%######################################
\section{Estimation of fair optimal predictions}
\label{sec:est}
%######################################

In the previous sections, we demonstrated that given the unconstrained parameter $\psi_0$, we can derive a closed-form solution for the constrained parameter $\psi^*_0$. However, in practice, we do not have access to the true distribution $P_0$ and instead rely on estimates of the nuisance parameters to obtain an estimate of $\psi^*_0$. Subsection~\ref{subsec:est_nuisance} discusses flexible semiparametric estimations of the involved nuisance parameters, while Subsection~\ref{subsec:est_conditions} focuses on the sufficient conditions under which the optimal penalized risk can be achieved and the constraint can be controlled.

\subsection{Estimation of nuisance parameters}
\label{subsec:est_nuisance}

In this section, we introduce a flexible semiparametric estimation framework targeting the nuisance parameters crucial for implementing the proposed fairness adjustments. By employing a semiparametric approach, we cater to diverse model specifications. Initially, we concentrate on estimation within the MSE risk framework, deferring the discussion on the cross-entropy risk framework to later in this section. 

There is a substantial body of literature on flexible estimation of the ATE, as well as the NDE and NIE within non/semiparametric models \citep{van2000asymptotic, bang2005doubly, tchetgen2012semiparametric, chernozhukov2018double}. While discussions on flexible estimation of mediation effects involving multiple mediators have emerged \citep{miles2017quantifying, miles2020semiparametric, benkeser2021nonparametric, zhou2022semiparametric}, the development of a general estimation theory for arbitrary PSEs within a semiparametric union model---particularly one that ensures robustness against partial model misspecification---remains an area less explored.

In this paper, we focus our estimation discussions on two explicit scenarios. Consider the DAG in Figure~\ref{fig:dag_examples}(c), where $W=(X, M, L)$ and $Z = (S, W)$. In the first scenario, we consider the direct effect and the effect through $L$ as the unfair mechanisms; that is, $\rho_1 = \{S \rightarrow Y, \ S \rightarrow L \rightarrow Y\}$. In the second scenario, we consider the direct effect and the effects through $M$ as the unfair mechanisms; that is, $\rho_2 = \{S \rightarrow Y, \ S \rightarrow M \rightarrow Y, \ S \rightarrow M \rightarrow L \rightarrow Y\}$. According to Lemma~\ref{lem:id_constraints}, the $\rho_1$-specific and $\rho_2$-specific effects are identified as follows: 
\begin{equation}\label{eq:pse_id_examples}
\begin{aligned}
    \Theta_{\rho_1, P_0}(\psi_0) 
    &= \int \left\{\psi_0(1, w) dP_0(\ell \mid 1, m, x) - \psi_0(0, x) dP_0(\ell \mid 0, m, x) \right\} dP_0(m \mid 0, x) dP_0(x) \ . 
    \\
    \Theta_{\rho_2, P_0}(\psi_0) 
    &= \int \left\{\psi_0(1, w) dP_0(m \mid 1, x) - \psi_0(0, x) dP_0(m \mid 0, x) \right\} dP_0(\ell \mid 0, m, x) dP_0(x) \ . 
\end{aligned}
\end{equation}

We identify the following key nuisance components essential for our analysis: 
(i) the outcome regression $\E_0[Y \mid Z]$, denoted by $\psi_0(Z)$; 
(ii) the propensity score $P_0(S \mid \pa_\G(S))$, denoted by $\pi_0(S \mid \pa_\G(S))$; and 
(iii) the conditional densities of mediators $P_0(M \mid \pa_\G(M))$ and $P_0(L \mid \pa_\G(L))$, denoted by $f_{0, M}(M \mid \pa_\G(M))$ and $f_{0, L}(L \mid \pa_\G(L))$. 
% ; and (iv) the distribution of covariates, $P_{0, X}$. 
Estimates of these nuisance parameters are indicated by subscript $n$, representing the sample size; i.e., $\psi_n, \pi_n, f_{n, M}, f_{n, L}$. We 
% let $P_n f = \frac{1}{n} \sum_{i=1}^n f(O_i)$, and 
use the notation $|| h || = (Ph^2)^{1/2}$ to denote the $L^2(P)$-norm of the function $h$. 

We propose the following estimators of the constraint $\Theta_{\rho_1, P_0}(\psi_0)$ in \eqref{eq:pse_id_examples}. We first define various marginalized estimates: $\bar{\psi}_{n,L,1}(M_i,X_i) = \int \psi_n(S=1,\ell,M_i, X_i) f_{n,L}(\ell \mid S = 1, M_i, X_i) d\ell$, $\theta_{n,1}(X_i) = \int \bar{\psi}_{n,L, 1}(m,X_i) f_{n,M}(m \mid S = 0, X_i) dm$, and $\theta_{n,0}(X_i) = \iint \psi_n(S = 0,\ell,m,X_i) f_{n,L}(\ell \mid S = 0, m, X_i) f_{n,M}(\ell \mid S = 0, X_i) d\ell dm$. Our estimators of $\Theta_{\rho_1, P_0}(\psi_0)$ can then be written as
{\small 
\begin{align*}
    \Theta^\text{plug-in}_{\rho_1, n} &= \frac{1}{n} \sum_{i=1}^n  \left[ \theta_{n, 1}(X_i) - \theta_{n, 0}(X_i) \right]  \ , 
    \\
    \Theta^\text{ipw}_{\rho_1, n} &= \frac{1}{n} \sum_{i=1}^n  \left[ \frac{2S_i -1}{\pi_n(S_i \mid X_i)} \frac{f_{n,M}(M_i \mid S=0, X_i)}{f_{n,M}(M_i \mid S_i, X_i)} Y_i  \right]  \ , 
    \\
    \Theta^\text{ipw-alt}_{\rho_1, n} &= \frac{1}{n} \sum_{i=1}^n  \left[ \frac{\I(S_i=0)}{\pi_n(0 \mid X_i)} \frac{f_{n,L}(L_i \mid S=1, M_i, X_i)}{f_{n,L}(L_i \mid S_i, M_i, X_i)} \psi_n(1, W_i)  - \frac{\I(S_i=0)}{\pi_n(0 \mid X_i)}  \psi_n(0, W_i) \right] \ ,  
    \\ 
    \Theta^\text{aipw}_{\rho_1, n} &= \frac{1}{n} \sum_{i=1}^n  \left[ 
    \frac{\I(S_i = 1)}{\pi_n(1 \mid X_i)} \frac{f_{n,M}(M_i \mid S=0, X_i)}{f_{n,M}(M_i \mid S_i, X_i)} \{Y_i - \bar{\psi}_{n,L,1}(M_i,X_i) \} \right. \\
    &\hspace{6em} \left. + \frac{\I(S_i = 0)}{\pi_n(0 \mid X_i)} \{\bar{\psi}_{n,L,1}(M_i,X_i) - \theta_{n,1}(X_i)\}  + \theta_{n,1}(X_i) \right] \\
    &\hspace{2em} - \frac{1}{n} \sum_{i=1}^n  \left[ \frac{\I(S_i = 0)}{\pi_n(0 \mid X_i)} \{Y_i - \theta_{n,0}(X_i) \} + \theta_{n,0}(X_i) \right] \ .
\end{align*}
}% 

In the following lemma, we establish the conditions under which the proposed estimators above are consistent. 
\begin{lemma}\label{lem:theta_1_est} 
    The estimators are consistent if: 
    \begin{enumerate}
        \item For $\Theta^\text{plug-in}_{\rho_1, n}$: \ $|| \psi_n - \psi_0 || = o_p(1)$ and $|| f_{n, L} - f_{0, L} || = o_p(1)$ and $|| f_{n, M} - f_{0, M} || = o_p(1)$ \ , 
        
        \item  For $\Theta^\text{ipw}_{\rho_1, n}$: \ $|| \pi_n - \pi_0 || = o_p(1)$ and $|| f_{n, M} - f_{0, M} || = o_p(1)$ \ , 
        
        \item $\Theta^\text{ipw-alt}_{\rho_1, n}$: \ $|| \pi_n - \pi_0 || = o_p(1)$ and $|| f_{n, L} - f_{0, L} || = o_p(1)$ and $|| \psi_n - \psi_0 || = o_p(1)$  \ .  
    \end{enumerate}
    Further, $\Theta^\text{aipw}_{\rho_1, n}$ is consistent if at least one of the above conditions hold. 
\end{lemma} 
According to Lemma~\ref{lem:theta_1_est}, $\Theta^\text{aipw}_{\rho_1, n}$ provides a \textit{triply} robust estimator of $\Theta_{\rho_1, P_0}(\psi_0)$. This robustness suggests that the estimator remains consistent if at least two sets of the nuisance estimates--$\{\pi_n\}$, $\{f_{n, M}\}$, $\{\psi_n, f_{n, L}\}$--are consistently estimated. 

For $\Theta_{\rho_2, P_0}(\psi_0)$ in \eqref{eq:pse_id_examples}, we define $\bar{\psi}_{n,L,0}(M, X) = \int \psi_n(S = 1, \ell, M, X) f_{n,L}(\ell \mid S=0, M, X)d\ell$, $\tilde{\theta}_{n, 1}(X) = \iint \bar{\psi}_{n,L,0}(m, X) f_{n,M}(m \mid S=1, X) dm$, and $\tilde{\theta}_{n, 0}(X) = \iint \bar{\psi}_{n,L,0}(m, X) f_{n,M}(m \mid S=0, X) dm$. We then propose the following estimators 
{\small 
\begin{align*}
    \Theta^\text{plug-in}_{\rho_2, n} &= \frac{1}{n} \sum_{i=1}^n  \left[ \tilde{\theta}_{n, 1}(X_i) - \tilde{\theta}_{n, 0}(X_i) \right]  \ , 
    \\
    \Theta^\text{ipw}_{\rho_2, n} &= \frac{1}{n} \sum_{i=1}^n  \left[ \frac{2S_i -1}{\pi_n(S_i \mid X_i)} \frac{f_{n,L}(L_i \mid S=0, M_i, X_i)}{f_{n,L}(L_i \mid S_i, M_i, X_i)} Y_i  \right]  \ , 
    \\
    \Theta^\text{ipw-alt}_{\rho_2, n} &= \frac{1}{n} \sum_{i=1}^n  \left[ \frac{\I(S_i=0)}{\pi_n(0 \mid X_i)} \frac{f_{n,M}(M_i \mid S=1, X_i)}{f_{n,M}(M_i \mid S_i, X_i)} \psi_n(1, W_i) - \frac{\I(S_i=0)}{\pi_n(0 \mid X_i)} \psi_n(0, W_i)   \right] 
    \\
    \Theta^\text{aipw}_{\rho_2, n} &= \frac{1}{n} \sum_{i=1}^n  \left[ \frac{\I(S_i = 1)}{\pi_n(1 \mid X_i)} \frac{f_{n,L}(L_i \mid S=0, M_i, X_i)}{f_{n,L}(L_i \mid S_i, M_i, X_i)} \{ Y_i - \psi_n(1, W_i) \}    \right. \\
    &\hspace{4em} + \frac{\I(S_i = 0)}{\pi_n(0 \mid X_i)} \frac{f_{n,M}(M_i \mid S = 1, X_i)}{f_{n,M}(M_i \mid S_i, X_i)} \{\psi_n(1, W_i) - \bar{\psi}_{n,L,0}(M_i, X_i) \} \\
    &\hspace{6em} \left. + \frac{\I(S_i = 1)}{\pi_n(1 \mid X_i)} 
    \{ \bar{\psi}_{n,L,0}(M_i, X_i) - \tilde{\theta}_{n,1}(X_i) \} + \tilde{\theta}_{n,1}(X_i) \right] \\
    &\hspace{2em} - \frac{1}{n} \sum_{i=1}^n  \left[ \frac{\I(S_i = 0)}{\pi_n(0 \mid X_i)} \{Y_i - \tilde{\theta}_{n,0}(W_i) \} + \tilde{\theta}_{n,0}(W_i) \right] \ .
\end{align*}
}%

We have the following result on the statistical consistency of the above estimators. 
\begin{lemma}\label{lem:theta_2_est} 
    The estimators are consistent if: 
    \begin{enumerate}
        \item For $\Theta^\text{plug-in}_{\rho_2, n}$: \ $|| \psi_n - \psi_0 || = o_p(1)$ and $|| f_{n, L} - f_{0, L} || = o_p(1)$ and $|| f_{n, M} - f_{0, M} || = o_p(1)$ \ , 
        
        \item  For $\Theta^\text{ipw}_{\rho_2, n}$: \ $|| \pi_n - \pi_0 || = o_p(1)$ and $|| f_{n, L} - f_{0, L} || = o_p(1)$ \ , 
        
        \item $\Theta^\text{ipw-alt}_{\rho_2, n}$: \ $|| \pi_n - \pi_0 || = o_p(1)$ and $|| f_{n, M} - f_{0, M} || = o_p(1)$ and $|| \psi_n - \psi_0 || = o_p(1)$  \ .  
    \end{enumerate}
    Further, $\Theta^\text{aipw}_{\rho_2, n}$ is consistent if at least one of the above conditions hold. 
\end{lemma} 

Similar to the previous result, $\Theta^\text{aipw}_{\rho_2, n}$ exhibits a triple robustness behavior; that is, the estimator remains consistent for $\Theta_{\rho_2, P_0}(\psi_0)$ if at least two sets of the nuisance estimates--$\{\pi_n\}$, $\{f_{n, L}\}$, $\{\psi_n, f_{n, M}\}$--are correctly specified. 

According to Lemma~\ref{lemma:gradients}, the gradients for $\Theta_{\rho_1, P_0}(\psi_0)$ and $\Theta_{\rho_2, P_0}(\psi_0)$ are:
{\small 
\begin{align}
    D_{\Theta_{\rho_1}, P_0}(Z) 
    = \frac{2S -1}{\pi_0(S \mid X)} \frac{f_{0,M}(M \mid S=0, X)}{f_{0,M}(M \mid S, X)} 
    \ , \
    D_{\Theta_{\rho_2}, P_0}(Z) 
    = \frac{2S -1}{\pi_0(S \mid X)} \frac{f_{0,L}(L \mid S=0, M, X)}{f_{0,L}(L \mid S, M, X)} \ . 
\end{align}
}
We suggest the following plug-in estimates of these gradients: 
{\small 
\begin{align}
    D_{\Theta_{\rho_1}, n}(Z) 
        &= \frac{2S -1}{\pi_n(S \mid X)} \frac{f_{n,M}(M \mid S=0, X)}{f_{n,M}(M \mid S, X)} \ , \ 
        D_{\Theta_{\rho_2}, P_0}(Z) 
        = \frac{2S -1}{\pi_n(S \mid X)} \frac{f_{n,L}(L \mid S=0, M, X)}{f_{n,L}(L \mid S, M, X)} \ .
\end{align}
}

For $j = 1,2$, the variance of the gradient $D_{\Theta_{\rho_j}, P_0}$ can be estimated using a plug-in estimator
\begin{align}
    \sigma^2_n(D_{\Theta_{\rho_j}, n}) = \frac{1}{n} \sum_{i=1}^n [D_{\Theta_{\rho_j}, n}(Z_i)^2] \ . 
\end{align}
 Thus, our suggested estimator for the fair optimal minimizer for path $\rho_j$, denoted by $\psi_{n,j}^*$ is
 \begin{align}
    \psi_{n,j}^*(z) &= \psi_n(z) -  \Theta_{\rho_j, n}(\psi_n) \ \frac{D_{\Theta_{\rho_j}, n}(z)}{\sigma^2_n(D_{\Theta_{\rho_j}, n})} \ . 
    \label{eq:mse_closed-form_est}
\end{align}%

Under the cross-entropy risk with a constraint defined on a mean difference scale (as per Definition~\ref{def:unfair_eff}), we propose estimating $\psi_{0, \lambda}$ for a specific $\lambda$ using a plug-in, according to \eqref{eq:psi0lambda_cross}, denoted by $\psi_{n, \lambda}$. In this estimation, $\psi_0$ and $D_{\Theta_\Delta, P_0}$ are substituted with their estimates $\psi_n$ and $D_{\Theta_\Delta, n}$, respectively. An estimate for $\lambda$, denoted as $\lambda_n$, is obtained by solving $\lambda_n = \argmin_{\lambda \in \mathbb{R}} |\Theta_{\Delta, n}(\psi_{n, \lambda})|$, where the minimization can be performed using a grid search. Here, $\Theta_{\Delta, n}(\psi_{n, \lambda})$ is an estimate of the constraint, derived from $\Theta^\text{plug-in}_{\Delta, n}$, $\Theta^\text{ipw}_{\Delta, n}$, $\Theta^\text{ipw-alt}_{\Delta, n}$, or the robust combination $\Theta^\text{aipw}_{\Delta, n}$. An estimate of the constrained minimizer is then given by $\psi^*_{n, \lambda_n}$. 

\subsection{Conditions for optimal risk and constraint satisfaction}
\label{subsec:est_conditions}

In this subsection, we formalize the asymptotic conditions under which our estimate of the constrained parameter $\psi^*_n$ (i) achieves the optimal penalized risk such that $R_{P_0}(\psi^*_n) - R_{P_0}(\psi^*_0) = o_{P_0}(1)$, and (ii) satisfies the constraint such that $\Theta_{P_0}(\psi^*_n) = o_{P_0}(1)$. Here, we concentrate on asymptotic conditions
within the MSE risk framework. Similar statements can be made within the cross-entropy risk framework while the unfair effect is measured on a log-odds ratio scale. 

We first fix some notations. For a $P_0$-measurable function $f$, let $P_0f = \int f(o) \ d P_0(o)$, let $|| f ||_1 = P_0(|f|)$ denote the $L^1(P_0)$-norm, and let $|| f ||_2 = (P_0f^2)^{1/2}$ denote the $L^2(P_0)$-norm of the function $f$. Let $P_n f = \frac{1}{n}\sum_{i = 1}^n f(O_i)$ denote the empirical evaluation of the function $f$. Let $\theta_n$ be the selected estimate of $\theta_0 = \Theta_{\Delta, P_0}(\psi_0)$, the constraint. Let $D_n$ denote an estimate of $D_0 = D_{\Theta_{\Delta}, P_0}$, the gradient of the constraint. Let $\sigma_n^2$ be an estimate of $\sigma_0^2 = P_0 D_{0}^2$, the variance of the gradient. Let $\mathcal{W}$ denote the domain of $W$. 

We assume the following regularity conditions: 
\begin{enumerate}
    \item[] (R1) \ There exists $C \in \mathbb{R}^{> 0}$ such that $\sup_{s \in \{0, 1\}, \ w \in \mathcal{W}} D_0(s, w) < C$, and 

    \item[] (R2) \ There exists $\delta \in \mathbb{R}^{> 0}$ such that for any $\epsilon > 0$, $P_0( \sigma^2_n > \delta ) > 1 - \epsilon$. 
\end{enumerate}

 We have the following result. 
\begin{theorem}
    Under regularity conditions (R1) and (R2), the estimate $\psi^*_n$ achieves the optimal penalized risk, i.e., $R_{P_0}(\psi^*_n) - R_{P_0}(\psi^*_0) = o_{P_0}(1)$, if the following conditions are met: 
    \begin{enumerate}
        \item[C1.] $L^2(P_0)$-consistency for the estimator $\psi_n$: $|| \psi_n - \psi_0 ||_2 = o_{P_0}(1)$, 
        \item[C2.] $L^2(P_0)$-consistency for the estimator $D_n$: $|| D_n - D_0 ||_2 = o_{P_0}(1)$, 
        \item[C3.] Consistency of the estimate $\theta_n$: $\theta_n - \theta_0 = o_{P_0}(1)$, and 
        \item[C4.] Consistency of the estimate $\sigma^2_n$: $\sigma^2_n - \sigma^2_0 = o_{P_0}(1)$. 
    \end{enumerate}
    \label{thm:risk_at_P0}
\end{theorem}
See a proof in Appendix~\ref{app:proof_risk_at_P0}. 

Theorem~\ref{thm:risk_at_P0} establishes a set of sufficient conditions under which the estimate of the constrained minimizer, $\psi^*_n$, achieves the lowest possible risk amongst prediction functions that satisfy the constraint. This optimal risk is attainable under the correct specification of the distribution $P_0$. However, a more relaxed scenario arises when we use an estimator $\theta_n$ that incorporates the same nuisance parameters as the ones used to estimate the gradient $D_0$, i.e., the propensity score $\pi_n$ and mediator densities in $\mathbb{M}_\rho$. Examples include the IPW estimator or the multiply robust AIPW estimator described earlier. This way, we can tolerate misspecifications of the mediator densities in $\mathbb{L}_\rho$. 

We now formalize the asymptotic conditions required to satisfy the constraint. Assume that $\psi_n$ converges in probability to $\tilde{\psi} \in \Psi$, where $\tilde{\psi}$ may differ from our unconstrained minimizer $\psi_0$. We then obtain the following result.   
\begin{theorem}
    Under regularity conditions (R1) and (R2), the estimate $\psi^*_n$ satisfies the constraint, i.e., $\Theta_{P_0}(\psi^*_n) = o_{P_0}(1)$, if the following conditions are met: 
    \begin{enumerate}
        % \item[C1.] $L^1(P_0)$-consistency for the estimator $\psi_n$: $|| \psi_n - \psi_0 ||_1 = o_{P_0}(1)$, 
        \item[C1.] $L^1(P_0)$-consistency for the estimator $D_n$: $|| D_n - D_0 ||_1 = o_{P_0}(1)$, 
        \item[C2.] Consistency of the estimate $\theta_n$: $\theta_n - \Theta_{\Delta, P_0}(\tilde{\psi}) = o_{P_0}(1)$, and 
        \item[C3.] Consistency of the estimate $\sigma^2_n$: $\sigma^2_n - \sigma^2_0 = o_{P_0}(1)$. 
    \end{enumerate}
    \label{thm:constraint_at_P0}
\end{theorem}

See a proof in Appendix~\ref{app:proof_constraint_at_P0}. 

Theorem~\ref{thm:constraint_at_P0} establishes sufficient conditions for satisfying the constraint at $\psi^*_n$. These conditions include consistently estimating the gradient $D_0$ and using an estimate of the constraint that is consistent for $\Theta_{P_0}(\tilde{\psi})$ for any choice of $\tilde{\psi}$.

Our theorems focus specifically on showing general conditions under which our estimator $\psi^*_n$ achieve optimal risk and proper control of the constraint in large samples. We have chosen to present these results  agnostic to how the various nuisance components are estimated. 
 However, it is possible to refine these theorems further when considering specific estimators of nuisance quantities. For instance, if specific convergence rates are available for nuisance estimators, then it should generally be possible to restate our theorems in terms of $O_{P_0}$ convergence. Additionally, if a plug-in estimator of the gradient $D_0$ is used wherein one substitutes estimates of unknown nuisance parameters in $D_0$ with their estimated counterparts, then it should generally be possible to derive sufficient conditions on these nuisance estimators that would imply appropriate convergence of $D_n$ to $D_0$. However, because there are often multiple approaches to estimation of nuisance parameters appearing in these gradients (e.g., estimating mediator density ratios rather than mediator densities directly), we have opted to leave our results in terms of convergence of $D_n$ to $D_0$. However, we wish here to maintain generality in terms of the nuisance estimators used in our proposed procedure.

%######################################
\section{Simulations}
\label{sec:sims}
%######################################

We evaluated our estimators in several settings. For each constraint and for a number of bounds on the constraint, we studied (i) the risk of our estimator $R_{P_0}(\psi^*_{n})$ as compared to the optimal constrained risk $R_{P_0}(\psi^*_{0})$ and (ii) the true fairness constraint under our estimator. For each setting, we generated 1000 simulated data sets of sizes $n$ = 100, 200, 400, 800, 1600. We also generated an independent test set of 1$e$6 observations to numerically approximate the value of the true risk and the true constraint. For each simulated data set, we generated our proposed estimators based on six different constraint bounds and the four proposed estimators of the constraint. We focus on results for $\Theta_{\rho_1,P_0}$; results for $\Theta_{\rho_2, P_0}$ were similar and are included in Appendix~\ref{app:sims}.

\subsection{Impact of inconsistent nuisance estimation}

Based on our theoretical results, we expect that certain patterns of misspecification will lead to sub-optimal risk such that we expect that $R_{P_0}(\psi_n^*) > R_{P_0}(\psi_0^*)$ for all $n$. On the other hand, our results suggest that there is some robustness to misspecification in terms of controlling the constraint. That is, for some formulations of our estimator we expect that $\Theta_{P_0}(\psi_n^*) \approx 0$ even in these settings with misspecification.

The goal of this simulation is to provide empirical support to our theoretical claims presented in Theorems \ref{thm:risk_at_P0} and \ref{thm:constraint_at_P0} pertaining to the behavior of the risk and constraint under inconsistent nuisance estimation. Therefore, we utilized simple generalized linear models and maximum likelihood, an approach that has well understood behavior under model misspecification. First, we correctly specified all nuisance models to confirm optimal performance in terms of risk and constraint. We then studied the performance when various combinations of nuisance parameters were inconsistently estimated. 

In this simulation, the covariate vector $X$ was two-dimensional with $X_1 \sim \mbox{Bern}(1/2)$ and $X_2 \sim \mbox{Unif}(-2, 2)$. Given $X$, $S$ was generated from a $\mbox{Bern}(\mbox{expit}(-X_1 + 2 X_1 X_2))$. Given $X$ and $S$, $M$ was generated from a $\mbox{Bern}(\mbox{expit}(-1/2 + S/2 - X_1 + 2X_1 X_2))$. Given $(X, S, M)$, $L$ was generated from a $\mbox{Bern}(\mbox{expit}(-1 + S/2 - M/2 - X_1 + 2X_1X_2))$. Given $(X, S, M, L)$, $Y$ was generated from a $\mbox{Normal}(-1/2 + S + M/2 - L/2 - X_1 + 2 X_1 X_2, 4)$ distribution. Under this data generating mechanism, $\Theta_{\rho_1, P_0}(\psi_0) \approx 0.96$.

We note the presence of the cross-product between $X_1$ and $X_2$ in the various components of the data generating process. In the scenario where all nuisance parameters are consistently estimated, these cross-product terms are included in the regression model specification. In the case that the nuisance parameters are inconsistently estimated, these cross-product terms are omitted, leading to biased estimation of the nuisance parameter.

\begin{figure}
\centering
\includegraphics[width=4cm, height=3.6cm]{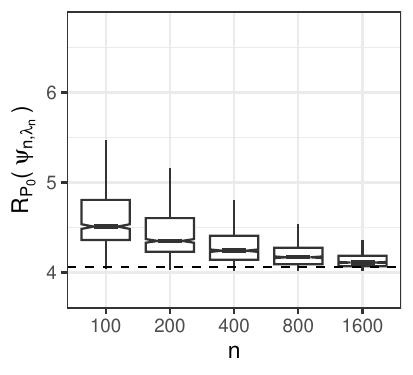}
\includegraphics[width=4cm, height=3.6cm]{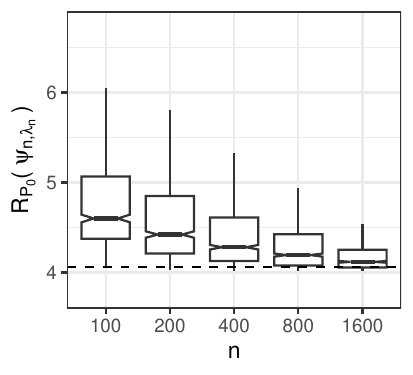}
\includegraphics[width=4cm, height=3.6cm]{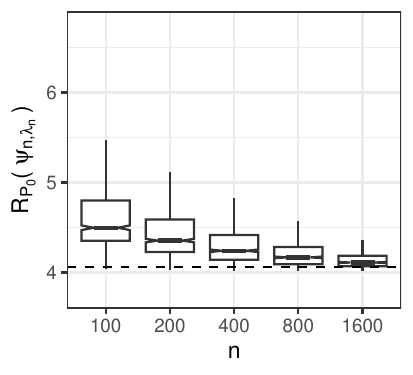}
\includegraphics[width=4cm, height=3.6cm]{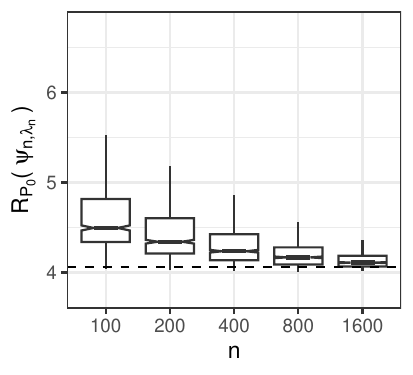}
\includegraphics[width=4cm, height=3.6cm]{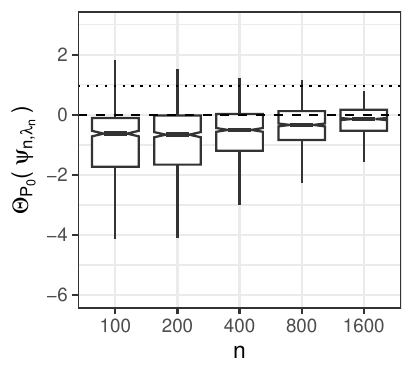}
\includegraphics[width=4cm, height=3.6cm]{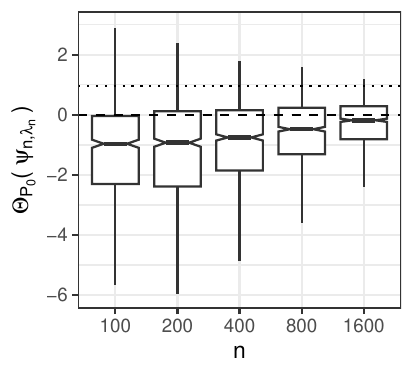}
\includegraphics[width=4cm, height=3.6cm]{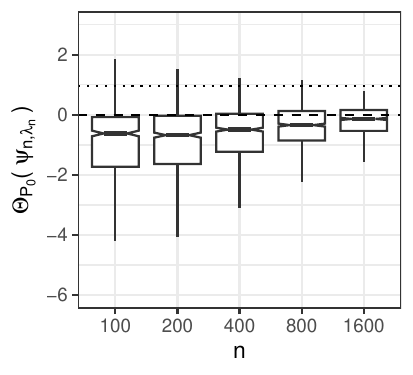}
\includegraphics[width=4cm, height=3.6cm]{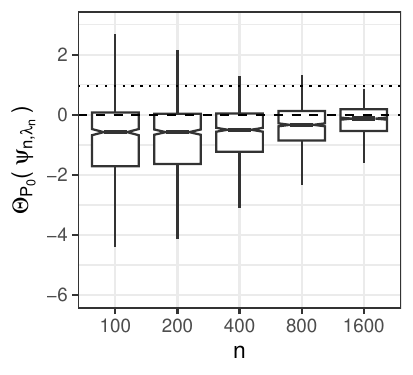}
\caption{\textbf{Estimates of optimal predictions under $\rho_1$-pathway constraint for mean squared error risk.} \underline{Top row:} Various estimators of the constraint are shown; from left to right: $\Theta^\text{plug-in}_{\rho_1, n}$, $\Theta^\text{ipw}_{\rho_1, n}$, $\Theta^\text{ipw-alt}_{\rho_1, n}$, $\Theta^\text{aipw}_{\rho_1, n}$. For each estimator, we show the distribution of risk of $\psi_{n,\lambda_n}$ over 1000 realizations for each sample size for the equality constraint $\Theta_{P_0}(\psi) = 0$. The dashed line indicates the optimal risk $R_{P_0}(\psi_0^*)$. \underline{Bottom row:} Distribution of the true constraint over 1000 realizations for each sample size. The dashed line indicates the equality constraint value of zero. The dotted line indicates the true value of the constraint under $\psi_0$.
}
\label{fig:pse1_sim_all_correct}
\end{figure}

\begin{figure}
\centering
\includegraphics[width=4cm, height=3.6cm]{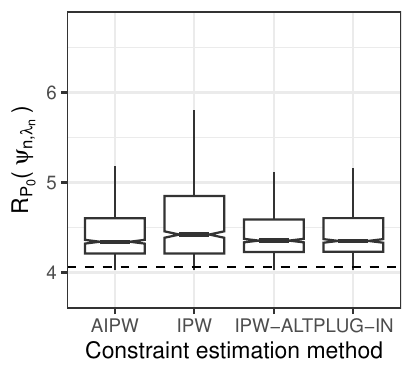}
\includegraphics[width=4cm, height=3.6cm]{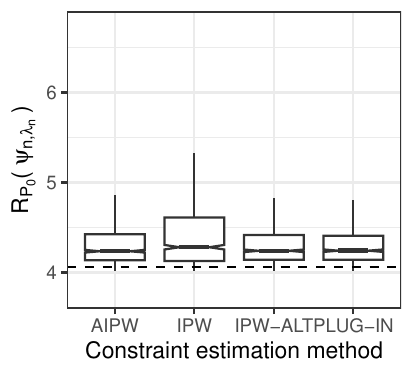}
\includegraphics[width=4cm, height=3.6cm]{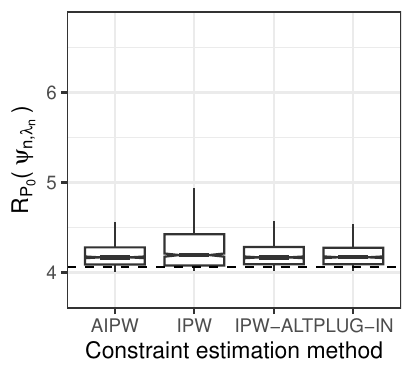}
\includegraphics[width=4cm, height=3.6cm]{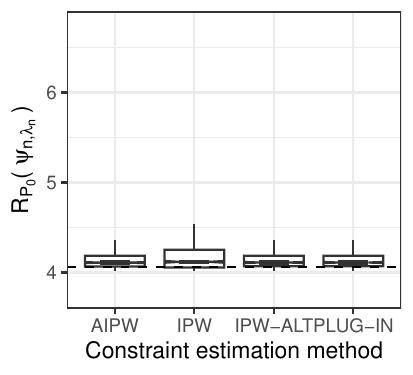}
\includegraphics[width=4cm, height=3.6cm]{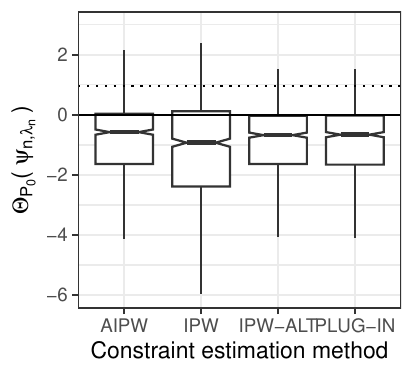}
\includegraphics[width=4cm, height=3.6cm]{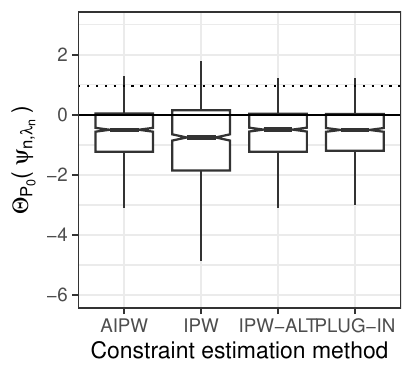}
\includegraphics[width=4cm, height=3.6cm]{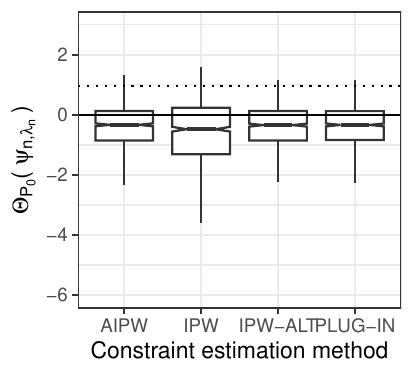}
\includegraphics[width=4cm, height=3.6cm]{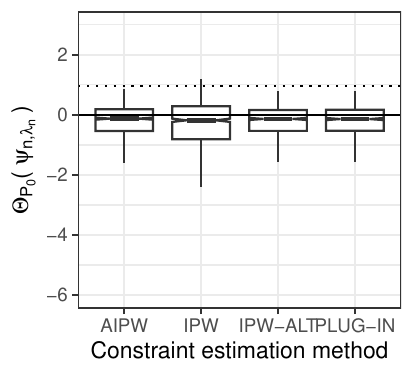}
\caption{\textbf{Comparison of pathway specific effect estimates used in construction of $\psi_n^*$.} \underline{Top row:} The various estimators are shown in each figure for sample sizes (from left-to-right) of $n=200,400,800, 1600$. For each estimator, we show the distribution of mean squared error of $\psi_{n,\lambda_n}$ over 1000 realizations for each sample size for the equality constraint $\Theta_{P_0}(\psi) = 0$. The dashed line indicates the optimal risk $R_{P_0}(\psi_0^*)$. \underline{Bottom row:} Distribution of the true constraint over 1000 realizations for each sample size. The dashed line indicates the equality constraint value of zero. The dotted line indicates the true value of the constraint under $\psi_0$.
}
\label{fig:pse1_sim_by_theta_est_method_all_correct}
\end{figure}

As expected, when all nuisance parameters are consistently estimated, our estimator asymptotically achieves optimal risk and appropriate control of the constraint (Figure \ref{fig:pse1_sim_all_correct}). The various estimators of $\Theta_{\rho_1, P_0}(\psi_0)$ generated estimates of $\psi_0^*$ that had similar performance in finite samples. However, we did find that using the IPW estimator in construction of $\psi_n^*$ led to worse average risk and greater variability in the constraint in finite samples, when compared to the other estimators (Figure \ref{fig:pse1_sim_by_theta_est_method_all_correct}).

\begin{figure}
\centering
\includegraphics[width=4cm, height=3.6cm]{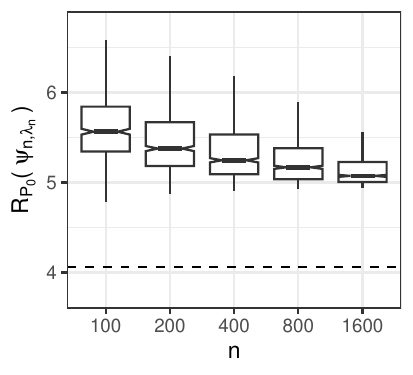}
\includegraphics[width=4cm, height=3.6cm]{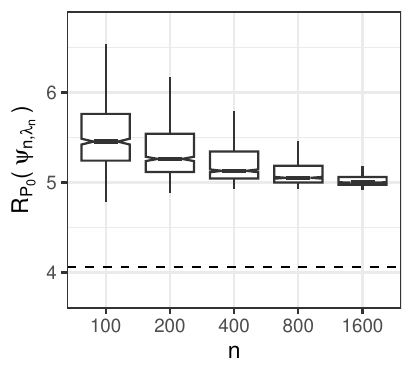}
\includegraphics[width=4cm, height=3.6cm]{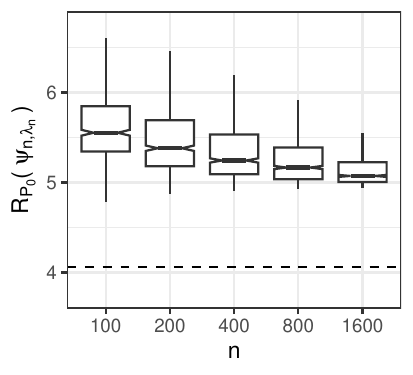}
\includegraphics[width=4cm, height=3.6cm]{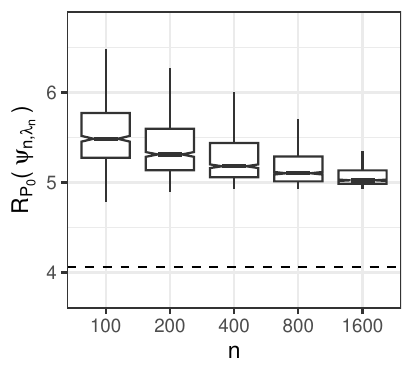}
\includegraphics[width=4cm, height=3.6cm]{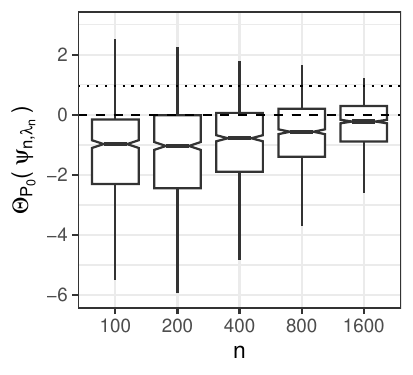}
\includegraphics[width=4cm, height=3.6cm]{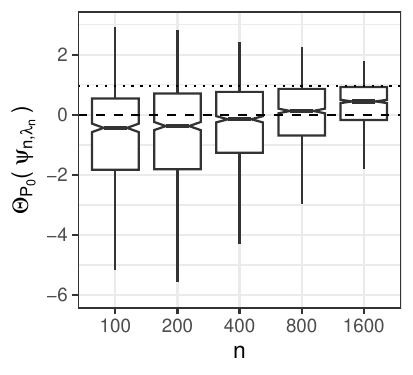}
\includegraphics[width=4cm, height=3.6cm]{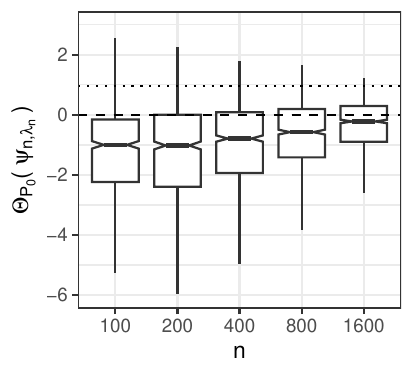}
\includegraphics[width=4cm, height=3.6cm]{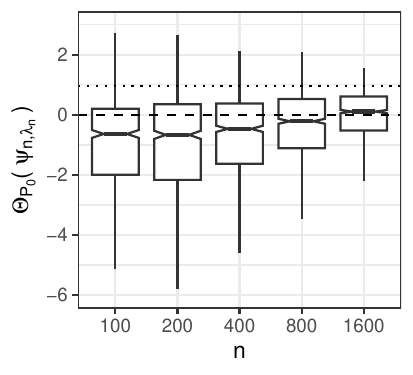}
\caption{\textbf{Estimates of optimal predictions under $\rho_1$-pathway constraint for mean squared error risk when $\psi_0$ is inconsistently estimated.} \underline{Top row:} Various estimators of the constraint are shown; from left to right: $\Theta^\text{plug-in}_{\rho_1, n}$, $\Theta^\text{ipw}_{\rho_1, n}$, $\Theta^\text{ipw-alt}_{\rho_1, n}$, $\Theta^\text{aipw}_{\rho_1, n}$. For each estimator, we show the distribution of risk of $\psi_{n,\lambda_n}$ over 1000 realizations for each sample size for the equality constraint $\Theta_{P_0}(\psi) = 0$. The dashed line indicates the optimal risk $R_{P_0}(\psi_0^*)$. \underline{Bottom row:} Distribution of the true constraint over 1000 realizations for each sample size. The dashed line indicates the equality constraint value of zero. The dotted line indicates the true value of the constraint under $\psi_0$.
}
\label{fig:pse1_sim_EY}
\end{figure}

\begin{figure}
\centering
\includegraphics[width=4cm, height=3.6cm]{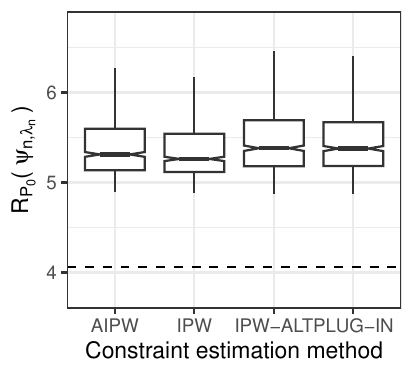}
\includegraphics[width=4cm, height=3.6cm]{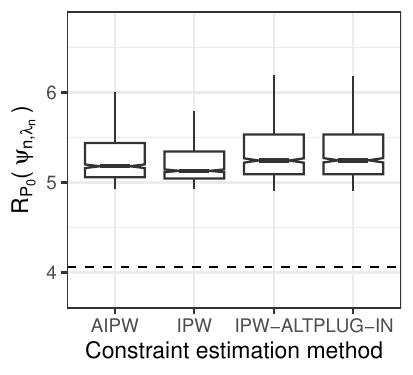}
\includegraphics[width=4cm, height=3.6cm]{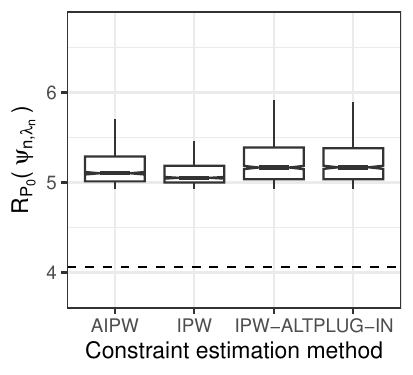}
\includegraphics[width=4cm, height=3.6cm]{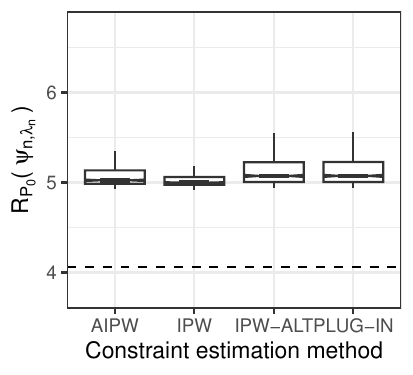}
\includegraphics[width=4cm, height=3.6cm]{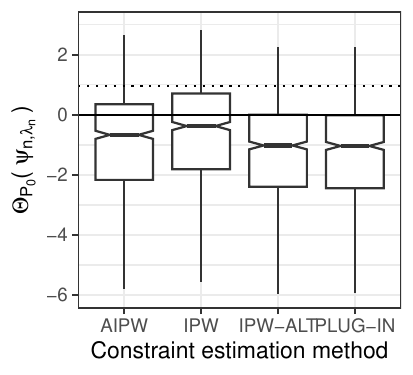}
\includegraphics[width=4cm, height=3.6cm]{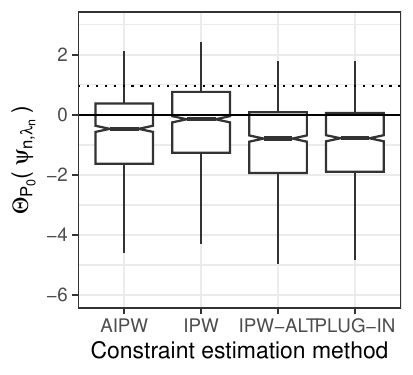}
\includegraphics[width=4cm, height=3.6cm]{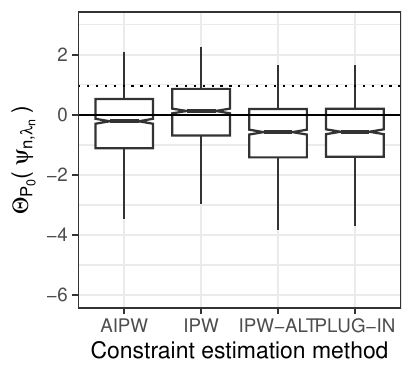}
\includegraphics[width=4cm, height=3.6cm]{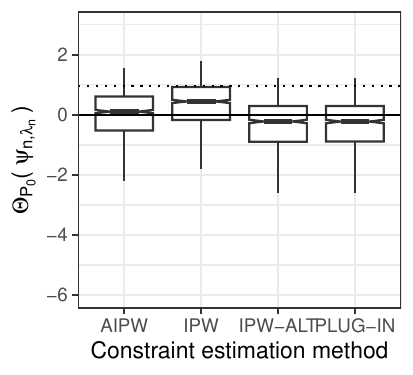}
\caption{\textbf{Comparison of pathway specific effect estimates used in construction of $\psi_n^*$ when $\psi_0$ is inconsistently estimated.} \underline{Top row:} The various estimators are shown in each figure for sample sizes (from left-to-right) of $n=200,400,800, 1600$. For each estimator, we show the distribution of mean squared error of $\psi_{n,\lambda_n}$ over 1000 realizations for each sample size for the equality constraint $\Theta_{P_0}(\psi) = 0$. The dashed line indicates the optimal risk $R_{P_0}(\psi_0^*)$. \underline{Bottom row:} Distribution of the true constraint over 1000 realizations for each sample size. The dashed line indicates the equality constraint value of zero. The dotted line indicates the true value of the constraint under $\psi_0$.
}
\label{fig:pse1_sim_by_theta_est_method_EY}
\end{figure}

When the outcome regression $\psi_0$ is inconsistently estimated, as expected, we found that none of the estimators achieved the optimal constrained risk. On the other hand, we found that constructing $\psi_n^*$ using either the plug-in estimator or the alternative IPW estimator still yielded proper control of the constraint in large samples. This is also predicted by our theory, as these estimators of $\Theta_{\rho_1,0}(\psi_0)$ are consistent for $\Theta_{\rho_1,0}(\tilde{\psi})$ even when $\tilde{\psi} \ne \psi_0$. On the other hand, the IPW and AIPW estimators of $\Theta_{\rho_1,0}(\psi_0)$ are consistent for $\Theta_{\rho_1,0}(\psi_0)$ even when $\psi_0$ is inconsistently estimated. Thus, our theory predicts that estimators of $\psi_0^*$ built using these estimates will not properly control the constraint. We see evidence of this behavior in Figures \ref{fig:pse1_sim_EY} and \ref{fig:pse1_sim_by_theta_est_method_EY}. The result is clear for the IPW estimator, in large samples the constraint is not solved. The results for AIPW are more equivocal, but the trend observed seems to indicate that the value of the pathway specific effect under $\psi_n^*$ tending to a value larger than zero in large samples.

Additional patterns of misspecification are shown in the supplementary material. We found that our theory correctly predicted the behavior of the estimators in large samples. However, we did find that the behavior of the estimators was relatively robust to certain patterns of misspecification. Further discussion and potential explanations for this robustness is included in the supplemental material.

\subsection{High-dimensional covariates and penalized regression}

In this simulation study, we compared the finite-sample performance of our estimators in settings with high-dimensional covariates, when nuisance parameters were estimated using $L^1$-penalized regression (LASSO). In this case, all nuisance parameters are consistently estimated, though penalization should lead to relatively slow convergence to the true nuisance parameter values. Thus, we were interested in understanding finite-sample differences between the performance of our estimators when building the estimators with different estimates of the constraint. We hypothesized that the superior robustness properties of the AIPW estimator could lead to improved performance in small sample sizes, relative to its singly robust counterparts.

We performed three separate simulations with $p = 10, 50, 100$-dimensional covariate vector $X$, respectively. We present here only the results for $p=100$; the other results were similar and can be found in the supplementary materials. The covariates were generated from a standard multivariate Normal distribution. In each simulation, we considered nuisance parameter formulations that were sparse in covariates, only depending on the first five components of $X$. In particular, we set $\pi_0(S = 1 \mid X) = \mbox{expit}(X_1 - X_2/2 + X_3/3 - X_4/4 + X_5/5)$, $f_{0,M}(M = 1 \mid S, X) = \mbox{expit}(-X + X_1 - X_2/2 + X_3/3 - X_4/4 + X_5/5)$, and $f_{0,L}(L = 1 \mid S, X, M) = \mbox{expit}(X - X_1 + X_2/2 - X_3/3 + X_4/4 - X_5/5)$. Finally we generated $Y$ from a Normal distribution with variance 9 and conditional mean $X + L + M + X_1 - X_2/2 + X_3/3 - X_4/4 + X_5/5$. Under this data-generating process, $\Theta_{\rho_1, P_0}(\psi_0) \approx 1.19$. All nuisance parameters were estimated using LASSO with level of penalization selected via 10-fold cross-validation \citep{glmnet_pkg}.

We found that while all estimators exhibited expected asymptotic performance, the estimators constructed based on the plug-in formulation and based on the alternative IPW formulation performed best in finite samples, both in terms of risk, as well as in terms of control of the constraint. The alternative IPW formulation performed slightly better than the plug-in formulation in all sample sizes.

\begin{figure}
\centering
\includegraphics[width=4cm, height=3.6cm]{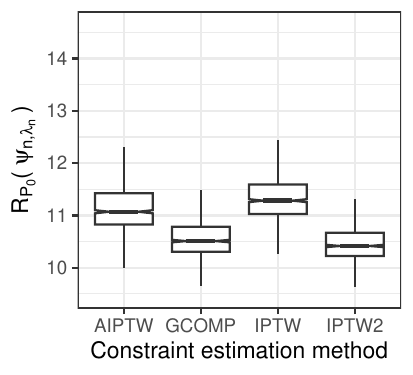}
\includegraphics[width=4cm, height=3.6cm]{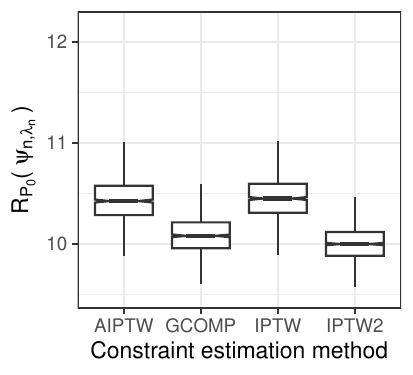}
\includegraphics[width=4cm, height=3.6cm]{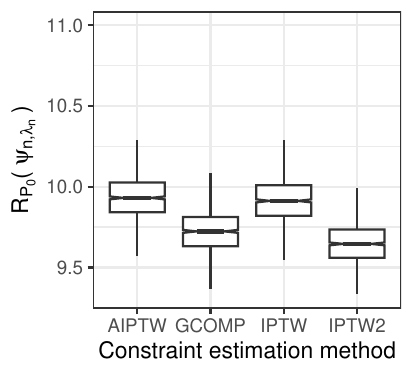}
\includegraphics[width=4cm, height=3.6cm]{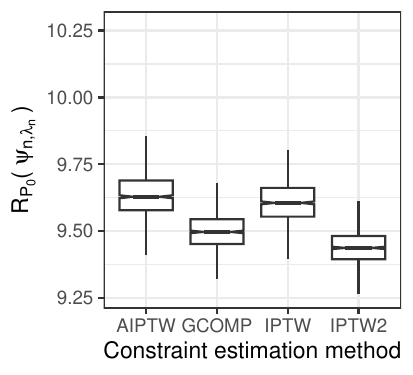}
\includegraphics[width=4cm, height=3.6cm]{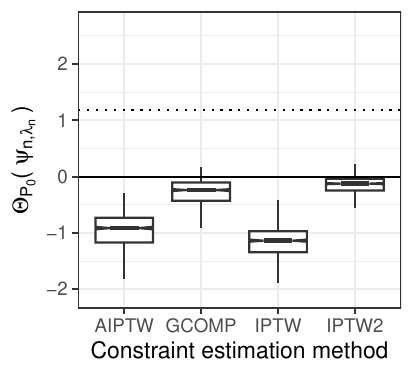}
\includegraphics[width=4cm, height=3.6cm]{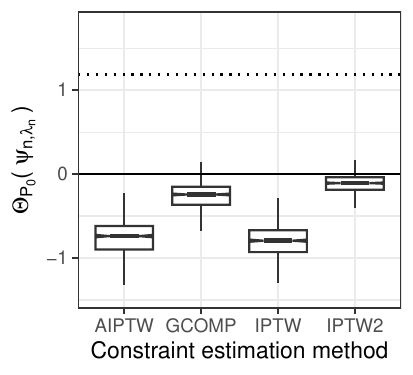}
\includegraphics[width=4cm, height=3.6cm]{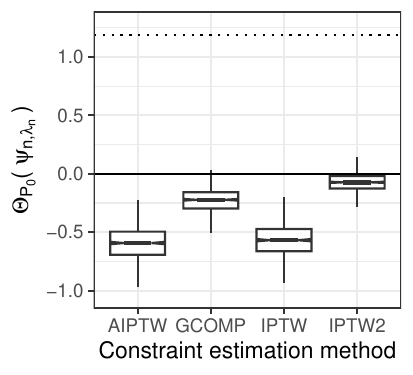}
\includegraphics[width=4cm, height=3.6cm]{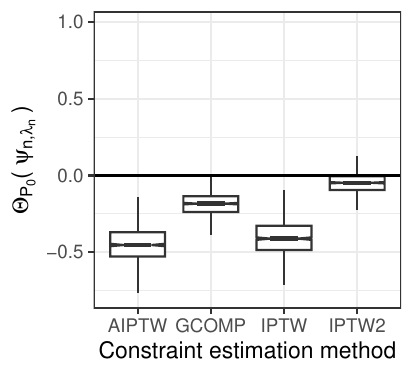}
\caption{\textbf{Comparison of pathway specific effect estimates used in construction of $\psi_n^*$ with high-dimensional covariates ($p = 100$).} \underline{Top row:} The various estimators are shown in each figure for sample sizes (from left-to-right) of $n=200,400,800, 1600$. For each estimator, we show the distribution of mean squared error of $\psi_{n,\lambda_n}$ over 1000 realizations for each sample size for the equality constraint $\Theta_{P_0}(\psi) = 0$. The dashed line indicates the optimal risk $R_{P_0}(\psi_0^*)$. \underline{Bottom row:} Distribution of the true constraint over 1000 realizations for each sample size. The dashed line indicates the equality constraint value of zero. The dotted line indicates the true value of the constraint under $\psi_0$.
}
\label{fig:pse1_sim_by_theta_est_method_glmnet_p100}
\end{figure}

%######################################
\section{Discussion}
\label{sec:conc}
%######################################

% summary 
The development of closed-form solutions for fair optimal predictions marks a significant advancement towards reconciling the often competing objectives of ensuring fairness and optimizing predictive models across various performance metrics. Our framework explains the complexities and nuances involved in embedding fairness into predictive models, demonstrating how fairness considerations can be systematically integrated through modifications to the unconstrained optimal risk minimizer. By providing closed-form solutions that incorporate the constraint, its gradient, and the variance of the gradient, we offer a structured methodology that not only advances theoretical understanding but also offers a pragmatic approach for implementing fairness in algorithmic decision-making. 

It is interesting to observe that in the context of fair prediction using marginal constraints, there seems to exist a ``blessing of positivity violations.'' If there is weak overlap with respect to identification of a $\rho$-specific constraint, then it is generally possible to obtain fair predictions by minimally alternating predictions made for most observations in the support of $W$, while dramatically altering predictions made in the region of  weak overlap. This observation may suggest a limitation of optimal fair prediction under marginal fairness constraints. While such constraints ensure that prediction systems make fair predictions \emph{on average}, our results indicate that relative to an unconstrained system, certain groups of individuals will bear that brunt of the fairness correction, while predictions for other groups will be relatively unchanged. While this is the optimal in a statistical sense, it may be sub-optimal in a practical sense, depending on the specific context. Thus, our results also suggest that it may be important in applications that leverage marginal fairness constraints to scrutinize which groups are more disadvantaged by fairness corrections and to query whether covariate-conditional constraints may be more suitable for achieving appropriate fairness in predictions.

% potential challenges: unknown DAGs + unmeasured confounding 
A potential challenge of our framework is its reliance on a specified DAG, which underpins the causal relationships critical to enforcing fairness. The specification of a DAG can be challenging when the complete causal structure is unknown. In such scenarios, we may employ data-driven causal discovery methods to construct the DAG \citep{spirtes2001causation, glymour2019review}. These approaches, while useful, introduce complex issues related to post-selection inference \citep{gradu2022valid, chang2024post}, which must be carefully addressed to ensure the robustness of causal conclusions. Moreover, the presence of unmeasured confounders poses a significant challenge to the identification of causal path-specific effects and fairness constraints. To navigate these difficulties, researchers might consider adopting more restrictive model assumptions or conducting sensitivity analyses. Development of relevant sensitivity analyses in the context of fair prediction is thus an important step in gauging the reliability of the fairness adjustments implemented.  

% potential challenges: race as treatment 
Furthermore, treating sensitive characteristics such as race or sex as treatments or causal factors is a contentious issue, not only within the realm of algorithmic fairness \citep{Hu2020what, hanna2020towards, hu2023race} but also across the broader field of causal inference \citep{holland1986statistics, vanderweele2013causal, glymour2014commentary, pearl2018does}. Within the research community, divergent viewpoints have surfaced: some researchers advocate concentrating solely on manipulable features—such as perceptions of race—while others challenge the strict manipulationist perspective as overly restrictive for scientific exploration. Although characteristics like race are intrinsically non-manipulable, comprehending how various outcomes and life experiences are counterfactually dependent on an individual's socially constructed race would be essential for analyzing systemic racism and the causal mechanisms driving disparities. Thus, estimating the effect of ``race'' may not necessarily quantify the impact of a specific intervention, but it provides vital insights that can guide policies designed to modify interdependencies among variables, such as ensuring equitable access to quality healthcare irrespective of race. Keeping this line of discussion active and critically examining these perspectives in specific contexts is crucial to advancing our understanding and application of fairness in both theory and practice.

% future direction 
Building on this foundation, future research may explore the extension of our framework to accommodate additional types of fairness constraints and loss functions, further broadening the applicability and impact of our approach. Additionally, empirical studies on diverse datasets across various domains will be invaluable for validating the practical efficacy of our methodology and refining our understanding of its implications for fair machine learning.

%######################################
% \newpage
\bibliographystyle{abbrvnat} 
\bibliography{refs}

\begin{thebibliography}{42}
\providecommand{\natexlab}[1]{#1}
\providecommand{\url}[1]{\texttt{#1}}
\expandafter\ifx\csname urlstyle\endcsname\relax
  \providecommand{\doi}[1]{doi: #1}\else
  \providecommand{\doi}{doi: \begingroup \urlstyle{rm}\Url}\fi

\bibitem[Bang and Robins(2005)]{bang2005doubly}
H.~Bang and J.~M. Robins.
\newblock Doubly robust estimation in missing data and causal inference models.
\newblock \emph{Biometrics}, 61\penalty0 (4):\penalty0 962--973, 2005.

\bibitem[Barocas et~al.(2023)Barocas, Hardt, and Narayanan]{barocas2023fairness}
S.~Barocas, M.~Hardt, and A.~Narayanan.
\newblock \emph{Fairness and machine learning: Limitations and opportunities}.
\newblock MIT Press, 2023.

\bibitem[Benkeser and Ran(2021)]{benkeser2021nonparametric}
D.~Benkeser and J.~Ran.
\newblock Nonparametric inference for interventional effects with multiple mediators.
\newblock \emph{Journal of Causal Inference}, 9\penalty0 (1):\penalty0 172--189, 2021.

\bibitem[Boltyanski et~al.(1998)Boltyanski, Martini, Soltan, and Soltan]{boltyanski1998geometric}
V.~Boltyanski, H.~Martini, V.~Soltan, and V.~P. Soltan.
\newblock \emph{Geometric methods and optimization problems}, volume~4.
\newblock Springer Science \& Business Media, 1998.

\bibitem[Chamon et~al.(2022)Chamon, Paternain, Calvo-Fullana, and Ribeiro]{chamon2022constrained}
L.~F. Chamon, S.~Paternain, M.~Calvo-Fullana, and A.~Ribeiro.
\newblock Constrained learning with non-convex losses.
\newblock \emph{IEEE Transactions on Information Theory}, 69\penalty0 (3):\penalty0 1739--1760, 2022.

\bibitem[Chang et~al.(2024)Chang, Guo, and Malinsky]{chang2024post}
T.-H. Chang, Z.~Guo, and D.~Malinsky.
\newblock Post-selection inference for causal effects after causal discovery.
\newblock \emph{arXiv preprint arXiv:2405.06763}, 2024.

\bibitem[Chernozhukov et~al.(2018)Chernozhukov, Chetverikov, Demirer, Duflo, Hansen, Newey, and Robins]{chernozhukov2018double}
V.~Chernozhukov, D.~Chetverikov, M.~Demirer, E.~Duflo, C.~Hansen, W.~Newey, and J.~Robins.
\newblock Double/debiased machine learning for treatment and structural parameters, 2018.

\bibitem[Chiappa(2019)]{chiappa2019path}
S.~Chiappa.
\newblock Path-specific counterfactual fairness.
\newblock In \emph{Proceedings of the AAAI Conference on Artificial Intelligence}, volume~33, pages 7801--7808, 2019.

\bibitem[Corbett-Davies and Goel(2018)]{corbett2018measure}
S.~Corbett-Davies and S.~Goel.
\newblock The measure and mismeasure of fairness: A critical review of fair machine learning.
\newblock \emph{arXiv preprint arXiv:1808.00023}, 2018.

\bibitem[Donini et~al.(2018)Donini, Oneto, Ben-David, Shawe-Taylor, and Pontil]{donini2018empirical}
M.~Donini, L.~Oneto, S.~Ben-David, J.~S. Shawe-Taylor, and M.~Pontil.
\newblock Empirical risk minimization under fairness constraints.
\newblock \emph{Advances in neural information processing systems}, 31, 2018.

\bibitem[Friedler et~al.(2021)Friedler, Scheidegger, and Venkatasubramanian]{friedler2021possibility}
S.~A. Friedler, C.~Scheidegger, and S.~Venkatasubramanian.
\newblock The (im)possibility of fairness: Different value systems require different mechanisms for fair decision making.
\newblock \emph{Communications of the ACM}, 64\penalty0 (4):\penalty0 136--143, 2021.

\bibitem[Glymour and Glymour(2014)]{glymour2014commentary}
C.~Glymour and M.~R. Glymour.
\newblock Commentary: race and sex are causes.
\newblock \emph{Epidemiology}, 25\penalty0 (4):\penalty0 488--490, 2014.

\bibitem[Glymour et~al.(2019)Glymour, Zhang, and Spirtes]{glymour2019review}
C.~Glymour, K.~Zhang, and P.~Spirtes.
\newblock Review of causal discovery methods based on graphical models.
\newblock \emph{Frontiers in genetics}, 10:\penalty0 524, 2019.

\bibitem[Gradu et~al.(2022)Gradu, Zrnic, Wang, and Jordan]{gradu2022valid}
P.~Gradu, T.~Zrnic, Y.~Wang, and M.~I. Jordan.
\newblock Valid inference after causal discovery.
\newblock \emph{arXiv preprint arXiv:2208.05949}, 2022.

\bibitem[Hanna et~al.(2020)Hanna, Denton, Smart, and Smith-Loud]{hanna2020towards}
A.~Hanna, E.~Denton, A.~Smart, and J.~Smith-Loud.
\newblock Towards a critical race methodology in algorithmic fairness.
\newblock In \emph{Proceedings of the 2020 conference on fairness, accountability, and transparency}, pages 501--512, 2020.

\bibitem[Holland(1986)]{holland1986statistics}
P.~W. Holland.
\newblock Statistics and causal inference.
\newblock \emph{Journal of the American statistical Association}, 81\penalty0 (396):\penalty0 945--960, 1986.

\bibitem[Hu(2023)]{hu2023race}
L.~Hu.
\newblock What is “race” in algorithmic discrimination on the basis of race?
\newblock \emph{Journal of Moral Philosophy}, 1\penalty0 (aop):\penalty0 1--26, 2023.

\bibitem[Hu and Kohler-Hausmann(2020)]{Hu2020what}
L.~Hu and I.~Kohler-Hausmann.
\newblock What's sex got to do with machine learning?
\newblock In \emph{Proceedings of the 2020 Conference on Fairness, Accountability, and Transparency}, 2020.

\bibitem[Kleinberg et~al.(2017)Kleinberg, Mullainathan, and Raghavan]{kleinberg2016inherent}
J.~Kleinberg, S.~Mullainathan, and M.~Raghavan.
\newblock Inherent trade-offs in the fair determination of risk scores.
\newblock \emph{arXiv preprint arXiv:1609.05807}, 2017.

\bibitem[Kusner et~al.(2017)Kusner, Loftus, Russell, and Silva]{kusner2017counterfactual}
M.~J. Kusner, J.~Loftus, C.~Russell, and R.~Silva.
\newblock Counterfactual fairness.
\newblock In \emph{Advances in Neural Information Processing Systems}, volume~30. PMLR, 2017.

\bibitem[Makhlouf et~al.(2020)Makhlouf, Zhioua, and Palamidessi]{makhlouf2020survey}
K.~Makhlouf, S.~Zhioua, and C.~Palamidessi.
\newblock Survey on causal-based machine learning fairness notions.
\newblock \emph{arXiv preprint arXiv:2010.09553}, 2020.

\bibitem[Miles et~al.(2017)Miles, Shpitser, Kanki, Meloni, and Tchetgen~Tchetgen]{miles2017quantifying}
C.~H. Miles, I.~Shpitser, P.~Kanki, S.~Meloni, and E.~J. Tchetgen~Tchetgen.
\newblock Quantifying an adherence path-specific effect of antiretroviral therapy in the nigeria pepfar program.
\newblock \emph{Journal of the American Statistical Association}, 112\penalty0 (520):\penalty0 1443--1452, 2017.

\bibitem[Miles et~al.(2020)Miles, Shpitser, Kanki, Meloni, and Tchetgen~Tchetgen]{miles2020semiparametric}
C.~H. Miles, I.~Shpitser, P.~Kanki, S.~Meloni, and E.~J. Tchetgen~Tchetgen.
\newblock On semiparametric estimation of a path-specific effect in the presence of mediator-outcome confounding.
\newblock \emph{Biometrika}, 107\penalty0 (1):\penalty0 159--172, 2020.

\bibitem[Mitchell et~al.(2021)Mitchell, Potash, Barocas, D'Amour, and Lum]{mitchell2018prediction}
S.~Mitchell, E.~Potash, S.~Barocas, A.~D'Amour, and K.~Lum.
\newblock Prediction-based decisions and fairness: A catalogue of choices, assumptions, and definitions.
\newblock \emph{Annual Review of Statistics and Its Application}, 2021.

\bibitem[Nabi and Shpitser(2018)]{nabi2018fair}
R.~Nabi and I.~Shpitser.
\newblock Fair inference on outcomes.
\newblock In \emph{Proceedings of the AAAI Conference on Artificial Intelligence}, volume~32, 2018.

\bibitem[Nabi et~al.(2022)Nabi, Malinsky, and Shpitser]{nabi2022optimal}
R.~Nabi, D.~Malinsky, and I.~Shpitser.
\newblock Optimal training of fair predictive models.
\newblock In \emph{Conference on Causal Learning and Reasoning}, pages 594--617. PMLR, 2022.

\bibitem[Nabi et~al.(2024)Nabi, Hejazi, van~der Laan, and Benkeser]{nabi2024statistical}
R.~Nabi, N.~S. Hejazi, M.~J. van~der Laan, and D.~Benkeser.
\newblock Statistical learning for constrained functional parameters in infinite-dimensional models with applications in fair machine learning.
\newblock \emph{arXiv preprint arXiv:2404.09847}, 2024.

\bibitem[Nilforoshan et~al.(2022)Nilforoshan, Gaebler, Shroff, and Goel]{nilforoshan2022causal}
H.~Nilforoshan, J.~D. Gaebler, R.~Shroff, and S.~Goel.
\newblock Causal conceptions of fairness and their consequences.
\newblock In \emph{International Conference on Machine Learning}, pages 16848--16887. PMLR, 2022.

\bibitem[Pearl(2009)]{pearl2009causality}
J.~Pearl.
\newblock \emph{Causality: Models, Reasoning, and Inference}.
\newblock Cambridge University Press, 2009.

\bibitem[Pearl(2018)]{pearl2018does}
J.~Pearl.
\newblock Does obesity shorten life? or is it the soda? on non-manipulable causes.
\newblock \emph{Journal of Causal Inference}, 6\penalty0 (2):\penalty0 20182001, 2018.

\bibitem[Plecko and Bareinboim(2022)]{plecko2022causal}
D.~Plecko and E.~Bareinboim.
\newblock Causal fairness analysis.
\newblock \emph{arXiv preprint arXiv:2207.11385}, 2022.

\bibitem[Shpitser and Tchetgen~Tchetgen(2016)]{shpitser2016causal}
I.~Shpitser and E.~Tchetgen~Tchetgen.
\newblock Causal inference with a graphical hierarchy of interventions.
\newblock \emph{Annals of statistics}, 44\penalty0 (6):\penalty0 2433, 2016.

\bibitem[Simon et~al.(2011)Simon, Friedman, Hastie, and Tibshirani]{glmnet_pkg}
N.~Simon, J.~Friedman, T.~Hastie, and R.~Tibshirani.
\newblock Regularization paths for cox's proportional hazards model via coordinate descent.
\newblock \emph{Journal of Statistical Software}, 39\penalty0 (5):\penalty0 1--13, 2011.
\newblock \doi{10.18637/jss.v039.i05}.

\bibitem[Spirtes et~al.(2001)Spirtes, Glymour, and Scheines]{spirtes2001causation}
P.~Spirtes, C.~Glymour, and R.~Scheines.
\newblock \emph{Causation, prediction, and search}.
\newblock MIT press, 2001.

\bibitem[Sundaram(1996)]{sundaram1996first}
R.~K. Sundaram.
\newblock \emph{A first course in optimization theory}.
\newblock Cambridge university press, 1996.

\bibitem[Tchetgen and Shpitser(2012)]{tchetgen2012semiparametric}
E.~J.~T. Tchetgen and I.~Shpitser.
\newblock Semiparametric theory for causal mediation analysis: efficiency bounds, multiple robustness, and sensitivity analysis.
\newblock \emph{Annals of statistics}, 40\penalty0 (3):\penalty0 1816, 2012.

\bibitem[Van~der Vaart(2000)]{van2000asymptotic}
A.~W. Van~der Vaart.
\newblock \emph{Asymptotic statistics}, volume~3.
\newblock Cambridge university press, 2000.

\bibitem[VanderWeele and Hernan(2013)]{vanderweele2013causal}
T.~J. VanderWeele and M.~A. Hernan.
\newblock Causal inference under multiple versions of treatment.
\newblock \emph{Journal of causal inference}, 1\penalty0 (1):\penalty0 1--20, 2013.

\bibitem[Zafar et~al.(2019)Zafar, Valera, Gomez-Rodriguez, and Gummadi]{zafar2019fairness}
M.~B. Zafar, I.~Valera, M.~Gomez-Rodriguez, and K.~P. Gummadi.
\newblock Fairness constraints: A flexible approach for fair classification.
\newblock \emph{Journal of Machine Learning Research}, 20\penalty0 (75):\penalty0 1--42, 2019.

\bibitem[Zhang and Bareinboim(2018)]{zhang2018fairness}
J.~Zhang and E.~Bareinboim.
\newblock Fairness in decision-making—the causal explanation formula.
\newblock In \emph{Proceedings of the AAAI Conference on Artificial Intelligence}, volume~32, 2018.

\bibitem[Zhang et~al.(2017)Zhang, Wu, and Wu]{zhang2017causal}
L.~Zhang, Y.~Wu, and X.~Wu.
\newblock A causal framework for discovering and removing direct and indirect discrimination.
\newblock In \emph{Association for the Advancement of Artificial Intelligence}, 2017.

\bibitem[Zhou(2022)]{zhou2022semiparametric}
X.~Zhou.
\newblock Semiparametric estimation for causal mediation analysis with multiple causally ordered mediators.
\newblock \emph{Journal of the Royal Statistical Society Series B: Statistical Methodology}, 84\penalty0 (3):\penalty0 794--821, 2022.

\end{thebibliography}
%######################################

\newpage 
\begin{appendix}

{\Large \bf APPENDIX}

The appendix is organized as follows. 
Appendix~\ref{app:notation} offers a summary of the notations used throughout the manuscript to aid in understanding and reference. 
Appendix~\ref{app:constraint-specific-path} provides details on the characterization of the constraint-specific path using the methodology developed by \citet{nabi2024statistical}. 
Appendix~\ref{app:more_on_pse} contains additional details on the conceptualizations of unfairness, including variations in our results when the direct effect is excluded from the set of considered unfair pathways. It also discusses methods for measuring the unfair effect on a log-odds ratio scale within the cross-entropy risk framework. 
Appendix~\ref{app:proofs} presents the proofs for all the results mentioned in the manuscript. 
Appendix~\ref{app:sims} includes additional simulation results.  

%##############################################
\section{Glossary of terms and notations} 
\label{app:notation} 
%##############################################

%To simplify navigation through the manuscript's notations, a detailed list is provided in Table~\ref{tab:notations}. 

\begin{table}[!h]
\begin{center}
\caption{\centering Glossary of terms and notations}
\label{tab:notations}
\addtolength{\tabcolsep}{0pt}
{\small
\begin{tabular}{ ll | ll} 
    \hline  
    \textbf{Symbol} & \textbf{Definition} & \textbf{Symbol} & \textbf{Definition}  
    \\ \hline 
    $S, Y$  & Sensitive, outcome variables  &  $\mathcal{M}$ & Statistical model    \\ 
    $W$ & Additional covariates  & $P_0 \in \mathcal{M}$ & True data distribution  \\ 
    $Z$ & Collection of $S$ and $W$ & $P \in \mathcal{M}$ & An arbitrary distribution  \\
    $O$ & Observed variables, $(S, W, Y)$ & $\G$ & Directed acyclic graph \\ 
    $\psi_0$ & Risk minimizer under $P_0$  & $\pa_\G(O_i)$ & Parents of $O_i$ in $\G$ \\ 
    $\Psi$ & Parameter space &  $f_{O_i}( . )$ & NPSEM for $O_i$ \\
    $L(\psi)$ & Loss function & $\rho$ & Collection of unfair paths \\
    $R_{P_0}(\psi)$ & Risk function, $\int L(\psi)(o)dP_0(o)$ & $\epsilon_{O_i}$ & Error term in NPSEM \\ 
    $\Delta^\rho$ & Constraint on difference scale & $\Pi^\rho$ & Constraint on log-odds ratio scale \\
    $\Theta_{\Delta, P_0}(\psi)$ & Constraint ID functional for  $\Delta^\rho$  & 
    $\Theta_{\Pi, P_0}(\psi)$ & Constraint ID functional for  $\Pi^\rho$
    \\ 
    $\lambda$ & Lagrange multiplier & $|| h ||$ & $L^2_P$-norm of function $h$ \\
    $\psi^*_0$ & Constrained risk minimizer &  $D_{R, P_0}(\psi)$ & Canonical gradient of the risk \\  
    $\pi_0(s \mid x)$ & $P_0(S= s \mid x)$ &  $D_{\Theta_\Delta, P}(\psi)$ & Canonical gradient of constraint $\Delta^\rho$ \\  
     $f_{0, Z}(z \mid \pa_\G(z))$ & $P_0(Z = z \mid \pa_\G(z))$ &  $\sigma^2(D_{\Theta_\Delta, P_0})$ & Variance of $D_{\Theta_\Delta, P_0}(\psi)$ \\  
    $\psi_n, \pi_n, f_{n, Z}$ & Nuisance estimates & $D_{\Theta_\Pi, P_0}(\psi)$ & Canonical gradient of constraint $\Pi^\rho$  \\ 
    % $\tangent_P(\psi)$ & \multicolumn{3}{l}{Closure in $\hilbert_P(\psi)$ of linear span of  $\mathcal{H}_P(\psi)$} \\
    % \multicolumn{4}{l}{$D_{R, P}(\psi), D_{\Theta, P}(\psi), D_{\Omega, P}(\psi) \quad $ Gradients of risk, equality constraint, inequality constraint} 
\end{tabular}
}
\end{center}
\end{table}

%######################################
\section{More details on constraint-specific path}
\label{app:constraint-specific-path}
%######################################

Here, we provide an overview of the characterization of the constraint-specific path developed by \citet{nabi2024statistical}. For a comprehensive treatment, refer to the original paper. 

For each function $\psi$ within the parameter space $\Psi$, we define a \emph{path} through $\psi$ characterized by a specific \emph{direction} $h$, where $h$ is a function dependent on $Z$. This path is represented as $\{\psi_{\delta, h} : \delta \in \mathbb{R}\} \subset \Psi$ and is parameterized by the scalar $\delta$. We describe this path as \emph{through} $\psi$ if, at $\delta = 0$, $\psi_{\delta,h}$ equals $\psi$. 

The \emph{direction} of the path is defined by the derivative $\frac{d}{d\delta} \psi_{\delta,h} |_{\delta = 0}$. We embed the set of all possible directions from $\psi$ in the parameter space $\Psi$ within the Hilbert space $\hilbert_{P}(\psi)$, equipped with a covariance inner product $\langle f,g \rangle = \int f(z) g(z) dP_{Z}(z)$. This embedding facilitates an inner-product characterization of \emph{bounded linear functionals} on $\hilbert_P(\psi)$ of the form: $f_P(\psi_{\delta,\cdot}): \hilbert_P(\psi) \rightarrow \mathbb{R}$, which  maps $h \in \hilbert_P(\psi)$ to $\mathbb{R}$. A functional $f_P(\psi_{\delta,\cdot})$ is called \emph{linear} on $\hilbert_P(\psi)$ if for any $h_1, h_2 \in \hilbert_P(\psi)$ and $\alpha, \beta \in \mathbb{R}$, it holds that $f_P(\psi_{\delta,\alpha h_1 + \beta h_2}) = \alpha f_P(\psi_{\delta, h_1}) + \beta f_P(\psi_{\delta, h_2})$. A functional $f_P(\psi_{\delta,\cdot})$ is called \emph{bounded} on $\hilbert_P(\psi)$ if for any $h \in \hilbert_P(\psi)$ and $M > 0$, it holds that $|f_P(\psi_{\delta, h})| \le M \langle h, h \rangle^{1/2}$. Leveraging the Reisz representation theorem, any bounded linear functional on $\hilbert_P(\psi)$ can be represented as an inner product, $f_P(\psi_{\delta,h}) = \langle h, D_P(\psi) \rangle$, with $D_P(\psi)$ being the unique canonical gradient of the functional within $\hilbert_P(\psi)$.

In this paper, we consider $\hilbert_{P}(\psi) = L^2(P_{Z})$, consisting of all square-integrable functions $h$ of $Z$. Our methodology involves examining the derivatives of the risk $R_P(\psi)$ and constraint $\Theta_{P}(\psi)$ along paths $\{\psi_{\delta, h} : \delta \in \mathbb{R}\}$ for $h \in L^2(P_{Z})$. These derivatives, evaluated at $\delta = 0$, are conceptualized as a functional $f_P$ on $L^2(P_{Z})$, with the mapping defined by $h \rightarrow \frac{d}{d\delta} f_P(\psi_{\delta, h})|_{\delta = 0}$. The pathwise derivatives of the risk $R_P(\psi)$ (MSE and cross-entropy) and constraint $\Theta_P(\psi)$ ($\rho$-specific effect) are bounded linear functionals on $L^2(P_{Z})$. Thus, we have the following inner-product representations: 
\begin{align*}
  \frac{d}{d\delta} R_P(\psi_{\delta,h})\bigg|_{\delta = 0} = \langle
  D_{R,P}(\psi), h \rangle \quad \mbox{and} \quad \frac{d}{d\delta}
  \Theta_P(\psi_{\delta, h}) \bigg|_{\delta = 0} = \langle D_{\Theta,P}(\psi),
  h \rangle \ ,
\end{align*}
for uniquely defined canonical gradients $D_{R,P}$ and $D_{\Theta,P}$ within the Hilbert space $L^2(P_{Z})$. 

For a given $\lambda$, we have $\psi_{0, \lambda} = \argmin_{\psi \in {\bf \Psi}} R_{P_0}(\psi) + \lambda \Theta_{P_0}(\psi)$ as the minimizer of penalized risk under sampling from $P_0$. The constraint-specific path $\{\psi_{0, \lambda} : \lambda \in \openr\}$ is then characterized as the collection of minimizers of the penalized risk under $P_0$. For any $\psi_{0, \lambda}$ on this path, it holds that $\psi_{0, \lambda}$ minimizes the criterion $\psi \rightarrow R_{P_0}(\psi) + \lambda \Theta_{P_0}(\psi)$. We can define a path through $\psi_{0, \lambda}$, $\{\psi_{0, \lambda, \delta} : \delta \in \openr\}$. Thus, for any $h \in L^2(P_Z)$, a derivative of the minimization criterion along this path equals zero, that is $\frac{d}{d \delta}R_{P_0}(\psi_{0, \lambda, \delta, h})|_{\delta = 0} + \lambda \left. \frac{d}{d \delta} \Theta_{P_0}(\psi_{0, \lambda, \delta, h}) \right |_{\delta = 0} = 0$. Given the inner-product representations, we get: 
\begin{align*}
  D_{R, P_0}(\psi_{0, \lambda}) + \lambda D_{\Theta,P_0}(\psi_{0,\lambda}) = 0 \ , \forall \lambda \in \mathbb{R} \ . 
\end{align*}%
The constrained risk minimmizer is established as $\psi^*_0 = \psi_{0, \lambda_0}$ where $\lambda_0$ is selected such that $\Theta_{P_0}(\psi_{0, \lambda_0}) = 0$.  

\begin{figure}[!t]
    \centering
    \scalebox{0.9}{
	\begin{tikzpicture}[
		bullet/.style={circle,fill,inner sep=2pt}, 
		long dash/.style={dash pattern=on 10pt off 2pt}, 
		decoration = {snake, pre length=0pt,post length=0pt,}
		]
        \begin{scope}[xshift=0cm, yshift=0cm]
    		% parameter space 
    		\draw[black, thick, long dash, fill=orange!2] plot[smooth cycle, tension=1]
    		% coordinates{(0,1.25) (3,2.8) (6.75, 1.5) (8.25,0.) (6.,-2) (0,-2)};
            coordinates{(0,-2.) (0, 2.) (7, 2.) (7, -2.)}; 
    		\node[] at (0.25, 2.5){$\Psi$}; % (parameter space)};  
    		
    		% constrained path
    		\draw[thick] (0.2,-0.4)  node[bullet]{} to[out=40,in=175] (6.75,0.4) node[bullet]{}; 
    		\node[] at (0.4,-0.7) {$\psi_{0} = \psi_{0, \lambda = 0}$}; 
    		\node[] at (7.0, 0.) {$\psi^*_{0} = \psi_{0, \lambda_0}$};
    		
    		\node[bullet] at (3.1,0.7){};
    		\node[] at (4.95, 0.9){$\psi_{0, \lambda} = \psi_{0,\lambda,\delta_j, h_j}|_{\delta_j = 0}$};  		
    		% directions and scores 
    		\draw[dashed, thick, decorate, segment length=200] (2.4,-1) -- (3.8,2.5); 
    		\draw[dashed, thick, decorate, segment length=200] (3.9,-1) -- (2.5,2); 
    		\node[] at (1.7, 1.65){$\psi_{0, \lambda, \delta_1, h_1}$}; 
    		\node[] at (4.6, 2.2){$\psi_{0, \lambda, \delta_2, h_2}$};  
    		
    		% horizental axis for lambda
    		\draw[very thick] (1.5,-1.9) -- (2.65,-1.9);
    		\node[] at (4.25, -1.85){$\{ \psi_{0, \lambda}: \ \lambda \in \openr \ \}$};   	
        \end{scope} 
  
	\end{tikzpicture}
    }
    % \vspace{-0.5cm}
    \caption{A schematic shows the constraint-specific path indexed by Lagrange multiplier $\lambda$. It includes the unconstrained risk minimizer $\psi_0$ and the constrained minimizer $\psi_{0, \lambda_0}$. }
    \label{fig:schematic}
\end{figure}
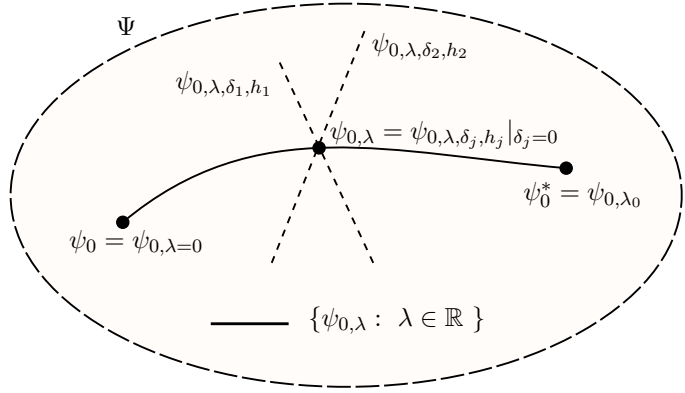
Figure~\ref{fig:schematic} offers a visual representation of the above constraint-specific path, with the dashed oval shape denoting the parameter space $\Psi$ and the solid black line denoting the constrained-specific path $\{\psi_{0, \lambda}, \lambda \in \mathbb{R}\}$. This path represents how adjustments to $\lambda$ shift $\psi_0$ to minimize the combined risk and constraint measure, inherently defining the trajectory of $\psi_0$ within the model's constraint framework. The unconstrained parameter $\psi_{0}$ and the constrained parameter $\psi^*_0$ are both part of this path. 

%######################################
\section{More on notions of unfair effects} 
\label{app:more_on_pse}
%######################################

\subsection{On the inclusion or exclusion of direct pathway in unfair pathways} 
\label{app:in_ex_direct_eff}

In the main draft, we assumed the direct causal path $S \rightarrow Y$ is included in the set of unfiar pathways $\rho$. Here, we consider the general case where the direct effect may or may not be included in $\rho$. Let $s_y \in \{0, 1\}$ represent the assignment of $S$ in the counterfactual outcome, with the alternative assignment denoted by $s'_y$. We can subsequently modify Lemma~\ref{lem:id_constraints} as follows. Consider an NPSEM-IE associated with DAG $\mathcal{G}$. Asume a $\rho$-specific effect $\Delta^\rho$, as detailed in Definition~\ref{def:unfair_eff}, does not yield any recanting witnesses. Consequently, the identification functional, denoted by $\Theta_{\Delta, P_0}(\psi_0)$, is given by:
{\small 
\begin{equation}\label{eq:pse_id_general}
\begin{aligned}
    \Theta_{\Delta, P_0}(\psi_0) 
    &= \I(s_y=1) \bigg\{ \displaystyle \int  \Big\{ \psi_0(1, w) \prod_{L_i \in \mathbb{L}_\rho \setminus Y} dP_0(\ell_i \mid \pa_\G(\ell_i)) \Big|_{S=1} - \psi_0(0, w) \prod_{L_i \in \mathbb{L}_\rho \setminus Y} dP_0(\ell_i \mid \pa_\G(\ell_i)) \Big|_{S=0} \Big\} \\ 
    &\hspace{3cm} \times \prod_{M_i \in \mathbb{M}_\rho} dP_0(m_i \mid \pa_\G(m_i)) \Big|_{S=0} \  dP_0(x) \bigg\} \\
    &\hspace{0.25cm} + \I(s_y = 0) \bigg\{ \displaystyle \int  \Big\{ \prod_{M_i \in \mathbb{M}_\rho} dP_0(m_i \mid \pa_\G(m_i)) \Big|_{S=1} -  \prod_{M_i \in \mathbb{M}_\rho} dP_0(m_i \mid \pa_\G(m_i)) \Big|_{S=0} \Big\} \\ 
    &\hspace{3cm} \times \psi_0(0, w) \prod_{L_i \in \mathbb{L}_\rho} dP_0(\ell_i \mid \pa_\G(\ell_i)) \Big|_{S=0} \  dP_0(x)  \bigg\} \ . 
\end{aligned}
\end{equation}
}

As an example, consider the DAG in Figure~\ref{fig:dag_examples}(c), and assume $\rho = \{S \rightarrow L \rightarrow Y\}$. Then, $Y(1, 0; \rho)$ is defined as $Y(0, M(0), L(1, M(0)))$ with $s_y = 0$, $\mathbb{M}_\rho = \{L\}$, and $\mathbb{L}_\rho = \{M, Y\}$. According to Equation~\eqref{eq:pse_id_general}, the $\rho$-specific effect is identified as 
{\small 
\begin{align*}
    \displaystyle \int  \psi_0(S=0, m, \ell, x) dP_0(m \mid S=0, x) \left\{ dP_0(\ell \mid S=1, m, x) - dP_0(\ell \mid S=0, m, x)  \right\}  dP_0(x) \ . 
\end{align*}
}%

The expression of the canonical gradient of the constraint in Lemma~\ref{lemma:gradients} is also modified as follows. Assuming the constraint is defined in Definition~\ref{def:unfair_eff} and identified via \eqref{eq:pse_id_general}, the canonical gradient is expressed as: 
{\small 
\begin{equation}\label{eq:theta_gradient_general}
\begin{aligned}
    D_{\Theta_\Delta, P_0}(Z) 
    &= \I(s_y = 1) \left\{ \frac{2S - 1}{P_0(S \mid \pa_\G(S))} \ \prod_{M_i \in \mathbb{M}_\rho} \frac{P_0(M_i \mid \pa_\G(M_i) \setminus S, S = 0)}{P_0(M_i \mid \pa_\G(M_i))} \right\} \\
    &\hspace{0.25cm}+\I(s_y = 0) \left\{ \frac{1-S}{P_0(S \mid \pa_\G(S))} \left\{ \prod_{M_i \in \mathbb{M}_\rho} \frac{P_0(M_i \mid \pa_\G(M_i) \setminus S, S = 1)}{P_0(M_i \mid \pa_\G(M_i))} - 1 \right\} \right\} .
\end{aligned}
\end{equation}
}%
For a proof, see Appendix~\ref{app:proofs_gradients}. 

It still holds that the constraint gradient is mean zero, i.e., 
$$\E[D_{\Theta_\Delta, P_0}(Z) \mid s_y=1] =  \E[D_{\Theta_\Delta, P_0}(Z) \mid s_y=0] = 0.$$  

We note that the statement of Theorems~\ref{thm:closed-mse} and \ref{thm:closed-cross} do not change. The closed-form solution for $\psi_{0, \lambda}(z)$ under the cross-entropy risk in Lemma~\ref{lem:closed_quadratic_cross} modifies to be: 
\begin{align} 
    \psi_{0, \lambda}(s,w) = \frac{ 1 + \lambda D_{\Theta_\Delta, P_0}(s,w) - (2s - 1)  \left[ \left\{ 1 + \lambda D_{\Theta_\Delta, P_0}(s,w) \right\}^2 - 4\lambda \psi_0(z)D_{\Theta_\Delta, P_0}(s,w) \right]^{1/2}  }{2\lambda D_{\Theta_\Delta, P_0}(s,w)} \ , \label{eq:psi0lambda_cross_general}   
\end{align}%
where $D_{\Theta_\Delta, P_0}(S, W)$ is given by \eqref{eq:theta_gradient_general}. For a proof, see Appendix~\ref{app:quadratic_equation}. 

\subsection{Fairness on a log-odds ratio scale}
\label{app:odds_ratio_scale}

% In order to encompass scenarios where the causal relationship between $S$ and $Y$ is more appropriately quantified on a multiplicative scale, we consider the odds ratio scale. 
% \begin{definition}[Unfair effect on an odds ratio scale]
%     Given a collection of unfair paths $\rho$, the unfair effect of $S$ on $Y$ on an odds ratio scale is defined as the $\rho$-specific effect, denoted by $\Pi^\rho$ and expressed as:  
%     \begin{align}
%        \Pi^\rho =  \E\left[ \log \frac{  \E[ Y(1, 0; \rho) \mid X] }{1- \E[ Y(1, 0; \rho) \mid X] } - \log \frac{  \E[ Y(0) \mid X] }{1- \E[ Y(0) \mid X] } \right] \ .
%     \end{align}
% \label{def:unfair_eff_odds-ratio}
% \end{definition}

In this supplemental section, we discuss a notable observation regarding the cross-entropy risk framework. Specifically, we note that the constrained parameter $\psi^*_0$ can be determined in a closed-form solution when examining the constraint on a different scale. This involves considering a logit transformation of $\psi_0(s, w)$, defined as $\text{logit}(\psi_0(s, w)) = \log\left(\frac{\psi_0(s, w)}{1 - \psi_0(s, w)}\right)$. Given an NPSEM-IE associated with DAG $\G$, we consider the $\rho$-specific effect on a log-odds ratio scale to be denoted by $\Pi^\rho$ and, in the absence of no recanting witnesses, identified as: 
{\small 
\begin{equation}\label{eq:pse_id_odds_general}
\begin{aligned}
    &\Theta_{\Pi, P_0}(\psi_0) \\
    &= \I(s_y=1) \bigg\{ \displaystyle \int  \Big\{ \text{logit}(\psi_0(1, w)) \!\!\! \prod_{L_i \in \mathbb{L}_\rho \setminus Y} dP_0(\ell_i \mid \pa_\G(\ell_i)) \Big|_{S=1} - \text{logit}(\psi_0(0, w)) \!\!\! \prod_{L_i \in \mathbb{L}_\rho \setminus Y} dP_0(\ell_i \mid \pa_\G(\ell_i)) \Big|_{S=0} \Big\} \\ 
    &\hspace{3cm} \times \prod_{M_i \in \mathbb{M}_\rho} dP_0(m_i \mid \pa_\G(m_i)) \Big|_{S=0} \  dP_0(x) \bigg\} \\
    &\hspace{0.25cm} + \I(s_y = 0) \bigg\{ \displaystyle \int  \Big\{ \prod_{M_i \in \mathbb{M}_\rho} dP_0(m_i \mid \pa_\G(m_i)) \Big|_{S=1} -  \prod_{M_i \in \mathbb{M}_\rho} dP_0(m_i \mid \pa_\G(m_i)) \Big|_{S=0} \Big\} \\ 
    &\hspace{3cm} \times \text{logit}(\psi_0(0, w)) \prod_{L_i \in \mathbb{L}_\rho} dP_0(\ell_i \mid \pa_\G(\ell_i)) \Big|_{S=0} \  dP_0(x)  \bigg\} \ . 
\end{aligned}
\end{equation}
% \begin{align}
%     \Theta_{\Pi, P_0}(\psi_0) 
%     &= \!\! \int \!\! \Big\{ \text{logit}(\psi_0(1, w)) \!\! \prod_{L_i \in \mathbb{L}_\rho \setminus Y} \!\!\! \! dP_0(\ell_i \mid \pa_\G(\ell_i)) \Big|_{S=1} \!\!\!\!\! - \text{logit}(\psi_0(0, w)) \!\! \prod_{L_i \in \mathbb{L}_\rho \setminus Y} \!\!\!\! dP_0(\ell_i \mid \pa_\G(\ell_i)) \Big|_{S=0} \Big\} \notag \\ 
%     &\hspace{4cm} \times \prod_{M_i \in \mathbb{M}_\rho} dP_0(m_i \mid \pa_\G(m_i)) \Big|_{S=0} \  dP_0(x) \ . 
%     \label{eq:pse_id_odds}
% \end{align}%  
}

The canonical gradient of the constraint in \eqref{eq:pse_id_odds_general} is denoted by $D_{\Theta_\Pi, P_0}(\psi)(Z)$ and expressed as: 
    \begin{align}
        D_{\Theta_\Pi, P_0}(\psi)(Z) = D_{\Theta_\Delta, P_0}(Z) \times\frac{1}{\psi(Z) (1 - \psi(Z)) } \ , 
        \label{eq:theta_gradient_odds}
    \end{align}
where $D_{\Theta_\Delta, P_0}(Z)$ is given in \eqref{eq:theta_gradient_general}; see Appendix~\ref{app:proofs_gradients} for a proof.  

Under the cross-entropy risk and $\rho$-specific constraint on a log-odds ratio scale, we are able to find a closed-form solution for optimal constrained parameter $\psi^*_0$. The results are stated in the following theorem. 
\begin{theorem}[Cross-entropy risk and constraint on log-odds ratio scale] 
    \mbox{} \\
    Assuming $\psi^*_0(z) = \argmin_{\psi \in \Psi, \Theta_{\Pi, P_0}(\psi) = 0} P_0 L(\psi)$, with $L(\psi)$ representing the negative log loss and $\Theta_{\Pi, P_0}(\psi)$ specified by \eqref{eq:pse_id_odds_general}, condition~\eqref{eq:pathcondition} and $\Theta_{\Pi, P_0}(\psi^*_0) = 0$ together imply
    \begin{align}
        \psi^*_0(z) = \psi_0(z) - \lambda_0 D_{\Theta_\Delta, P_0}(z) \ , 
        \label{eq:cross_closed-form_odds}
    \end{align}%
    where $D_{\Theta_\Delta, P_0}(Z)$ is given in \eqref{eq:theta_gradient_general}. 
\label{thm:closed-cross-odds}
\end{theorem}

The formulation posited in \eqref{eq:cross_closed-form_odds} shares conceptual parallels with the MSE risk scenario in \eqref{eq:mse_closed-form}, particularly in the aspect of adding a mean-zero random variable to the unconstrained minimizer to effectuate population-level fairness adjustments. As with MSE, the amount added is greatest for data points that significantly influence the constraint as quantified by the value of the gradient at that point. However, in contrast to MSE, a closed-form solution for $\lambda_0$ is not available. Thus, estimation of $\lambda_0$ must instead rely on a grid search.

To estimate $\psi_0^*$ in this setting, we first estimate $\psi_{0, \lambda}$ for a given $\lambda$ via $\psi_{n, \lambda}(z) = \psi_n(z) - \lambda D_{\Theta_\Delta, n}(z)$. We then compute $\psi_{n, \lambda_n}$ as the constrained minimizer where $\lambda_n$ is determined by a grid search as a minimizer to $\lambda_n = \argmin_{ \epsilon_\ell \leq \lambda \leq \epsilon_u} |\Theta_{\Pi, n}(\psi_{n, \lambda})|$. The bounds, for when $s_y=1$, are calculated as follows:
{\small 
\begin{align*}
   \epsilon_\ell &=  \min_{w}\max \left\{ \frac{\psi_0(0, w)}{D_{\Theta_\Delta, P_0}(0, w)}, -\frac{1 - \psi_0(1, w)}{ D_{\Theta_\Delta, P_0}(1, w)}  \right\} \ , \quad \epsilon_u = \max_{w} \min \left\{ \frac{\psi_0(1, w)}{ D_{\Theta_\Delta, P_0}(1, w)}, \frac{\psi_0(0, w)-1}{D_{\Theta_\Delta, P_0}(0, w) }  \right\} \ . 
\end{align*}
}
See Appendix~\ref{app:proofs_closed-cross-odds} for a proof of Theorem~\ref{thm:closed-cross-odds} and detailed derivations of the above bounds.

%######################################
\section{Proofs} 
\label{app:proofs}
%######################################

\subsection{Lemma~\ref{lem:id_constraints}} 

The identification functional is known as the edge g-formula, and the proof is given by \cite{shpitser2016causal}. 

%+++++++++++++++++++++++++++++++++++++++++++++++
\subsection{Lemma~\ref{lemma:gradients}}
\label{app:proofs_gradients}

\textbf{Gradient of the MSE risk.} \  

Consider paths through $\psi \in {\Psi}$ at $\delta = 0$ of the form $\{\psi_{\delta, h}(s, w) = \psi(s,w) + \delta h(s,w): \delta \in \openr\}$, with direction $\frac{d}{d\delta} \psi_{\delta,h} |_{\delta = 0} = h$ in the space of bounded real-valued functions defined on the support of $(S, W)$ implied by $P$, denoted by $L^2(P_{S,W})$. For any direction $h \in L^2(P_{S,W})$, we have: 
\begin{align*}
	\left. \frac{d}{d \delta} R_{P_0}(\psi_{\delta, h}) \right |_{\delta = 0} 
    &= \left. \frac{d}{d \delta}  P_0L(\psi_{\delta, h})(Z) \right |_{\delta = 0} \\
    &= -\E_{P_0}\left[ 2 \{Y - \psi(S,W)\} h(S,W) \right] \\
    &= \E_{P_0} \left[2\{\psi(S,W) - \psi_0(S,W)\}
	h(S,W)\right] \ . 
\end{align*}
Thus, $D_{R, P_0}(\psi)(z) = 2\{\psi(z) - \psi_0(z)\}$.

\textbf{Gradient of the constraint} $\Theta_{\Delta, P_0}(\psi).$  

We first consider the constraint identified via \eqref{eq:pse_id} (i.e., where direct effect is part of $\rho$). Given the same setup as above, for any $h \in L^2(P_{S,W})$, we can write: 
{\small 
\begin{align*}
	&\left. \frac{d}{d \delta} \Theta_{\Delta, P_0}(\psi_{\delta, h}) \right |_{\delta = 0}  \\
    &\hspace{0.5cm} = \displaystyle \int \Big\{ h(1, w) \prod_{L_i \in \mathbb{L}_\rho \setminus Y} dP_0(\ell_i \mid \pa_\G(\ell_i)) \Big|_{S=1} - h(0, w) \prod_{L_i \in \mathbb{L}_\rho \setminus Y} dP_0(\ell_i \mid \pa_\G(\ell_i)) \Big|_{S=0} \Big\} \notag \\ 
        &\hspace{4cm} \times \prod_{M_i \in \mathbb{M}_\rho} dP_0(m_i \mid \pa_\G(m_i)) \Big|_{S=0} \  dP_0(x) \\
	&= \int \left\{ \frac{\mathbb{I}(s = 1)}{P_0(s=1 \mid \pa_\G(s))} -
	\frac{\mathbb{I}(s = 0)}{P_0(s=0 \mid \pa_\G(s))} \right\} \prod_{M_i \in \mathbb{M}_\rho} \frac{P_0(m_i \mid \pa_\G(m_i)\setminus s, S=0) }{P_0(m_i \mid \pa_\G(m_i))} h(s, w) dP_0(s, w) \\
	&= \int \frac{2\I(s=1) - 1}{P_0(s \mid \pa_\G(s))} \prod_{M_i \in \mathbb{M}_\rho} \frac{P_0(m_i \mid \pa_\G(m_i)\setminus s, 0) }{P_0(m_i \mid \pa_\G(m_i))} \ h(s, w) \ dP_{0,S,W}(s, w) \ . 
\end{align*}
}%
Thus, $D_{\Theta_\Delta, P_0}(Z) = \frac{2S - 1}{P_0(S \mid \pa_\G(S))} \ \prod_{M_i \in \mathbb{M}_\rho} \frac{P_0(M_i \mid \pa_\G(M_i) \setminus S, S = 0)}{P_0(M_i \mid \pa_\G(M_i))}$.

To obtain the constraint gradient for the  general case defined in \eqref{eq:pse_id_general}, we can derive the gradient assuming $s_y = 0$: 
{\small 
\begin{align*}
	\left. \frac{d}{d \delta} \Theta_{\Delta, P_0}(\psi_{\delta, h}) \right |_{\delta = 0} 
    &= \displaystyle \int \Big\{  \prod_{M_i \in \mathbb{M}_\rho} dP_0(m_i \mid \pa_\G(m_i)) \Big|_{S=1} - \prod_{M_i \in \mathbb{M}_\rho} dP_0(m_i \mid \pa_\G(m_i)) \Big|_{S=0} \Big\} \notag \\ 
        &\hspace{4cm} \times h(0, w) \prod_{L_i \in \mathbb{L}_\rho \setminus Y} dP_0(\ell_i \mid \pa_\G(\ell_i)) \Big|_{S=0} \  dP_0(x) \\
	&=  \int  \frac{\I(s=0)}{P_0(s \mid \pa_\G(s))} \left\{ \prod_{M_i \in \mathbb{M}_\rho} \frac{P_0(m_i \mid \pa_\G(m_i) \setminus s, S=1)}{P_0(m_i \mid \pa_\G(m_i))} - 1 \right\}  h(s, w) dP_{0, S, W}(s, w) \ . 
\end{align*}
}
Thus, $D_{\Theta_\Delta, P_0}(Z) = \frac{(1-S)}{P_0(S \mid \pa_\G(S))} \left\{ \prod_{M_i \in \mathbb{M}_\rho} \frac{P_0(M_i \mid \pa_\G(M_i) \setminus S, S=1)}{P_0(M_i \mid \pa_\G(M_i))} - 1 \right\}$. 

Putting these derivations together, we obtain the general form of the constraint in \eqref{eq:theta_gradient_general}.

\textbf{Gradient of the cross-entropy risk.} \  

Consider paths through $\psi \in \Psi$ at $\delta=0$ of the form $\big\{ \psi_{\delta, h}(s,w) = \text{expit}\left[ \text{logit}\{\psi(s,w)\} + \delta h(s, w)  \right]: \delta \in \openr \big\},$ with direction $\frac{d}{d\delta} \psi_{\delta,h}|_{\delta = 0} =  h$ in $L^2(P_{S, W})$, the space of real-valued functions defined on the support of $(S,W)$ implied by $P$. For any $h \in L^2(P_{S, W})$, 
\begin{align*}
	\left. \frac{d}{d \delta} R_{P_0}(\psi_{\delta, h}) \right |_{\delta = 0} 
    &= \left. \frac{d}{d \delta}  P_0L(\psi_{\delta, h})(Z) \right |_{\delta = 0} \\ 
    &= \int \left\{\frac{-y}{\psi(s, w)} + \frac{1-y}{1- \psi(s, w)} \right\} h(s,w) \ dP_0(o) \\
    &= \int \left\{\frac{-\psi_0(s, w)}{\psi(s, w)} + \frac{1-\psi_0(s, w)}{1- \psi(s, w)} \right\} h(s,w) \ dP_{0,S,W}(s, w) \ .  
\end{align*}
Thus, $D_{R, P_0}(\psi)(z) = \displaystyle \frac{1-\psi_0(s, w)}{1- \psi(s, w)} - \frac{\psi_0(s, w)}{\psi(s, w)}$.

\textbf{Gradient of the constraint $\Theta_{\Pi, P_0}$.} \ 

Consider the constraint identified via \eqref{eq:pse_id_odds_general}. Given the same setup as above and when $s_y = 1$, for any $h \in L^2(P_{S,W})$, we can write: 
{\small 
\begin{align*}
	&\left. \frac{d}{d \delta} \Theta_{\Pi, P_0}(\psi_{\delta, h}) \right |_{\delta = 0}  \\
    &\hspace{0.0cm} = \displaystyle \int \Big\{ \frac{h(1, w)}{\psi(1, w) (1-\psi(1, w))} \!\! \prod_{L_i \in \mathbb{L}_\rho \setminus Y} dP_0(\ell_i \mid \pa_\G(\ell_i)) \Big|_{S=1} \\ 
    &\hspace{0.5cm}- \frac{h(0, w)}{\psi(0, w)(1-\psi(0, w))} \prod_{L_i \in \mathbb{L}_\rho \setminus Y} dP_0(\ell_i \mid \pa_\G(\ell_i)) \Big|_{S=0} \Big\}  \times \prod_{M_i \in \mathbb{M}_\rho} dP_0(m_i \mid \pa_\G(m_i)) \Big|_{S=0} \  dP_0(x) \\
	&= \int \frac{2\I(s=1) - 1}{P_0(s \mid \pa_\G(s))} \prod_{M_i \in \mathbb{M}_\rho} \frac{P_0(m_i \mid \pa_\G(m_i)\setminus s, S=0) }{P_0(m_i \mid \pa_\G(m_i))} \ \frac{h(s, w)}{\psi(s,w)(1-\psi(s,w))} \ dP_{0, S,W}(s, w) \ . 
\end{align*}
}%
And when $s_y=0$, we can write: 
{\small 
\begin{align*}
	&\left. \frac{d}{d \delta} \Theta_{\Pi, P_0}(\psi_{\delta, h}) \right |_{\delta = 0}  \\
    &\hspace{0.5cm} = \displaystyle \int \Big\{ \prod_{M_i \in \mathbb{M}_\rho} dP_0(m_i \mid \pa_\G(m_i)) \Big|_{S=1} -  \prod_{M_i \in \mathbb{M}_\rho} dP_0(m_i \mid \pa_\G(m_i)) \Big|_{S=0} \Big\} \\ 
    &\hspace{3cm} \times \frac{h(0, w)}{\psi(0, w) (1-\psi(0, w))} \!\! \prod_{L_i \in \mathbb{L}_\rho} dP_0(\ell_i \mid \pa_\G(\ell_i)) \Big|_{S=0} \  dP_0(x) \\
	&\hspace{0.5cm}= \int \frac{\I(s=0)}{P_0(s \mid \pa_\G(s))} \left\{ \prod_{M_i \in \mathbb{M}_\rho} \frac{P_0(m_i \mid \pa_\G(m_i)\setminus s, S=1) }{P_0(m_i \mid \pa_\G(m_i))} - 1 \right\} \ \frac{h(s, w)}{\psi(s,w)(1-\psi(s,w))} \ dP_{0, S,W}(s, w) \ .  
\end{align*}
}

Thus, $D_{\Theta_\Pi, P_0}(\psi)(z) = D_{\Theta_\Delta, P_0}(z)/ \{\psi(z) (1-\psi(z)) \}$, where $D_{\Theta_\Delta, P_0}(z)$ is given in \eqref{eq:theta_gradient_general}.

%+++++++++++++++++++++++++++++++++++++++++++++++
\subsection{Theorem~\ref{thm:closed-mse}}

Given the canonical gradient of the MSE risk and Condition~\ref{eq:pathcondition}, we can write: 
\begin{align*}
    &2 \{ \psi_{0, \lambda_0}(z) - \psi_0(z) \} + \lambda_0 D_{\Theta_\Delta, P_0}(z) = 0 \quad \rightarrow \quad \psi_{0, \lambda_0}(z) = \psi_0(z) - 0.5\lambda_0 D_{\Theta_\Delta, P_0}(z) \ . 
\end{align*}
On the other hand, $\Theta_{\Delta, P_0}(\psi_{0, \lambda_0}) = 0$, where $\Theta_{\Delta, P_0}(\psi)$ is given in \eqref{eq:pse_id}. This implies: 
\begin{align*}
    &\displaystyle \int \Big\{ \psi_{0, \lambda_0}(1, w) \prod_{L_i \in \mathbb{L}_\rho \setminus Y} dP_0(\ell_i \mid \pa_\G(\ell_i)) \Big|_{S=1} - \psi_{0, \lambda_0}(0, w) \prod_{L_i \in \mathbb{L}_\rho \setminus Y} dP_0(\ell_i \mid \pa_\G(\ell_i)) \Big|_{S=0} \Big\} \notag \\ 
    &\hspace{4cm} \times \prod_{M_i \in \mathbb{M}_\rho} dP_0(m_i \mid \pa_\G(m_i)) \Big|_{S=0} \  dP_0(x) = 0 \ . 
\end{align*}
Substituting $\psi_{0, \lambda_0}$ into the above equation and simplifying, we obtain: 
\begin{align*}
   &\Theta_{\Delta, P_0}(\psi_0) - 0.5 \lambda_0 \int D^2_{\Theta_\Delta, P_0}(z) dP_0(z) = 0 \quad \rightarrow \quad  \lambda_0 = 2 \Theta_{\Delta, P_0}(\psi_0)/ \sigma^2(D_{\Theta_\Delta, P_0}) \ . 
\end{align*}
Putting all these together, we obtain: 
$\psi^*_0(z) \equiv \psi_{0, \lambda_0}(z) = \psi_0(z) - \Theta_{\Delta, P_0}(\psi_0) \frac{D_{\Theta_\Delta, P_0}(z)}{\sigma^2(D_{\Theta_\Delta, P_0})}.$ 

The mean squared difference between $\psi^*_0$ and $\psi_0$ is simply
\begin{align*}
    \E[(\psi^*_0(Z) - \psi_0(Z))^2] = \frac{\Theta^2_{\Delta, P_0}(\psi_0)}{\sigma^4(D_{\Theta_\Delta, P_0})} \ \E[D^2_{\Theta_\Delta, P_0}(Z)] = \frac{\Theta^2_{\Delta, P_0}(\psi_0)}{\sigma^2(D_{\Theta_\Delta, P_0})} \ . 
\end{align*}

Under the more general constraint of $\Theta_{\Delta, P_0}(\psi)$ in \eqref{eq:pse_id_general}, we can show that the same result holds: if $s_y=0$, $\Theta_{\Delta, P_0}(\psi_0, \lambda_0)$ = 0 implies: 
\begin{align*}
    &\displaystyle \int \Big\{  \prod_{M_i \in \mathbb{M}_\rho} dP_0(m_i \mid \pa_\G(m_i)) \Big|_{S=1} - \prod_{M_i \in \mathbb{M}_\rho} dP_0(m_i \mid \pa_\G(m_i)) \Big|_{S=0} \Big\} \notag \\ 
    &\hspace{4cm} \times \psi_{0, \lambda_0}(0, w) \prod_{L_i \in \mathbb{L}_\rho \setminus Y} dP_0(\ell_i \mid \pa_\G(\ell_i)) \Big|_{S=0} \  dP_0(x) = 0 \ . 
\end{align*}
Substituting $\psi_{0, \lambda_0}$ into the above equation and simplifying, we obtain: $$\Theta_{\Delta, P_0}(\psi_0) - 0.5 \lambda_0 \int D^2_{\Theta_\Delta, P_0}(z) dP_0(z) = 0, $$ where $D_{\Theta_\Delta, P_0}$ is given by \eqref{eq:theta_gradient_general} under $s_y=0$. This yields the exact same closed-form solution for $\psi^*_0 \equiv \psi_{0, \lambda_0}$ as above. 

%+++++++++++++++++++++++++++++++++++++++++++++++
\subsection{Theorem~\ref{thm:closed-cross}}

Given the canonical gradient of the cross-entropy risk, canonical gradient of the constraint $\Theta_{\Delta, P_0}$, and Condition~\ref{eq:pathcondition}, we can write: 
\begin{align}
    & \frac{\psi_{0, \lambda_0}(z) - \psi_0(z)}{\psi_{0, \lambda_0}(z) (1-\psi_{0, \lambda_0}(z))} + \lambda_0 D_{\Theta_\Delta, P_0}(z) = 0  \label{eq:quadratic_path} \\ 
    \rightarrow \qquad & \psi_{0, \lambda_0}(z) = \psi_0(z) - \lambda_0 D_{\Theta_\Delta, P_0}(z) \sigma^2_{\psi_{0, \lambda_0}}(z) \ , \notag
\end{align}
where $\sigma^2_{\psi_{0, \lambda_0}}(z) = \psi_{0, \lambda_0}(z) (1-\psi_{0, \lambda_0}(z))$. On the other hand, $\Theta_{\Delta, P_0}(\psi_{0, \lambda_0}) = 0$, where $\Theta_{\Delta, P_0}(\psi)$ is given in \eqref{eq:pse_id}. This implies $\lambda_0 = \Theta_{\Delta, P_0}(\psi_0)/ \E[D^2_{\Theta_\Delta, P_0}(Z) \sigma^2_{\psi_{0, \lambda_0}}(Z) ]$. Together, we have that $\psi^*_0$ can be characterized by: 
$$\psi_{0, \lambda_0}(z) = \psi_0(z) - \Theta_{\Delta, P_0}(\psi_0) \frac{D_{\Theta_\Delta, P_0}(z) \sigma^2_{\psi_{0, \lambda_0}}(z)}{\E[D^2_{\Theta_\Delta, P_0}(Z) \sigma^2_{\psi_{0, \lambda_0}}(Z) ]}.$$ 

The mean squared difference between $\psi^*_0$ and $\psi_0$ can be written down as: 
\begin{align*}
    \E[(\psi^*_0(Z) - \psi_0(Z))^2] 
    &= \Theta^2_{\Delta, P_0}(\psi_0)\frac{\E[D^2_{\Theta_\Delta, P_0}(Z) \sigma^4_{\psi_{0, \lambda_0}}(Z)]}{\E^2[D^2_{\Theta_\Delta, P_0}(Z) \sigma^2_{\psi_{0, \lambda_0}}(Z) ]}  \\
    &\leq \frac{\Theta^2_{\Delta, P_0}(\psi_0)}{4 \times \E[D^2_{\Theta_\Delta, P_0}(Z) \sigma^2_{\psi_{0, \lambda_0}}(Z) ]} \ , 
\end{align*}% 
where the last equality holds since $\sigma^2_{\psi_{0, \lambda_0}}(Z) \leq 0.25$, and thus $$\E[D^2_{\Theta_\Delta, P_0}(Z) \sigma^4_{\psi_{0, \lambda_0}}(Z)] \leq \frac{\E[D^2_{\Theta_\Delta, P_0}(Z) \sigma^2_{\psi_{0, \lambda_0}}(Z)]}{4} \ . $$

%+++++++++++++++++++++++++++++++++++++++++++++++
\subsection{Lemma~\ref{lem:closed_quadratic_cross}}
\label{app:quadratic_equation}

According to \eqref{eq:quadratic_path}, we have the following quadratic equation: 
\begin{align*}
    \lambda D_{\Theta_\Delta, P_0}(z)  \psi^2_{0, \lambda}(z) - \left\{1 + \lambda D_{\Theta_\Delta, P_0}(z)  \right\} \psi_{0, \lambda}(z) + \psi_0(z) = 0 \ . 
\end{align*}
The roots are given by the quadratic formula, which simplifies to: 
\begin{align*}
\frac{ \{ \lambda + D_{\Theta_\Delta, P_0}(z)^{-1} \} \pm  \big\{ ( \lambda + D_{\Theta_\Delta, P_0}(z)^{-1})^2 - 4\lambda D_{\Theta_\Delta, P_0}(z)^{-1}\psi_0(z)   \big\}^{1/2}  }{2\lambda} \ . 
\end{align*}
Given $D_{\Theta_\Delta, P_0}$ in Lemma~\ref{lemma:gradients}, the discriminant is non-negative, and it ensures two solutions exist: 
\begin{itemize}
    \item For $\lambda > 0$, we observe: $(\lambda + D_{\Theta_\Delta, P_0}(1, w)^{-1})^2 - 4\lambda D_{\Theta_\Delta, P_0}(1, w)^{-1} \psi_0(s, w) \geq  (\lambda - D_{\Theta_\Delta, P_0}(1, w)^{-1})^2$ and $(\lambda + D_{\Theta_\Delta, P_0}(s_y=0, w)^{-1})^2 - 4\lambda D_{\Theta_\Delta, P_0}(s_y=0, w)^{-1} \psi_0(s, w) \geq 0. $

    \item For $\lambda < 0$, we observe: $(\lambda + D_{\Theta_\Delta, P_0}(1, w)^{-1})^2 - 4\lambda D_{\Theta_\Delta, P_0}(1, w)^{-1} \psi_0(s, w) \geq 0$ and $(\lambda + D_{\Theta_\Delta, P_0}(s_y=0, w)^{-1})^2 - 4\lambda D_{\Theta_\Delta, P_0}(s_y=0, w)^{-1} \psi_0(s, w) \geq (\lambda - D_{\Theta_\Delta, P_0}(s_y=0, w)^{-1})^2. $ 
\end{itemize}
However, as $\psi_{0, \lambda}(z)$ represents a probability, it must be confined to the unit interval. Given this constraint and knowing the above inequalities, it is straightforward to see that the following solution for $\psi_{0, \lambda}(s,w)$ remains within this range: 
\begin{align*}
\psi_{0, \lambda}(s,w) = \frac{ 1 + \lambda D_{\Theta_\Delta, P_0}(s,w) - (2s-1)   \left[ \left\{ 1 + \lambda D_{\Theta_\Delta, P_0}(s,w) \right\}^2 - 4\lambda \psi_0(s,w)D_{\Theta_\Delta, P_0}(s,w) \right]^{1/2}  }{2\lambda D_{\Theta_\Delta, P_0}(s,w)} \ . 
\end{align*}

%+++++++++++++++++++++++++++++++++++++++++++++++
\subsection{Lemma~\ref{lem:theta_1_est}}

It suffices to show the following equations hold:
{\small 
\begin{equation}\label{eq:pse_ex1_proof} 
\begin{aligned}
    \Theta_{\rho_1, P_0}(\psi_0) 
    &= \int \left\{\psi_0(1, w) dP_0(\ell \mid 1, m, x) - \psi_0(0, x) dP_0(\ell \mid 0, m, x) \right\} dP_0(m \mid 0, x) dP_0(x) \\
    &= \E\left[ \frac{2S -1}{\pi_0(S \mid X)} \frac{f_{0,M}(M \mid S=0, X)}{f_{0,M}(M \mid S, X)} Y \right] \\
    &= \E\left[ \frac{\I(S=0)}{\pi_0(0 \mid X)} \frac{f_{0,L}(L \mid S=1, M, X)}{f_{0,L}(L \mid S, M, X)} \psi_0(1, W)  - \frac{\I(S=0)}{\pi_0(0 \mid X)}  \psi_0(0, W) \right] \ . 
\end{aligned}
\end{equation}
}

The first equality holds by Lemma~\ref{lem:id_constraints}. The second equality holds since: 
{\small 
\begin{align*}
    &\E\left[ \frac{2S -1}{\pi_0(S \mid X)} \frac{f_{0,M}(M \mid S=0, X)}{f_{0,M}(M \mid S, X)} Y \right] \\
    &\hspace{0.5cm} = \int \frac{2s -1}{\pi_0(s \mid x)} \frac{f_{0,M}(m \mid S=0, x)}{f_{0,M}(m \mid s, x)} \psi_0(s, \ell, m, x)  f_{0, L}(\ell | s, m, x) f_{0, M}(m | s, x) \pi_0(s | x) p_{0, X}(x) d\ell \ dm \ ds \ dx 
    \\
    &\hspace{0.5cm} = \int (2s -1)  \psi_0(s, \ell, m, x)  f_{0, L}(\ell \mid s, m, x) f_{0,M}(m \mid S=0, x) p_{0, X}(x) \ d\ell \ dm \ ds \ dx 
    \\
    &\hspace{0.5cm} =  \Theta_{\rho_1, P_0}(\psi_0)  \ . 
\end{align*}
}

The last equality in \eqref{eq:pse_ex1_proof} holds since: 
{\small 
\begin{align*}
    &\E\left[ \frac{\I(S=0)}{\pi_0(0 \mid X)} \frac{f_{0,L}(L \mid S=1, M, X)}{f_{0,L}(L \mid S, M, X)} \psi_0(1, W)  - \frac{\I(S=0)}{\pi_0(0 \mid X)}  \psi_0(0, W) \right] 
    \\
    &\hspace{0.5cm}= \int \{ \psi_0(1, w) f_{0,L}(\ell \mid S=1, m, x) -  \psi_0(0, w) f_{0,L}(\ell \mid S=0, m, x) \} f_{0, M}(m \mid S=0, m, x) p_{0, X}(x) d\ell \ dm \ dx \ , 
    \\
    &\hspace{0.5cm} =  \Theta_{\rho_1, P_0}(\psi_0)  \ . 
\end{align*}
}

Lastly, $\Theta^\text{aipw}_{\rho_1, n}$ is based on the efficient influence function (EIF) for $\Theta_{\rho_1, P_0}(\psi_0)$, which can be derived as follows. Let $P_{0,\varepsilon}$ denote a submodel through $P_0$ at $\varepsilon = 0$ indexed by score function $\mathcal{S}_0$. To derive the efficient influence function, we can compute $\frac{\partial}{\partial \varepsilon} \Theta_{\rho_1, P_{0, \epsilon}}|_{\varepsilon=0}$ and represent the resultant derivative as an inner product in $L^2_0(P_0)$.
{\small 
\begin{align*}
\frac{\partial}{\partial \varepsilon} &\Theta_{\rho_1, P_{0, \epsilon}} \Big|_{\varepsilon=0} 
=\frac{\partial}{\partial \varepsilon} \int y \ d P_{0, \varepsilon}\left(y \mid S=1, \ell, m, x\right) d P_{0, \varepsilon}\left(\ell \mid S=1, m, x\right) dP_{0, \varepsilon}(m \mid S=0, x)  d P_{0, \varepsilon}(x)\Big|_{\varepsilon=0}  \\ 
&\hspace{3cm} - \frac{\partial}{\partial \varepsilon} \int y \ d P_{0, \varepsilon}\left(y, \ell, m \mid S=0, x\right)  d P_{0, \varepsilon}(x)\Big|_{\varepsilon=0}  
\\
& =\int y \ \mathcal{S}_0\left(y \mid S=1, \ell, m, x\right) d P_0\left(y \mid S=1, \ell, m, x\right) dP_0\left(\ell \mid S=1, m, x\right) dP_0(m \mid S=0, x)  d P_0(x)  \quad(1) \\
&\hspace{0.2cm} + \int y \ \mathcal{S}_0\left(\ell \mid S=1, m, x\right) d P_0\left(y \mid S=1, \ell, m, x\right) dP_0\left(\ell \mid S=1, m, x\right) dP_0(m \mid S=0, x)  d P_0(x)  \quad(2) \\
&\hspace{0.2cm} + \int y \  \mathcal{S}_0\left(m \mid S=0, x\right) d P_0\left(y \mid S=1, \ell, m, x\right) dP_0\left(\ell \mid S=1, m, x\right) dP_0(m \mid S=0, x)  d P_0(x)  \quad(3) \\
&\hspace{0.2cm} - \int y  \ \mathcal{S}_0\left(y, \ell, m \mid S=0, x\right) d P_0\left(y, \ell, m \mid S=0, x\right)  d P_0(x)  \quad(4) \\ 
&\hspace{0.2cm} + \int y \ \mathcal{S}_0\left(x\right) \Big\{ d P_0\left(y \mid S=1, \ell, m, x\right) dP_0\left(\ell \mid S=1, m, x\right) \\ 
&\hspace{2.5cm} - d P_0\left(y \mid S=0, \ell, m, x\right) dP_0\left(\ell \mid S=0, m, x\right) \Big\} dP_0(m \mid S=0, x)  d P_0(x)  \quad(5) \\
\end{align*}
}%
where $\mathcal{S}_0(. \mid .)$ denotes a conditional score function w.r.t. $P_0(. \mid .)$. 

With the shorthand notation of the nuisances, Line (1) simplifies to:
{\small 
\begin{align*}
    &\int y \ \mathcal{S}_0\left(y \mid S=1, \ell, m, x\right) d P_0\left(y \mid S=1, \ell, m, x\right) dP_0\left(\ell \mid S=1, m, x\right) dP_0(m \mid S=0, x)  d P_0(x)  \\
    &\hspace{0.75cm}= \int \frac{\I(S=1)}{\pi_0(s \mid x)} \frac{f_{0, M}(m \mid S=0, x)}{f_{0, M}(m \mid s, x)} y  \ \mathcal{S}_0\left(y \mid s, \ell, m, x\right) d P_0\left(y, \ell, m, s, x\right)  \\
    &\hspace{0.75cm}= \int \frac{\I(S=1)}{\pi_0(s \mid x)} \frac{f_{0, M}(m \mid S=0, x)}{f_{0, M}(m \mid s, x)} \ \left(y - \psi_0(s, w)\right)  \ \mathcal{S}_0\left(y \mid s, \ell, m, x\right) d P_0\left(y, \ell, m, s, x\right)  \\
    &\hspace{0.75cm}= \int \frac{\I(S=1)}{\pi_0(s \mid x)} \frac{f_{0, M}(m \mid S=0, x)}{f_{0, M}(m \mid s, x)} \ \left(y - \psi_0(s, w)\right)  \ \mathcal{S}_0\left(y, \ell, m, s, x\right) d P_0\left(y, \ell, m, s, x\right)  \\
    &\hspace{0.75cm}= \E\left[ \frac{\I(S=1)}{\pi_0(S \mid X)} \frac{f_{0, M}(M \mid S=0, X)}{f_{0, M}(M \mid S, X)} \ \left(Y - \psi_0(1, W)\right)  \ \mathcal{S}_0\left(Y, L, M, S, X\right) \right]  \ . 
\end{align*}
}

Line (2) simplifies to:
{\small 
\begin{align*}
     &\int y \ \mathcal{S}_0\left(\ell \mid S=1, m, x\right) d P_0\left(y \mid S=1, \ell, m, x\right) dP_0\left(\ell \mid S=1, m, x\right) dP_0(m \mid S=0, x)  d P_0(x)  \\
     &\hspace{0.75cm}= \int \frac{\I(S=1)}{\pi_0(s \mid x)} \frac{f_{0, M}(m \mid S=0, x)}{f_{0, M}(m \mid s, x)} \psi_0(1, w) \ \mathcal{S}_0\left(\ell \mid s, m, x\right)  dP_0\left(\ell, m, s, x\right)  \\ 
     &\hspace{0.75cm}= \int \frac{\I(S=1)}{\pi_0(s \mid x)} \frac{f_{0, M}(m \mid S=0, x)}{f_{0, M}(m \mid s, x)} \ \left(\psi_0(1, w) - \bar{\psi}_{0, L, 1}(m, x) \right) \ \mathcal{S}_0\left(\ell \mid s, m, x\right)  dP_0\left(\ell, m, s, x\right) \\ 
    &\hspace{0.75cm}= \int \frac{\I(S=1)}{\pi_0(s \mid x)} \frac{f_{0, M}(m \mid S=0, x)}{f_{0, M}(m \mid s, x)} \ \left(\psi_0(1, w) - \bar{\psi}_{0, L, 1}(m, x)  \right) \ \mathcal{S}_0\left(\ell, m, s, x\right)  dP_0\left(\ell, m, s, x\right) \\ 
    &\hspace{0.75cm}= \E\left[ \frac{\I(S=1)}{\pi_0(S \mid X)} \frac{f_{0, M}(M \mid S=0, X)}{f_{0, M}(M \mid S, X)} \ \left(\psi_0(1, W) - \bar{\psi}_{0, L, 1}(M, X) \right) \ \mathcal{S}_0\left(L, M, S, X\right) \right]  \ . 
\end{align*}
}%
where $\bar{\psi}_{0, L, 1}(M, X) = \int \psi_0(S=1, \ell, M, X) dP_0(\ell \mid S=1, M, X)$. \\

Line (3) simplifies to:
{\small 
\begin{align*}
    &\int y \  \mathcal{S}_0\left(m \mid S=0, x\right) d P_0\left(y \mid S=1, \ell, m, x\right) dP_0\left(\ell \mid S=1, m, x\right) dP_0(m \mid S=0, x)  d P_0(x) \\
    &\hspace{0.75cm}= \int \frac{\I(S=0)}{\pi_0(s \mid x)} \bar{\psi}_{0, L, 1}(m, x) \ \mathcal{S}_0\left(m \mid s, x\right) \ dP_0(m, s, x) \\ 
    &\hspace{0.75cm}= \int \frac{\I(S=0)}{\pi_0(s \mid x)}  \ \left( \bar{\psi}_{0, L, 1}(m, x) - \theta_{0, 1}(x) \right) \ \mathcal{S}_0\left(m \mid s, x\right) \ dP_0(m, s, x) \\ 
    &\hspace{0.75cm}= \int \frac{\I(S=0)}{\pi_0(s \mid x)}  \ \left( \bar{\psi}_{0, L, 1}(m, x) - \theta_{0, 1}(x) \right) \ \mathcal{S}_0\left(m,  s, x\right) \ dP_0(m, s, x) \\ 
    &\hspace{0.75cm}= \E\left[ \frac{\I(S=0)}{\pi_0(S \mid X)}  \ \left( \bar{\psi}_{0, L, 1}(M, X) - \theta_{0, 1}(X) \right) \ \mathcal{S}_0\left(M, S, X\right) \right] \ . 
\end{align*}
}% 
where $\theta_{0, 1}(X) = \int  \bar{\psi}_{0, L, 1}(m, X) dP_0(m \mid S=0, X).$ \\ 

Line (4) simplifies to:
{\small 
\begin{align*}
    &- \int y  \ \mathcal{S}_0\left(y, \ell, m \mid S=0, x\right) d P_0\left(y, \ell, m \mid S=0, x\right)  d P_0(x) \\
    &\hspace{0.75cm}= - \int \frac{\I(S=0)}{\pi_0(s \mid x)} y \ \mathcal{S}_0\left(y, \ell, m \mid s, x\right) d P_0\left(y, \ell, m, s, x\right) \\
    &\hspace{0.75cm}= - \int \frac{\I(S=0)}{\pi_0(s \mid x)} \ \left(y -  \theta_{0, 0}(x) \right) \ \mathcal{S}_0\left(y, \ell, m \mid s, x\right) d P_0\left(y, \ell, m, s, x\right) \\
     &\hspace{0.75cm}= - \int \frac{\I(S=0)}{\pi_0(s \mid x)} \ \left(y -  \theta_{0, 0}(x) \right) \ \mathcal{S}_0\left(y, \ell, m, s, x\right) d P_0\left(y, \ell, m, s, x\right) \\
      &\hspace{0.75cm}= \E\left[ - \frac{\I(S=0)}{\pi_0(S \mid X)} \ \left(Y -  \theta_{0, 0}(X) \right) \ \mathcal{S}_0\left(Y, S, X\right) \right] \ . 
\end{align*}
}%
where $\theta_{0, 0}(X) = \int y dP_0(y \mid S=0, \ell, m, X) dP_0(\ell \mid S=0, m, X) dP_0(m \mid S=0, X).$ \\ 

Line (5) simplifies to:
{\small 
\begin{align*}
    &\int y \ \mathcal{S}_0\left(x\right) \Big\{ d P_0\left(y \mid S=1, \ell, m, x\right) dP_0\left(\ell \mid S=1, m, x\right) \\ 
        &\hspace{2.5cm} - d P_0\left(y \mid S=0, \ell, m, x\right) dP_0\left(\ell \mid S=0, m, x\right) \Big\} dP(m \mid S=0, x)  d P_0(x)  \\
    &\hspace{0.75cm}= \int \left( \theta_{0, 1}(x) - \theta_{0, 0}(x) \right) \ \mathcal{S}_0\left(x\right) dP_0(x) \\ 
    &\hspace{0.75cm}= \int \left( \theta_{0, 1}(x) - \theta_{0, 0}(x) \right) \ \mathcal{S}_0\left(x\right) dP_0(x) \\ 
    &\hspace{0.75cm}= \int \left\{\left( \theta_{0, 1}(x) - \theta_{0, 0}(x) \right) - \Theta_{\rho_1, P_0}(\psi_0) \right\}  \ \mathcal{S}_0\left(x\right) dP_0(x) \\ 
    &\hspace{0.75cm}= \E\left[ \left\{ \theta_{0, 1}(X) - \theta_{0, 0}(X) \right\} - \Theta_{\rho_1, P_0}(\psi_0)  \ \mathcal{S}_0\left(X\right) \right] \ . 
\end{align*}
}

The EIF, denoted by $\Phi(\Theta_{\rho_1, P_0})$ is: 
{\small 
\begin{align*}
    \Phi(\Theta_{\rho_1, P_0})(O)
    &= 
    \underbrace{\frac{I(S = 1)}{\pi_0(1 \mid X)} \frac{f_{0,M}(M \mid S=0, X)}{f_{0,M}(M \mid S, X)} \{Y - \bar{\psi}_{0,L,1}(M,X) \} }_{\Phi_1(\Theta_{\rho_1, P_0})(O)} \\
    &\hspace{2em} \underbrace{ + \frac{I(S = 0)}{\pi_0(0 \mid X)} \{\bar{\psi}_{0,L,1}(M,X) - \theta_{0,1}(X)\} }_{\Phi_2(\Theta_{\rho_1, P_0})(O)}   \\
    &\hspace{2em} \underbrace{ - \frac{I(S = 0)}{\pi_0(0 \mid X)} \{Y - \theta_{0,0}(X) \} }_{\Phi_3(\Theta_{\rho_1, P_0})(O)}  \\
    &\hspace{2em} \underbrace{ + \theta_{0,1}(X) - \theta_{0, 0}(X) - \Theta_{\rho_1, P_0}(\psi_0) }_{\Phi_4(\Theta_{\rho_1, P_0})(O)}  \ .  
\end{align*}
}% 

To prove the triple robustness property, it suffices to show: $\int \Phi(\Theta_{\rho_1, \tilde{Q}})(o) dP_0(o) = 0$, under proper choices of $\tilde{Q}$ as shown below:  

Assume $\tilde{Q} = \{\pi_0, {f}_{0, M}, \tilde{f}_{L}, \tilde{\psi} \}$. Then,   
{\small 
\begin{align*}
     \E[\Phi_1(\Theta_{\rho_1, \tilde{Q}})(O) + \Phi_3(\Theta_{\rho_1, \tilde{Q}})(O) + \Phi_4(\Theta_{\rho_1, \tilde{Q}})(O)] = 0, 
    \quad \E[\Phi_2(\Theta_{\rho_1, \tilde{Q}})(O)] = 0. 
\end{align*}
}

Assume $\tilde{Q} = \{\pi_0, \tilde{f}_{M}, {f}_{0, L}, {\psi}_0 \}$. Then,  
{\small 
\begin{align*}
     \E[\Phi_1(\Theta_{\rho_1, \tilde{Q}})(O)] = 0, 
    \quad \E[\Phi_2(\Theta_{\rho_1, \tilde{Q}})(O) + \Phi_3(\Theta_{\rho_1, \tilde{Q}})(O) + \Phi_4(\Theta_{\rho_1, \tilde{Q}})(O)] = 0. 
\end{align*}
}

Assume $\tilde{Q} = \{\tilde{\pi}, {f}_{0, M}, {f}_{0, L}, {\psi}_0 \}$. Then, 
{\small 
\begin{align*}
     \E[\Phi_1(\Theta_{\rho_1, \tilde{Q}})(O)] = 0, 
    \quad \E[\Phi_2(\Theta_{\rho_1, \tilde{Q}})(O)] = 0, 
    \quad \E[\Phi_3(\Theta_{\rho_1, \tilde{Q}})(O)] = 0, 
    \quad \E[\Phi_4(\Theta_{\rho_1, \tilde{Q}})(O)] = 0. 
\end{align*}
}

%+++++++++++++++++++++++++++++++++++++++++++++++
\subsection{Lemma~\ref{lem:theta_2_est}}

It suffices to show the following equations hold: 

\begin{equation}\label{eq:pse_ex2_proof} 
\begin{aligned}
    \Theta_{\rho_2, P_0}(\psi_0) 
    &= \int \left\{\psi_0(1, w) dP_0(m \mid 1, x) - \psi_0(0, x) dP_0(m \mid 0, x) \right\} dP_0(\ell \mid 0, m, x) dP_0(x) \\
    &= \E\left[ \frac{2S -1}{\pi_0(S \mid X)} \frac{f_{0,L}(L \mid S=0, M, X)}{f_{0,L}(L \mid S, M, X)} Y \right] \\
    &= \E\left[ \frac{\I(S=0)}{\pi_0(0 \mid X)} \frac{f_{0,M}(M \mid S=1, X)}{f_{0,M}(M \mid S, X)} \psi_0(1, W) - \frac{\I(S=0)}{\pi_0(0 \mid X)} \psi_0(0, W)  \right] \ . 
\end{aligned}
\end{equation}

The first equality holds by Lemma~\ref{lem:id_constraints}. The second equality holds since: 
{\small 
\begin{align*}
    &\E\left[ \frac{2S -1}{\pi_0(S \mid X)} \frac{f_{0,L}(L \mid S=0, M, X)}{f_{0,L}(L \mid S, M, X)} Y \right] \\
    &\hspace{0.5cm}= \int \frac{2s -1}{\pi_0(s \mid x)} \psi_0(s, w) dP(\ell \mid S=0, m, x) dP(m \mid s, x) dP(s, x)   \\
    &\hspace{0.5cm}= \int \left\{ \psi_0(1, w) dP(m \mid S=1, x) - \psi_0(0, w) dP(m \mid S=0, x) \right\} dP(\ell \mid S=0, m, x)  dP(x)   \\
    &\hspace{0.5cm}=  \Theta_{\rho_2, P_0}(\psi_0) \ . 
\end{align*}
}

The last equality in \eqref{eq:pse_ex2_proof} holds since: 
{\small 
\begin{align*}
    &\E\left[ \frac{\I(S=0)}{\pi_0(0 \mid X)} \frac{f_{0,M}(M \mid S=1, X)}{f_{0,M}(M \mid S, X)} \psi_0(1, W) - \frac{\I(S=0)}{\pi_0(0 \mid X)} \psi_0(0, W)  \right] \\
    &\hspace{0.5cm}= \int \frac{\I(S=0)}{\pi_0(s \mid x)}  \left\{  \frac{f_{0,M}(m \mid S=1, x)}{f_{0,M}(m \mid s, x)} \psi_0(1, w) - \psi_0(0, w) \right\} dP(\ell \mid s, m, x) dP(m \mid s, x) dP(s, x)    \\ 
    &\hspace{0.5cm}= \int \left\{  \psi_0(1, w) dP(m \mid S=1, x) - \psi_0(0, w) dP(m \mid S=0, x) \right\} dP(\ell \mid S=0, m, x) dP(x)    \\ 
    &\hspace{0.5cm}=  \Theta_{\rho_2, P_0}(\psi_0) \ . 
\end{align*}
}

Lastly, $\Theta^\text{aipw}_{\rho_2, n}$ is based on the efficient influence function (EIF) for $\Theta_{\rho_2, P_0}(\psi_0)$, which can be derived as follows: 
{\small 
\begin{align*}
    \frac{\partial}{\partial \varepsilon} &\Theta_{\rho_2, P_{0, \epsilon}} \Big|_{\varepsilon=0} 
    =\frac{\partial}{\partial \varepsilon} \int y \ d P_{0, \varepsilon}\left(y \mid S=1, \ell, m, x\right) d P_{0, \varepsilon}\left(\ell \mid S=0, m, x\right) dP_{0, \varepsilon}(m \mid S=1, x)  d P_{0, \varepsilon}(x)\Big|_{\varepsilon=0}  \\ 
    &\hspace{3cm} - \frac{\partial}{\partial \varepsilon} \int y \ d P_{0, \varepsilon}\left(y, \ell, m \mid S=0, x\right)  d P_{0, \varepsilon}(x)\Big|_{\varepsilon=0}  
    \\
    & =\int y \ \mathcal{S}_0\left(y \mid S=1, \ell, m, x\right) d P_0\left(y \mid S=1, \ell, m, x\right) dP_0\left(\ell \mid S=0, m, x\right) dP_0(m \mid S=1, x)  d P_0(x)  \quad(1) \\
    &\hspace{0.2cm} + \int y \ \mathcal{S}_0\left(\ell \mid S=0, m, x\right) d P_0\left(y \mid S=1, \ell, m, x\right) dP_0\left(\ell \mid S=0, m, x\right) dP_0(m \mid S=1, x)  d P_0(x)  \quad(2) \\
    &\hspace{0.2cm} + \int y \  \mathcal{S}_0\left(m \mid S=1, x\right) d P_0\left(y \mid S=1, \ell, m, x\right) dP_0\left(\ell \mid S=0, m, x\right) dP_0(m \mid S=1, x)  d P_0(x)  \quad(3) \\
    &\hspace{0.2cm} - \int y  \ \mathcal{S}_0\left(y, \ell, m \mid S=0, x\right) d P_0\left(y, \ell, m \mid S=0, x\right)  d P_0(x)  \quad(4) \\ 
    &\hspace{0.2cm} + \int y \ \mathcal{S}_0\left(x\right) \Big\{ \ d P_0\left(y \mid S=1, \ell, m, x\right) dP_0\left(m \mid S=1, x\right) \\ 
    &\hspace{2.5cm} - d P_0\left(y \mid S=0, \ell, m, x\right) dP_0\left(m \mid S=0, x\right) \Big\} dP_0(\ell \mid S=0, m, x)  d P_0(x)  \quad(5) \\
\end{align*}
}%
where $\mathcal{S}_0(. \mid .)$ denotes a conditional score function w.r.t. $P_0(. \mid .)$. 

With the shorthand notation of the nuisances, Line (1) simplifies to:
{\small 
\begin{align*}
    &\int y \ \mathcal{S}_0\left(y \mid S=1, \ell, m, x\right) d P_0\left(y \mid S=1, \ell, m, x\right) dP_0\left(\ell \mid S=0, m, x\right) dP_0(m \mid S=1, x)  d P_0(x) \\
    &\hspace{0.75cm}= \int \frac{\I(S=1)}{\pi_0(s \mid x)} \frac{f_{0, L}(\ell \mid S=0, m, x)}{f_{0, L}(\ell \mid s, m, x)} y \ \mathcal{S}_0\left(y \mid s, \ell, m, x\right) d P_0\left(y,  \ell, m, s, x\right) \\
    &\hspace{0.75cm}= \int \frac{\I(S=1)}{\pi_0(s \mid x)} \frac{f_{0, L}(\ell \mid S=0, m, x)}{f_{0, L}(\ell \mid s, m, x)} \left\{ y - \psi_0(s, w) \right\}  \mathcal{S}_0\left(y \mid s, \ell, m, x\right) d P_0\left(y,  \ell, m, s, x\right) \\
    &\hspace{0.75cm}= \int \frac{\I(S=1)}{\pi_0(s \mid x)} \frac{f_{0, L}(\ell \mid S=0, m, x)}{f_{0, L}(\ell \mid s, m, x)} \left\{ y - \psi_0(s, w) \right\} \mathcal{S}_0\left(y, \ell, m, s, x\right) d P_0\left(y,  \ell, m, s, x\right) \\
    &\hspace{0.75cm}= \E\left[ \frac{\I(S=1)}{\pi_0(S \mid X)} \frac{f_{0, L}(L \mid S=0, M, X)}{f_{0, L}(L \mid S, M, X)} \left\{ Y - \psi_0(1, W) \right\} \ \mathcal{S}_0\left(y, \ell, m, s, x\right)  \right] \ . 
\end{align*}
}

Line (2) simplifies to:
{\small 
\begin{align*}
    &\int y \ \mathcal{S}_0\left(\ell \mid S=0, m, x\right) d P_0\left(y \mid S=1, \ell, m, x\right) dP_0\left(\ell \mid S=0, m, x\right) dP_0(m \mid S=1, x)  d P_0(x) \\
    &\hspace{0.75cm}= \int \frac{\I(S=0)}{\pi_0(s \mid x)} \frac{f_{0, M}(m \mid S=1, x)}{f_{0, M}(m \mid s, x)} \psi_0(1, w) \ \mathcal{S}_0\left(\ell \mid s, m, x\right) d P_0\left(\ell, m, s, x\right) \\
    &\hspace{0.75cm}=\int \frac{\I(S=0)}{\pi_0(s \mid x)} \frac{f_{0, M}(m \mid S=1, x)}{f_{0, M}(m \mid s, x)} \left\{ \psi_0(1, w) - \bar{\psi}_{0, L, 0}(m, x) \right\} \ \mathcal{S}_0\left(\ell \mid s, m, x\right) d P_0\left(\ell, m, s, x\right) \\
    &\hspace{0.75cm}=\int \frac{\I(S=0)}{\pi_0(s \mid x)} \frac{f_{0, M}(m \mid S=1, x)}{f_{0, M}(m \mid s, x)} \left\{ \psi_0(1, w) - \bar{\psi}_{0, L, 0}(m, x) \right\} \ \mathcal{S}_0\left(\ell, m, s, x\right) d P_0\left(\ell, m, s, x\right) \\
    &\hspace{0.75cm}=\E\left[ \frac{\I(S=0)}{\pi_0(S \mid X)} \frac{f_{0, M}(M \mid S=1, X)}{f_{0, M}(M \mid S, X)} \left\{ \psi_0(1, W) - \bar{\psi}_{0, L, 0}(M, X) \right\} \ \mathcal{S}_0\left(L, M, S, X\right)  \right]  \ , 
\end{align*}
}
where $\bar{\psi}_{0, L, 0}(M, X) = \int \psi_0(S=1, \ell, M, X) dP_0(\ell \mid S=0, M, X)$. \\

Line (3) simplifies to:
{\small 
\begin{align*}
    &\int y \  \mathcal{S}_0\left(m \mid S=1, x\right) d P_0\left(y \mid S=1, \ell, m, x\right) dP_0\left(\ell \mid S=0, m, x\right) dP_0(m \mid S=1, x)  d P_0(x) \\
    &\hspace{0.75cm}= \int \frac{\I(S=1)}{\pi_0(s \mid x)} \bar{\psi}_{0, L, 0}(m, x) \  \mathcal{S}\left(m \mid s, x\right) dP_0(m, s, x) \\
    &\hspace{0.75cm}= \int \frac{\I(S=1)}{\pi_0(s \mid x)} \left\{ \bar{\psi}_{0, L, 0}(m, x) - \tilde{\theta}_{0, 1}(x) \right\} \  \mathcal{S}_0\left(m \mid s, x\right) dP_0(m, s, x) \\
    &\hspace{0.75cm}= \int \frac{\I(S=1)}{\pi_0(s \mid x)} \left\{ \bar{\psi}_{0, L, 0}(m, x) - \tilde{\theta}_{0, 1}(x) \right\} \  \mathcal{S}_0\left(m, \mid s, x\right) dP_0(m, s, x) \\
    &\hspace{0.75cm}= \E\left[ \frac{\I(S=1)}{\pi_0(S \mid X)} \left\{ \bar{\psi}_{0, L, 0}(M, X) - \tilde{\theta}_{0, 1}(X) \right\} \  \mathcal{S}_0\left(M, S, X\right) \right] \ , 
\end{align*}
}%
where $\tilde{\theta}_{0, 1}(X) = \int \bar{\psi}_{0, L, 0}(m, X) dP_0(m \mid S=1, X)$. 

Line (4) simplifies to:
{\small 
\begin{align*}
    &- \int y  \ \mathcal{S}_0\left(y, \ell, m \mid S=0, x\right) d P_0\left(y, \ell, m \mid S=0, x\right)  d P(x) \\
    &\hspace{0.75cm}= - \int \frac{\I(S=0)}{\pi_0(s \mid x)} y \ \mathcal{S}_0\left(y, \ell, m \mid s, x\right) d P_0\left(y, \ell, m, s, x\right) \\
    &\hspace{0.75cm}= - \int \frac{\I(S=0)}{\pi_0(s \mid x)} \ \left(y -  \tilde{\theta}_{0, 0}(x) \right) \ \mathcal{S}_0\left(y, \ell, m \mid s, x\right) d P_0\left(y, \ell, m, s, x\right) \\
     &\hspace{0.75cm}= - \int \frac{\I(S=0)}{\pi_0(s \mid x)} \ \left(y -  \tilde{\theta}_{0, 0}(x) \right) \ \mathcal{S}_0\left(y, \ell, m, s, x\right) d P_0\left(y, \ell, m, s, x\right) \\
      &\hspace{0.75cm}= \E\left[ - \frac{\I(S=0)}{\pi_0(S \mid X)} \ \left(Y -  \tilde{\theta}_{0, 0}(X) \right) \ \mathcal{S}_0\left(Y, S, X\right) \right] \ . 
\end{align*}
}%
where $\tilde{\theta}_{0, 0}(X) = \int y dP_0(y \mid S=0, \ell, m, X) dP_0(\ell \mid S=0, m, X) dP_0(m \mid S=0, X).$ \\ 

Line (5) simplifies to:
{\small 
\begin{align*}
    &\int y \ \mathcal{S}_0\left(x\right) \Big\{ \ d P_0\left(y \mid S=1, \ell, m, x\right) dP_0\left(m \mid S=1, x\right) \\ 
        &\hspace{2.5cm} - d P_0\left(y \mid S=0, \ell, m, x\right) dP_0\left(m \mid S=0, x\right) \Big\} dP_0(\ell \mid S=0, m, x)  d P_0(x)  \\
    &\hspace{0.75cm}= \int \left\{ \tilde{\theta}_{0, 1}(x) - \tilde{\theta}_{0, 0}(x) \right\} \ \mathcal{S}_0\left(x\right) d P_0(x)  \\
     &\hspace{0.75cm}= \int \left\{ \tilde{\theta}_{0, 1}(x) - \tilde{\theta}_{0, 0}(x) - \Theta_{\rho_2, P_0}(\psi_0) \right\} \ \mathcal{S}_0\left(x\right) d P_0(x)  \\
     &\hspace{0.75cm}= \E\left[ \left\{ \tilde{\theta}_{0, 1}(X) - \tilde{\theta}_{0, 0}(X) - \Theta_{\rho_2, P_0}(\psi_0) \right\} \ \mathcal{S}_0\left(X\right) \right] \ . 
\end{align*}
}

The EIF, denoted by $\Phi(\Theta_{\rho_2, P_0})$, is: 
{\small 
\begin{align*}
    \Phi(\Theta_{\rho_2, P_0})(O) 
    &=  \underbrace{\frac{I(S = 1)}{\pi_0(1 \mid X)} \frac{f_{0,L}(L \mid S=0, M, X)}{f_{0,L}(L \mid S, M, X)} \{ Y - \psi_0(1, W) \} }_{ \Phi_1(\Theta_{\rho_2, P_0})(O) }   \\
    &\hspace{2em} \underbrace{ + \frac{I(S = 0)}{\pi_0(0 \mid X)} \frac{f_{0,M}(M \mid S = 1, X)}{f_{0,M}(M \mid S, X)} \{\psi_0(1, W) - \bar{\psi}_{0,L,0}(M, X) \} }_{ \Phi_2(\Theta_{\rho_2, P_0})(O) }  \\
    &\hspace{2em}  \underbrace{ + \frac{I(S = 1)}{\pi_0(1 \mid X)} 
    \{ \bar{\psi}_{0,L,0}(M, X) - \tilde{\theta}_{0,1}(X) \}  }_{ \Phi_3(\Theta_{\rho_2, P_0})(O) }  \\
    &\hspace{2em} \underbrace{ - \frac{I(S = 0)}{\pi_0(0 \mid X)} \{Y - \tilde{\theta}_{0,0}(X) \}  }_{ \Phi_4(\Theta_{\rho_2, P_0})(O) }  \\
    &\hspace{2em}  \underbrace{ + \tilde{\theta}_{0,1}(X)  - \tilde{\theta}_{0,0}(X) - \Theta_{\rho_2, P_0}(\psi_0) }_{ \Phi_5(\Theta_{\rho_2, P_0})(O) }  \ . 
\end{align*}
}%

To prove the triple robustness property, it suffices to show: $\int \Phi(\Theta_{\rho_2, \tilde{Q}})(o) dP_0(o) = 0$, under proper choices of $\tilde{Q}$ as shown below:  

Assume $\tilde{Q} = \{\pi_0, {f}_{0, L}, \tilde{f}_{M}, \tilde{\psi} \}$. Then,   
{\small 
\begin{align*}
     \E[\Phi_1(\Theta_{\rho_2, \tilde{Q}})(O) + \Phi_3(\Theta_{\rho_2, \tilde{Q}})(O) + \Phi_4(\Theta_{\rho_2, \tilde{Q}})(O) + \Phi_5(\Theta_{\rho_2, \tilde{Q}})(O)] = 0, 
    \quad \E[\Phi_2(\Theta_{\rho_2, \tilde{Q}})(O)] = 0. 
\end{align*}
}

Assume $\tilde{Q} = \{\pi_0, \tilde{f}_{L}, {f}_{0, M}, {\psi}_0 \}$. Then,  
{\small 
\begin{align*}
     \E[\Phi_1(\Theta_{\rho_2, \tilde{Q}})(O)] = 0, 
    \quad \E[\Phi_2(\Theta_{\rho_2, \tilde{Q}})(O) + \Phi_3(\Theta_{\rho_2, \tilde{Q}})(O) + \Phi_4(\Theta_{\rho_2, \tilde{Q}})(O) + \Phi_5(\Theta_{\rho_2, \tilde{Q}})(O)] = 0. 
\end{align*}
}

Assume $\tilde{Q} = \{\tilde{\pi}, {f}_{0, L}, {f}_{0, M}, {\psi}_0 \}$. Then, 
{\small 
\begin{align*}
     \E[\Phi_1(\Theta_{\rho_2, \tilde{Q}})(O)] \ = 
   \  \E[\Phi_2(\Theta_{\rho_2, \tilde{Q}})(O)] \ =  
   \  \E[\Phi_3(\Theta_{\rho_2, \tilde{Q}})(O)] \ =  
   \  \E[\Phi_4(\Theta_{\rho_2, \tilde{Q}})(O)] \ =  
    \ \E[\Phi_5(\Theta_{\rho_2, \tilde{Q}})(O)] \ = 0. 
\end{align*}
}

%+++++++++++++++++++++++++++++++++++++++++++++++
\subsection{Theorem~\ref{thm:closed-cross-odds}} 
\label{app:proofs_closed-cross-odds}

Given the canonical gradient of the cross-entropy risk, canonical gradient of the constraint $\Theta_{\Pi, P_0}$, and Condition~\ref{eq:pathcondition}, we can write: 
\begin{align*}
    & \frac{\psi_{0, \lambda_0}(z) - \psi_0(z)}{\psi_{0, \lambda_0}(z) (1-\psi_{0, \lambda_0}(z))} + \lambda_0 D_{\Theta_\Pi, P_0}(z) = 0 \\ 
    \rightarrow \quad & \frac{\psi_{0, \lambda_0}(z) - \psi_0(z)}{\psi_{0, \lambda_0}(z) (1-\psi_{0, \lambda_0}(z))} + \lambda_0 \frac{D_{\Theta_\Delta, P_0}(z)}{\psi_{0, \lambda_0}(z) (1-\psi_{0, \lambda_0}(z))} = 0  \\
    \rightarrow \quad &\psi_{0, \lambda_0}(z) = \psi_0(z) - \lambda_0 D_{\Theta_\Delta, P_0}(z) \ . 
\end{align*}

Since $\lambda_0$ does not have a closed-form solution, we must rely on a grid search, where the grid is bounded by the fact that $\psi_{0, \lambda_0}$ must be in the unit interval. Assume $s_y=1$. For any datum $o = (s, w)$, we have  
\begin{align*}
   0 \leq \psi_0(z) - \lambda_0 D_{\Theta_\Delta, P_0}(z)  \leq 1 \ ,  
\end{align*}
where $D_{\Theta_\Delta, P_0}(z)$ is given in Lemma~\ref{lemma:gradients}. 

For $S=1$, we have: 
\begin{align*}
   0 \leq \psi_0(1,w) - \lambda_0 D_{\Theta_\Delta, P_0}(1,w)  \leq 1 \quad \rightarrow \quad  
   -\frac{1-\psi_0(1,w) }{D_{\Theta_\Delta, P_0}(1,w)} \leq \lambda_0 \leq \frac{\psi_0(1,w) }{D_{\Theta_\Delta, P_0}(1,w)}
\end{align*}

For $S=0$, we have: 
\begin{align*}
   0 \leq \psi_0(0,w) + \lambda_0 | D_{\Theta_\Delta, P_0}(0,w)|  \leq 1 \quad \rightarrow \quad  
   \frac{\psi_0(0,w) }{D_{\Theta_\Delta, P_0}(0,w)} \leq \lambda_0 \leq \frac{\psi_0(0,w) - 1}{D_{\Theta_\Delta, P_0}(0,w)}
\end{align*}

The bounds on $\lambda$ also agree with the second-order conditions for $\psi^*_0$ to be indeed a minimizer, see Appendix~C of \citep{nabi2024statistical}. 

Given the canonical gradients $D_{R, P_0}(\psi)$ and $D_{\Theta, P_0}(\psi)$, we can write the following:
\begin{align*}
    \mathcal{L}_{P_0}(\psi, \lambda) &=  R_{P_0}(\psi) + \lambda \Theta_{\Delta, P_0}(\psi) , \\ 
    \dot{\mathcal{L}}_{P_0}(\psi, \lambda) &=  \frac{\psi - \psi_0}{\psi(1-\psi)} + \lambda \frac{D_{\Theta_\Delta, P_0}}{\psi(1-\psi)}  \ , \\
    \ddot{\mathcal{L}}_{P_0}(\psi, \lambda) &= \frac{1}{\psi(1-\psi)} + \big( \psi - \psi_0 + \lambda D_{\Theta_\Delta, P_0} \big) \frac{-(1-2\psi)}{\psi^2(1-\psi)^2}   \ .
\end{align*} %
For any given $\psi = (\psi^1, \psi^0) \in { \Psi}$ and $\lambda \in \openr$, a sufficient condition for satisfying the second-order criterion as outlined in \citep{nabi2024statistical} is to have $ \ddot{\mathcal{L}}_{P_0}(\psi, \lambda) > 0 $.  That is, 
\begin{align*}
   \psi^2 - 2(\psi_0 - \lambda D_{\Theta_\Delta, P_0}) \psi + (\psi_0 - \lambda D_{\Theta_\Delta, P_0})  & > 0 \ .  
\end{align*}
The above condition holds if and only if $0 < \psi_0 - \lambda D_{\Theta_\Delta, P_0} < 1$.  
This concludes that $\psi_{0, \lambda_0} = (\psi^1_{0, \lambda_0}, \psi^0_{0, \lambda_0})$ is indeed the optimal  minimizer of the penalized risk.

%+++++++++++++++++++++++++++++++++++++++++++++++
\subsection{Theorem~\ref{thm:risk_at_P0}} 
\label{app:proof_risk_at_P0}

% Let $\theta_n$ be the selected estimate of $\theta_0 = \Theta_{\Delta, P_0}(\psi_0)$, the constraint. Let $D_n$ denote an estimate of $D_0 = D_{\Theta_{\Delta}, P_0}$, the gradient of the constraint. Let $\sigma_n^2$ be an estimate of $\sigma_0^2 = P_0 D_{0}^2$, the variance of the gradient. 

% Let $\mathcal{W}$ denote the domain of $W$. 
% We assume the following regularity conditions. 
% \begin{itemize}
%      \item There exists a constant $M$ such that, for all $n$ 
%     \begin{equation}
%         \sup_{s \in \{0, 1\}, w \in \mathcal{W}} D_0(s, w) < M \ .  \tag{R1} \label{eq:regularity_bounded_D0}
%     \end{equation}
    
%     \item There exists a positive constant $\delta$ such that for any $\epsilon > 0$ and sufficiently large $n$, 
%     % \begin{equation}
%     \begin{align}
%         P_0( \sigma^2_n > \delta_1 ) > 1 - \epsilon \ . 
%         \tag{R2}
%         \label{eq:regularity_c1}
%     \end{align}%
% \end{itemize}

\subsubsection*{Risk condition: $R_{P_0}(\psi^*_n) - R_{P_0}(\psi^*_0) = o_{P_0}(1)$}

We assume the following consistency statements:  
\begin{itemize}
    \item $L^2(P_0)$-consistency for the estimators $\psi_n$ and $D_n$,
    \begin{align}
        || \psi_n - \psi_0 ||_2 &= o_{P_0}(1) \ , \tag{C1}
        \\
        || D_n - D_0 ||_2 &= o_{P_0}(1) \ . \tag{C2}
        \label{eq:convergence1}
    \end{align}
    where $|| f_n - f_0 ||_2 = [P_0 \{ f_n - f_0 \}^2]^{1/2}$. 

    \item Consistency of the estimates $\theta_n$ and $\sigma^2_n$: \begin{align}
    \theta_n - \theta_0 &= o_{P_0}(1) \ , \tag{C1}
    \\
    \sigma_n^2 - \sigma_0^2 &= o_{P_0}(1) \ . \tag{C2}
    \label{eq:convergence2}
\end{align}
\end{itemize}

% \label{subsub:risk_derivation} 
We have that
\begin{align*}
    R_{P_0}(\psi^*_n) &= \int \left\{y - \psi^*_n(z)\right\}^2 dP_0(z, y) \\ 
%%%%%%%%%%%%%%%
    &= \int  \left\{y - \psi_0(z) + \psi_0(z) - \psi^*_n(z)\right\}^2 dP_0(z,y) \\
%%%%%%%%%%%%%%%
    &= \int \left[\left\{y - \psi_0(z)\right\}^2  + \left\{\psi_0(z) - \psi^*_n(z)\right\}^2 \right] dP_0(z,y) \\
%%%%%%%%%%%%%%%
    &= \int \left[\left\{y - \psi_0(z)\right\}^2  +  \left\{ \psi_0(z) - \psi_n(z) +  \frac{\theta_n}{\sigma^2_n} D_n(z)\right\}^2 \right] dP_0(z,y) \\
%%%%%%%%%%%%%%%
    &= \int \left[ \left\{y - \psi_0(z)\right\}^2 + \left\{\psi_0(z) - \psi_n(z)\right\}^2 + \left\{\frac{\theta_n}{\sigma^2_n} D_n(z) \right\}^2 + 2\frac{\theta_n}{\sigma^2_n} \left\{\psi_0(z) - \psi_n(z)\right\} D_n(z) \right] dP_0(z,y) \\
    %&\hspace{1cm} \ \pm \ \frac{\Theta^2_0(\psi_0)}{\sigma^4_0(D_0)} \int  D^2_0(z) dP_0(z) 
    %\\
    &= \int \left[ \left\{y - \psi_0(z)\right\}^2 + \left(\frac{\theta_0}{\sigma^2_0}\right)^2 D^2_0(z) \right] dP_0(z,y)  \\
    &\hspace{4em}
    + P_0(\psi_0 - \psi_n)^2 + 
    \left(\frac{\theta_n}{\sigma^2_n} \right)^2 P_0 D_n^2
    - \left(\frac{\theta_0}{\sigma^2_0}\right)^2 P_0 D^2_0
    + 2 \frac{\theta_n}{\sigma^2_n} P_0 \left[ \left\{\psi_0 - \psi_n\right\} D_n \right]
\end{align*}

Note that the first term in the final line equals $R_{P_0}(\psi^*_0)$ and thus

\begin{equation} \label{eq:risk_diff_asymptotic}
\begin{aligned}
    &R_{P_0}(\psi_n^*) - R_{P_0}(\psi^*_0) \\
    &\hspace{0.5cm} = 
    P_0 (\psi_n - \psi_0)^2 + \left\{ \left(\frac{\theta_n}{\sigma^2_n}\right)^2 - \left(\frac{\theta_0}{\sigma^2_0} \right)^2 \right\} P_0 D_n^2 + \left(\frac{\theta_0}{\sigma^2_0} \right)^2 P_0 (D_n^2 - D_0^2) + 2 \frac{\theta_n}{\sigma^2_n} P_0 \left\{ (\psi_0 - \psi_n) D_n \right\} \ .
\end{aligned}
\end{equation}
The first term on the right-hand side of (\ref{eq:risk_diff_asymptotic}) equals to $o_{P_0}(1)$ by the assumption of $L^2(P_0)$-consistency of $\psi_n$.

The second term on the right-hand side of (\ref{eq:risk_diff_asymptotic}) can be written as \begin{align*}
&\left\{ \left(\frac{\theta_n}{\sigma^2_n}\right)^2 - \left(\frac{\theta_0}{\sigma^2_0} \right)^2 \right\} P_0 D_n^2 \\
&\hspace{1cm} = \left\{ \left(\frac{\theta_n}{\sigma^2_n} - \frac{\theta_0}{\sigma^2_0} \right)^2 + 2 \frac{\theta_0}{\sigma^2_0} \left( \frac{\theta_n}{\sigma_n^2} - \frac{\theta_0}{\sigma^2_0} \right) \right\} P_0 \left[ (D_n - D_0)^2 + 2 D_0(D_n - D_0) + D^2_0 \right] \ ,
\end{align*}
indicating that this term consists only of second- and higher-order terms that will all equal to $o_{P_0}(1)$ under our assumptions.

The third term on the right-hand side of (\ref{eq:risk_diff_asymptotic}) is \begin{align*}
    \left(\frac{\theta_0}{\sigma^2_0}\right)^2 P_0 (D_n^2 - D_0^2) = \left(\frac{\theta_0}{\sigma^2_0}\right)^2 P_0 \left\{(D_n - D_0)^2 + 2 D_0(D_n - D_0) \right\} \ .
\end{align*}
$L^2(P_0)$ consistency of $D_n$, along with the fact that $\theta_0$ is finite and $\sigma_0^2$ is non-zero, implies that \[
\left(\frac{\theta_0}{\sigma^2_0}\right)^2 P_0 (D_n - D_0)^2 = o_{P_0}(1) \ .
\]
These same assumptions combined with (R1) and H{\"o}lder's inequality implies that
\[
\left(\frac{\theta_0}{\sigma^2_0}\right)^2 P_0 D_0 (D_n - D_0) = o_{P_0}(1) \ .
\]

The fourth term on the right-hand side of (\ref{eq:risk_diff_asymptotic})
is \begin{align*}
2 \frac{\theta_n}{\sigma^2_n} P_0 \left\{ (\psi_0 - \psi_n) D_n \right\} = 2 \frac{\theta_n}{\sigma^2_n} P_0 \left\{ (\psi_0 - \psi_n) (D_n - D_0) + D_0(\psi_n - \psi_0) \right\} \ . 
\end{align*}
The term $P_0 \left\{ (\psi_0 - \psi_n) (D_n - D_0)\right\}$ is second-order and will equal to $o_{P_0}(1)$ under the $L^2(P_0)$ consistency assumptions by the Cauchy-Schwarz inequality. On the other hand, boundedness of $D_0$, $L^2(P_0)$ consistency, and H{\"o}lder's inequality imply that $P_0 \{ D_0 (\psi_n - \psi_0) \} = o_{P_0}(1)$.

%+++++++++++++++++++++++++++++++++++++++++++++++
\subsection{Theorem~\ref{thm:constraint_at_P0}} 
\label{app:proof_constraint_at_P0}

\subsubsection*{Constraint condition: $\Theta_{P_0}(\psi^*_n) = o_{P_0}(1)$} 
% \label{subsub:constraint_derivation} 

Let $dP_{0, L, 1}(\tilde{\ell}) = \prod_{L_i \in \mathbb{L}_\rho \setminus Y} dP_0(\ell_i \mid \pa_\G(\ell_i)) \Big|_{S=1} $, $dP_{0, L, 0}(\tilde{\ell}) = \prod_{L_i \in \mathbb{L}_\rho \setminus Y} dP_0(\ell_i \mid \pa_\G(\ell_i)) \Big|_{S=0} $, and $dP_{0, M, 0}(\tilde{m}) = \prod_{M_i \in \mathbb{M}_\rho} dP_0(m_i \mid \pa_\G(m_i)) \Big|_{S=0} $. We use $\tilde{\ell}$ and $\tilde{m}$ as shorthand notations for the collective arguments that appear in $dP_{0, L, s}$ and $dP_{0, M, 0}$, respectively.  

The constraint $\Theta_{P_0}(\psi^*_n)$ can be analyzed as follows:
\begin{equation*}
\begin{aligned}
    \Theta_{P_0}(\psi^*_n) 
    &= \int \left\{ \psi^*_n(1, w) dP_{0, L, 1}(\tilde{\ell}) - \psi^*_n(0, w) dP_{0, L, 0}(\tilde{\ell}) \right\} \ dP_{0, M, 0}(\tilde{m}) \  dP_0(x) 
    \\
    &= \displaystyle \int \left\{ \psi_n(1, w) dP_{0, L, 1}(\tilde{\ell}) - \psi_n(0, w) dP_{0, L, 0}(\tilde{\ell}) \right\} \ dP_{0, M, 0}(\tilde{m}) \  dP_0(x)   \\ 
    &\hspace{0.25cm} - \frac{\theta_{n}}{\sigma^2_n} \int \left\{ D_{n}(1, w) dP_{0, L, 1}(\tilde{\ell}) - D_{n}(0, w) dP_{0, L, 0}(\tilde{\ell}) \right\} \ dP_{0, M, 0}(\tilde{m}) \  dP_0(x) 
    \\
    &= \Theta_{P_0}(\psi_n) - \frac{\theta_{n}}{\sigma^2_n}  \int  \frac{2s-1}{P_0(s \mid x)} \frac{P_{0, M, 0}(\tilde{m})}{P_{0, M, s}(\tilde{m})} D_{n}(s, w) dP_{0, L, s}(\tilde{\ell}) dP_{0, M, s}(\tilde{m}) dP_0(s, x)  \\
    &= \Theta_{P_0}(\psi_n) - \frac{\theta_{n}}{\sigma^2_n} P_0 ( D_0 D_n ) \\
    &= \Theta_{P_0}(\psi_n) - \frac{\theta_{n}}{\sigma^2_n} P_0 \{ D_0 ( D_n - D_0 ) \} - \frac{\sigma^2_0}{\sigma^2_n} \theta_{n} \ .
    % &= \Theta_{P_0}(\psi_n) - \frac{\Theta_{n}(\psi_n)}{\sigma^2_n(D_{n})}  \int \Big\{ \big(D_n - D_0\big) D_0 + D^2_0 \Big\}  dP_0(.) \\ 
    % &= \Theta_{P_0}(\psi_n) - \frac{\Theta_{n}(\psi_n)}{\sigma^2_n(D_{n})} \sigma^2_0(D_0) - \frac{\Theta_{n}(\psi_n)}{\sigma^2_n(D_{n})}  \int \big(D_n - D_0\big) D_0  dP_0(.) \ . 
\end{aligned}
\end{equation*}

    We note that $\Theta_{P_0}(f) = P_0 ( D_0 f )$ for any $P_0$-measurable function $f$. Thus, we can write:\begin{align*}
    &\Theta_{P_0}(\psi_n) - \frac{\theta_{n}}{\sigma^2_n} P_0 \{ D_0 ( D_n - D_0 ) \} - \theta_{n} \frac{\sigma^2_0}{\sigma^2_n} \\
    &\hspace{2em} = P_0 \{ D_0 (\psi_n - \psi_0) \} - \frac{\theta_{n}}{\sigma^2_n} P_0 \{ D_0 ( D_n - D_0 ) \} - \sigma^2_0 \left( \frac{\theta_{n}}{\sigma^2_n} - \frac{\theta_0}{\sigma^2_0} \right) \ . 
\end{align*}

In this case, sufficient conditions are: (i) bounded $D_0$; (ii) $L^1(P_0)$ convergence of $D_n$ and $\psi_n$; (iii) consistency of $\theta_n, \sigma^2_n$ to $\theta_0, \sigma^2_0$.

One could also analyze as follows. Let $\tilde{\psi}$ be the in-probability limit of $\psi_n$. 
% and let $\tilde{D}$ be the in-probability limit of $D_0$. 
Then 
\begin{align*}
&\Theta_{P_0}(\psi_n) - \frac{\theta_{n}}{\sigma^2_n} P_0 \{ D_0 ( D_n - D_0 ) \} - \frac{\sigma^2_0}{\sigma^2_n} \theta_{n} \\
&\hspace{2em} = P_0 \{ D_0 (\psi_n - \tilde{\psi}) \} - \frac{\theta_{n}}{\sigma^2_n} P_0 \{ D_0 ( D_n - D_0 )\} 
% - \frac{\theta_{n}}{\sigma^2_n} P_0 \{ D_0 ( \tilde{D} - D_0 )\} 
- \sigma^2_0 \left( \frac{\theta_{n}}{\sigma^2_n} - \frac{\Theta_{P_0}(\tilde{\psi})}{\sigma^2_0}\right) \ . 
\end{align*}
In this case, the sufficient conditions for convergence are those given by the Theorem: (i) bounded $D_0$, (ii) $L^1$-consistency of $\psi_n$ and $D_n$ for $\tilde{\psi}$ and ${D}_0$, respectively; 
% (ii)  $\tilde{D} = D_0$; 
% or at least that $P_0 (D_0 \tilde{D}) = P_0 (D_0^2)$; 
and (iii) $\theta_n, \sigma^2_n$ being consistent for $\Theta_{P_0}(\tilde{\psi}), \sigma^2_0$. 

% Essentially, sufficient conditions are to consistently estimate the gradient and use an estimate of the constraint that is consistent for $\Theta_{P_0}(\tilde{\psi})$ for any choice of $\tilde{\psi}$. 

%######################################
\section{Additional simulation details}
\label{app:sims}
%######################################

\subsection{Impact of the gradient's variance} 

\begin{figure}[!t]
    \centering
    \includegraphics[scale=0.24]{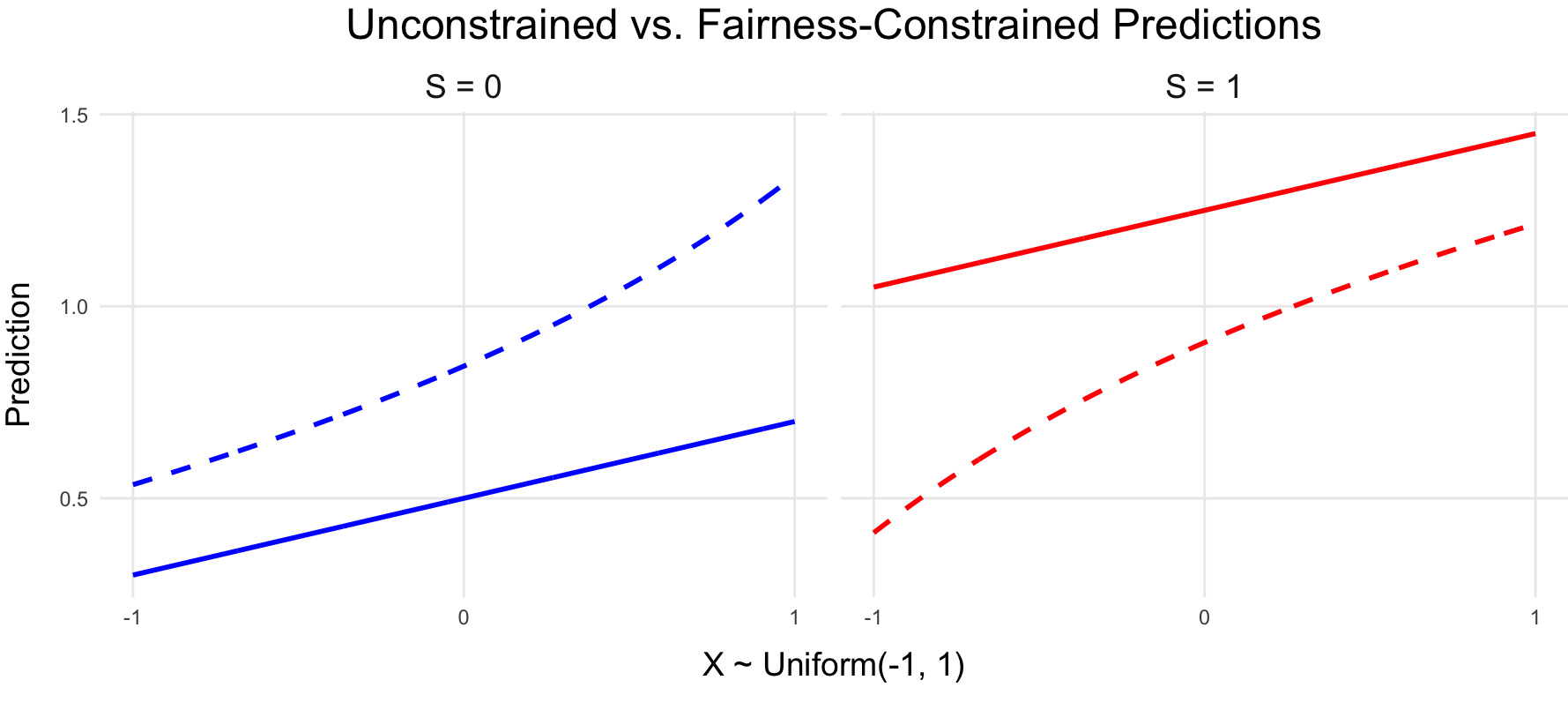} 

    \includegraphics[scale=0.24]{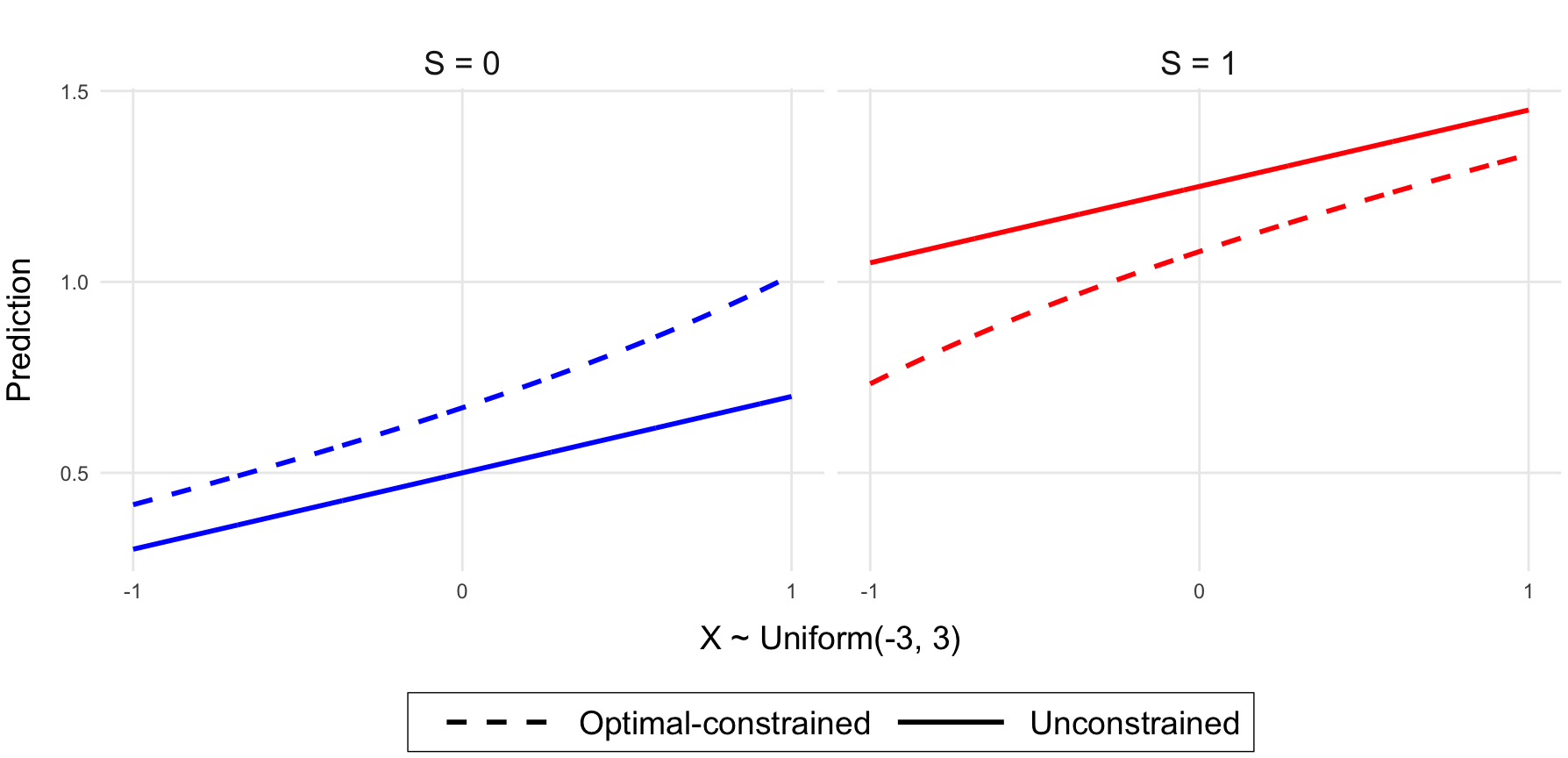}
    
    \caption{Comparisons of Unconstrained vs. Fairness-Constrained Predictions Across Different X Distributions. 
    \textit{(Top)} The covariate $X$ is uniformly distributed in the interval $(-1, 1)$ and $\sigma^2(D_{\Theta_\Delta, P_0})=4.36$. \textit{(Bottom)} The covariate $X$ is uniformly distributed in the interval $(-3, 3)$ and $\sigma^2(D_{\Theta_\Delta, P_0})=8.8$. In both panels, predictions are shown for $S=0$ (blue) and $S=1$ (red) with the unconstrained predictions (solid lines) and optimal fairness-constrained predictions (dashed lines). Adjustments are larger in the top panel due to the smaller variance of the gradient, necessitating more pronounced adjustments to meet fairness constraints. } 
    \label{fig:fair_adj_ate_2}
\end{figure}

To examine the influence of the variance of the gradient $\sigma^2(D_{\Theta_\Delta, P_0})$ on fairness adjustments, we revisit the ATE scenario from Section~\ref{sec:intuition}. In this example, the variance formula is expressed as:
\begin{align*}
\sigma^2(D_{\Theta_\Delta, P_0})
&= \int \left\{\frac{2s-1}{\pi_0(s \mid x)}\right\}^2 dP_0(s, x) = \int \frac{1}{\pi_0(1 \mid x) \pi_0(0 \mid x)} dP_0(x) \ .
\end{align*}
This variance depends on the distribution of the covariate $X$. For illustrative purposes, we contrast scenarios where $X$ follows a uniform distribution over the intervals $(-1, 1)$ and $(-3, 3)$. In both cases, $\pi_0(1 \mid x)$ and $\psi_0(s,x)$ are unchanged to maintain consistent constraints $\Theta_{\Delta, P_0}$ and gradients $D_{\Theta_\Delta, P_0}$ across our comparisons. Calculations reveal that the variance of the gradient is higher for $X$ in the range $(-3, 3)$, measuring at $8.8$, as opposed to $4.36$ for $(-1, 1)$. Consequently, more substantial fairness adjustments occur in the scenario with the narrower $X$ range due to the inverse relationship between gradient variance and adjustment magnitude. The results are provided in Figure~\ref{fig:fair_adj_ate_2}, where only the comparison within the $(-1, 1)$ interval is shown to ensure both examples have valid support for $X$.

\subsection{Impact of propensity score and mediator density ratio on adjustments}

\begin{figure}[!t]
    \centering
    \includegraphics[scale=0.25]{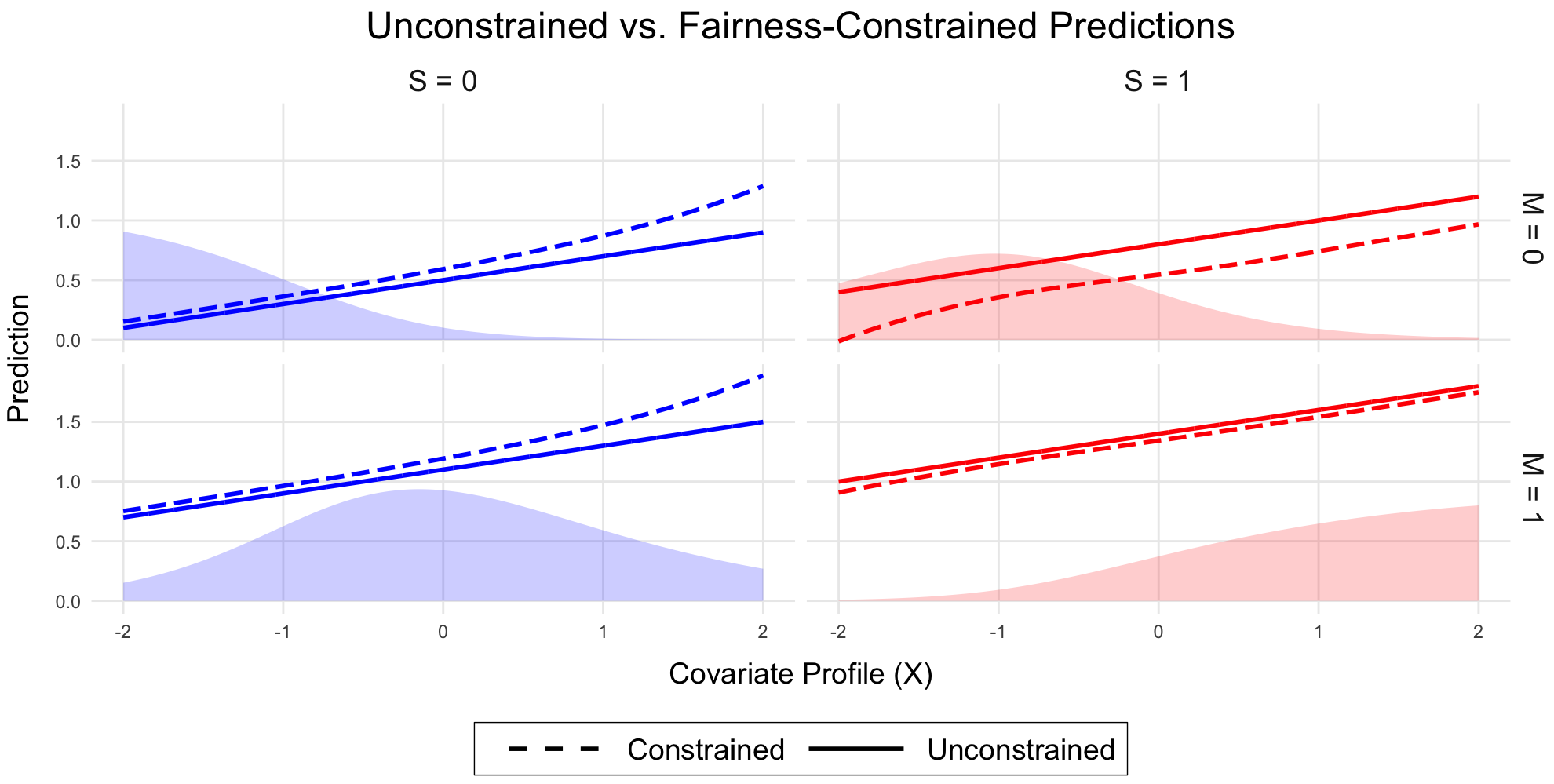}
    \caption{
    Illustrating the impact of covariate and mediator profiles on fairness adjustments in predictive modeling. Higher values of $X$ are associated with an increased likelihood of belonging to the $S=1$ group and a greater probability of $M=1$ for both $S=0$ and $S=1$ groups. The plots contrast unconstrained (solid lines) and fairness-constrained (dashed lines) predictions, distinguished by class status (color-coded) and mediator values (with predictions for $M=0$ at the top and $M=1$ at the bottom). Adjustments for the $(S=0)$ group are driven solely by the propensity score distribution, mirroring observations from Figure~\ref{fig:fair_adj_ate}. In contrast, adjustments for the $(S=1)$ group are shaped by a complex interplay between the propensity score and mediator density ratio, as detailed in Figures~\ref{fig:fair_adj_ate} and \ref{fig:fair_adj_de}. The shaded areas highlight the distribution of $X$ within each panel to demonstrate the effect of $X$'s variance on the extent of fairness adjustments required.} 
    \label{fig:fair_adj_de_2}
\end{figure}

To examine the simultaneous influence of the propensity score and mediator density ratio, we revisit the NDE example in Section~\ref{sec:intuition}. In this analysis, we modify the propensity score to depend explicitly on the covariate profile $X$, with $P_0(S=1 | X=x) = \text{expit}(x)$, as opposed to the initial model where $P_0(S=1 | X=x) = P_0(S=1)=0.5$. This modification allows us to investigate how changes in the dependency of $S$ on $X$---together with variations in the mediator density ratio---affect fairness adjustments. The results are illustrated in Figure~\ref{fig:fair_adj_de_2}. 

\subsection{Addition misspecification simulation results} \label{sec:addntl_sim_rslt_pse1}

\begin{figure}
\centering
\includegraphics[width=4cm, height=3.6cm]{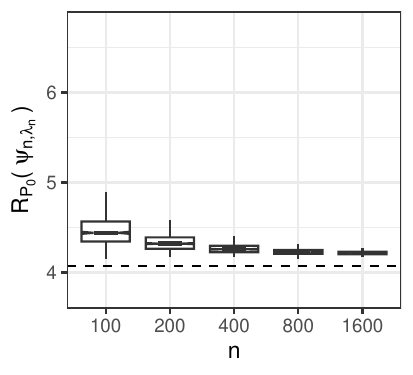}
\includegraphics[width=4cm, height=3.6cm]{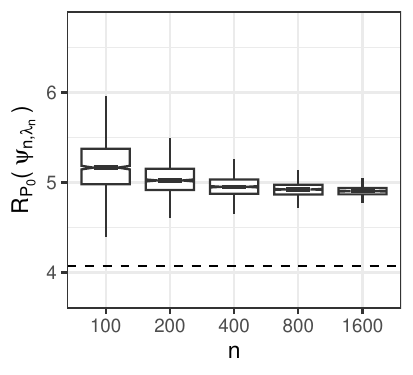}
\includegraphics[width=4cm, height=3.6cm]{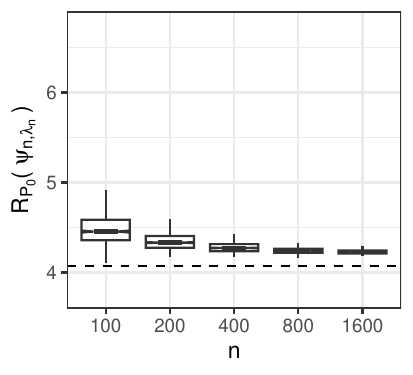}
\includegraphics[width=4cm, height=3.6cm]{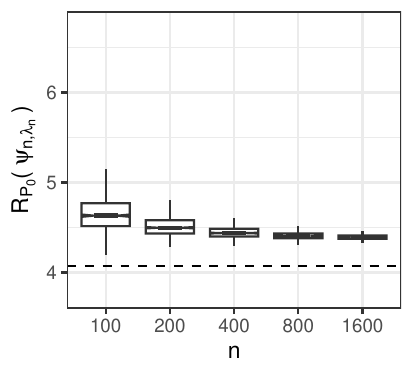}
\includegraphics[width=4cm, height=3.6cm]{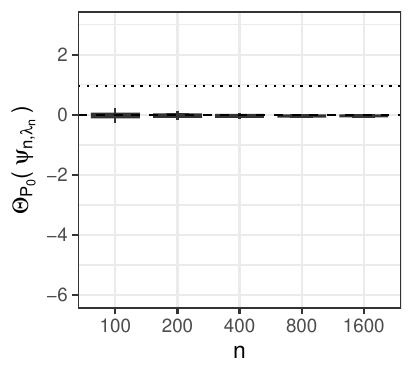}
\includegraphics[width=4cm, height=3.6cm]{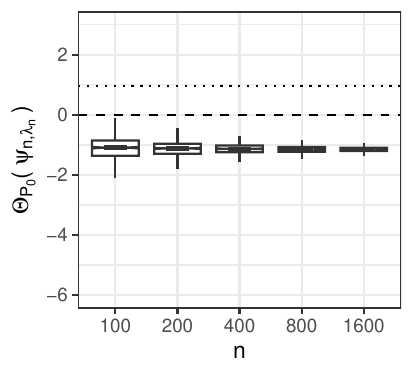}
\includegraphics[width=4cm, height=3.6cm]{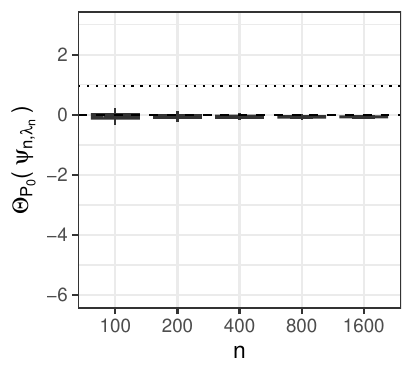}
\includegraphics[width=4cm, height=3.6cm]{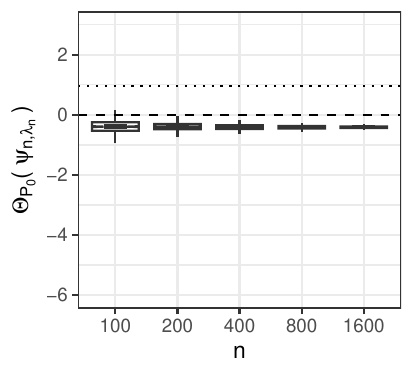}
\caption{\textbf{Estimates of optimal predictions under $\rho_1$-pathway constraint for mean squared error risk when $\pi_0$ is inconsistently estimated.} \underline{Top row:} Various estimators of the constraint are shown; from left to right: $\Theta^\text{plug-in}_{\rho_1, n}$, $\Theta^\text{ipw}_{\rho_1, n}$, $\Theta^\text{ipw-alt}_{\rho_1, n}$, $\Theta^\text{aipw}_{\rho_1, n}$. For each estimator, we show the distribution of risk of $\psi_{n,\lambda_n}$ over 1000 realizations for each sample size for the equality constraint $\Theta_{P_0}(\psi) = 0$. The dashed line indicates the optimal risk $R_{P_0}(\psi_0^*)$. \underline{Bottom row:} Distribution of the true constraint over 1000 realizations for each sample size. The dashed line indicates the equality constraint value of zero. The dotted line indicates the true value of the constraint under $\psi_0$.
}
\label{fig:pse1_sim_g}
\end{figure}

\begin{figure}
\centering
\includegraphics[width=4cm, height=3.6cm]{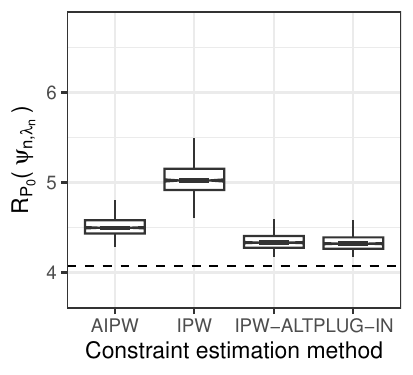}
\includegraphics[width=4cm, height=3.6cm]{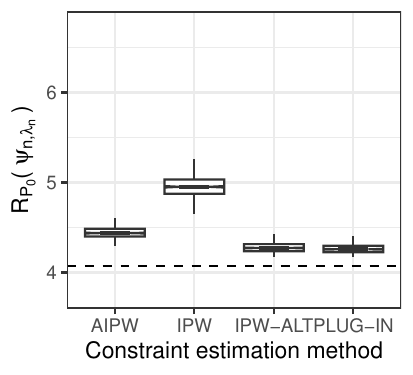}
\includegraphics[width=4cm, height=3.6cm]{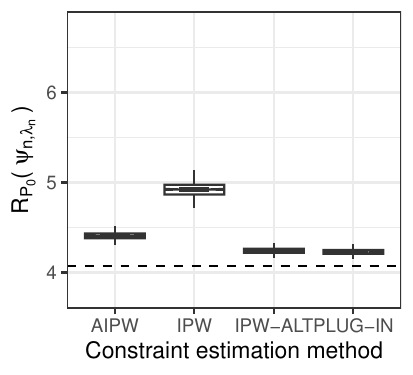}
\includegraphics[width=4cm, height=3.6cm]{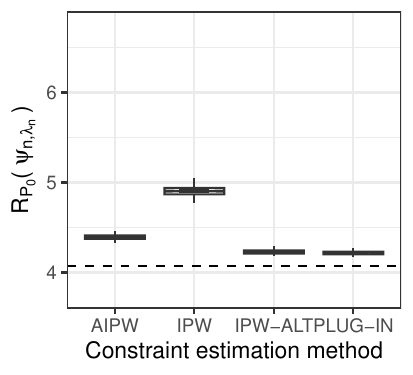}
\includegraphics[width=4cm, height=3.6cm]{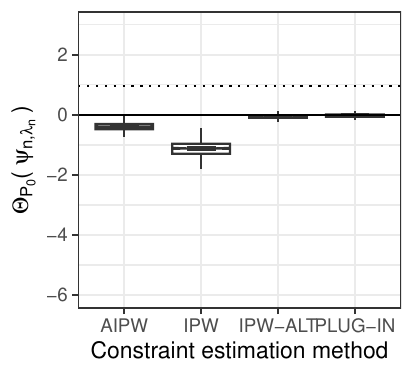}
\includegraphics[width=4cm, height=3.6cm]{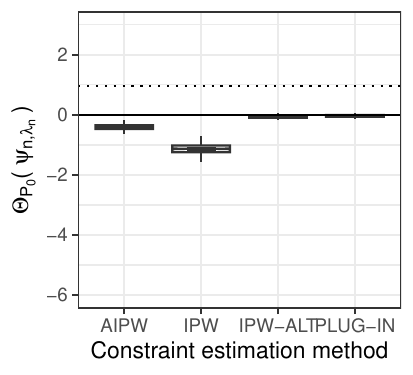}
\includegraphics[width=4cm, height=3.6cm]{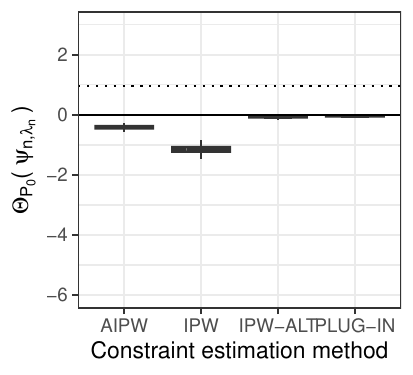}
\includegraphics[width=4cm, height=3.6cm]{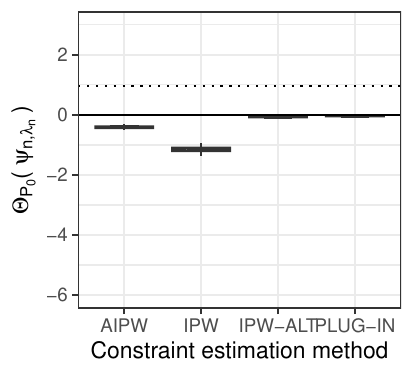}
\caption{\textbf{Comparison of pathway specific effect estimates used in construction of $\psi_n^*$ when $\pi_0$ is inconsistently estimated.} \underline{Top row:} The various estimators are shown in each figure for sample sizes (from left-to-right) of $n=200,400,800, 1600$. For each estimator, we show the distribution of mean squared error of $\psi_{n,\lambda_n}$ over 1000 realizations for each sample size for the equality constraint $\Theta_{P_0}(\psi) = 0$. The dashed line indicates the optimal risk $R_{P_0}(\psi_0^*)$. \underline{Bottom row:} Distribution of the true constraint over 1000 realizations for each sample size. The dashed line indicates the equality constraint value of zero. The dotted line indicates the true value of the constraint under $\psi_0$.
}
\label{fig:pse1_sim_by_theta_est_method_g}
\end{figure}

Figures \ref{fig:pse1_sim_g} and \ref{fig:pse1_sim_by_theta_est_method_g} show results for the misspecification simulation when $\pi_0$ is inconsistently estimated. Not surprisingly, the risk for estimators built using IPW were considerably worse than those built using the other estimators. Similarly, the constraint was poorly controlled when using the IPW estimator. Surprisingly, the constraint was reasonably well controlled using the plug-in estimator and the IPW estimator, in spite of the fact that the assumption on $L^2$-convergence of $D_n$ should be violated in this case. This illustrates that certain patterns of misspecification may still lead to reasonable control of the constraint even when assumptions of our theorems are violated.

\begin{figure}
\centering
\includegraphics[width=4cm, height=3.6cm]{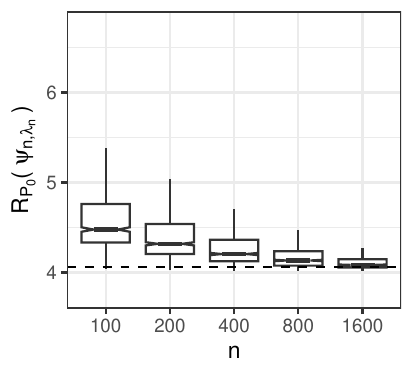}
\includegraphics[width=4cm, height=3.6cm]{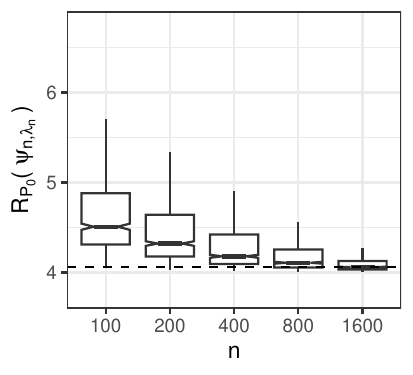}
\includegraphics[width=4cm, height=3.6cm]{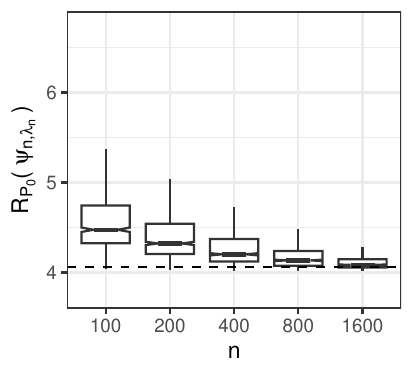}
\includegraphics[width=4cm, height=3.6cm]{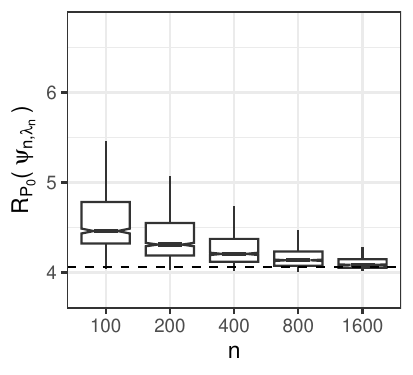}
\includegraphics[width=4cm, height=3.6cm]{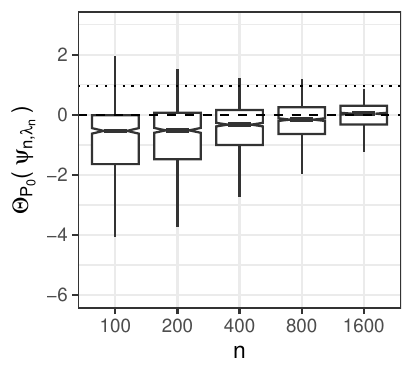}
\includegraphics[width=4cm, height=3.6cm]{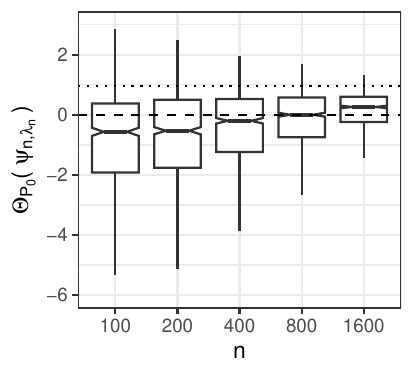}
\includegraphics[width=4cm, height=3.6cm]{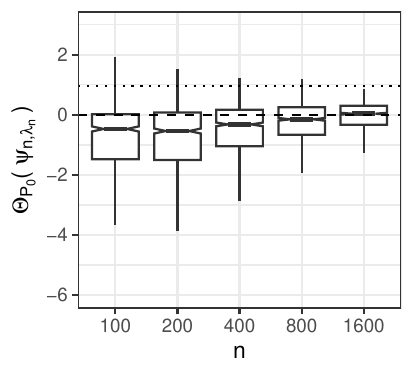}
\includegraphics[width=4cm, height=3.6cm]{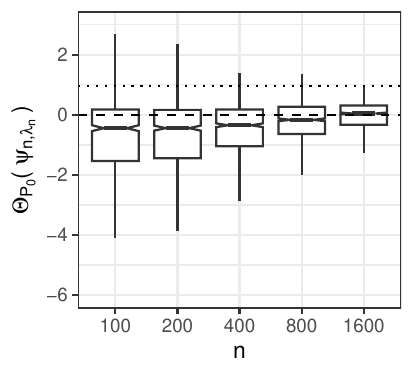}
\caption{\textbf{Estimates of optimal predictions under $\rho_1$-pathway constraint for mean squared error risk when $f_{0,M}$ is inconsistently estimated.} \underline{Top row:} Various estimators of the constraint are shown; from left to right: $\Theta^\text{plug-in}_{\rho_1, n}$, $\Theta^\text{ipw}_{\rho_1, n}$, $\Theta^\text{ipw-alt}_{\rho_1, n}$, $\Theta^\text{aipw}_{\rho_1, n}$. For each estimator, we show the distribution of risk of $\psi_{n,\lambda_n}$ over 1000 realizations for each sample size for the equality constraint $\Theta_{P_0}(\psi) = 0$. The dashed line indicates the optimal risk $R_{P_0}(\psi_0^*)$. \underline{Bottom row:} Distribution of the true constraint over 1000 realizations for each sample size. The dashed line indicates the equality constraint value of zero. The dotted line indicates the true value of the constraint under $\psi_0$.
}
\label{fig:pse1_sim_pMpL1}
\end{figure}

\begin{figure}
\centering
\includegraphics[width=4cm, height=3.6cm]{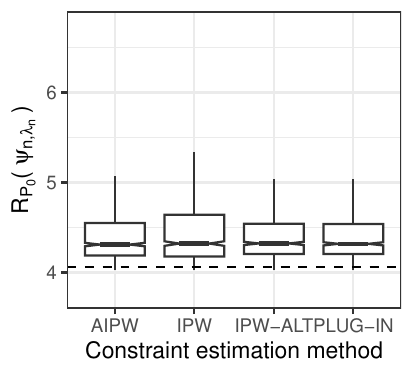}
\includegraphics[width=4cm, height=3.6cm]{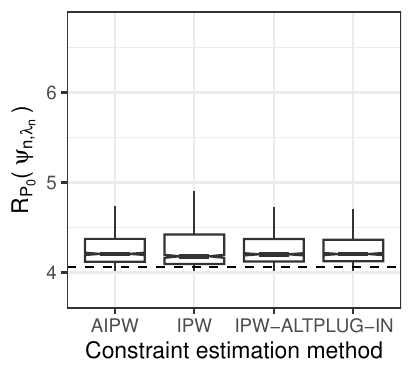}
\includegraphics[width=4cm, height=3.6cm]{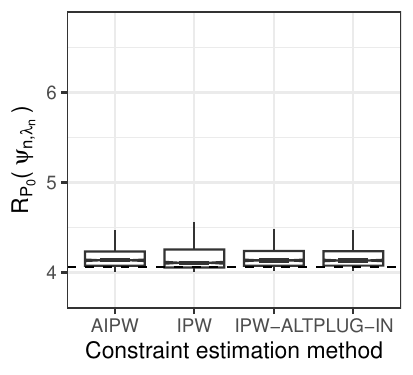}
\includegraphics[width=4cm, height=3.6cm]{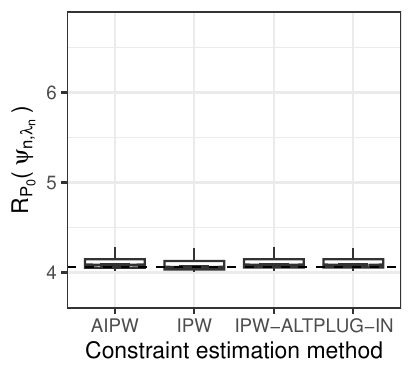}
\includegraphics[width=4cm, height=3.6cm]{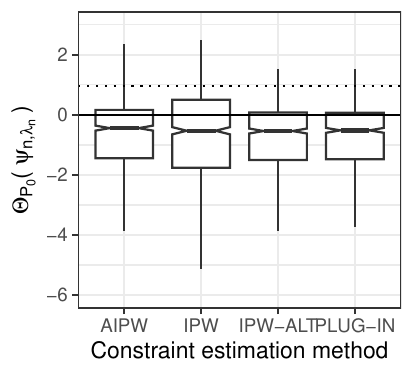}
\includegraphics[width=4cm, height=3.6cm]{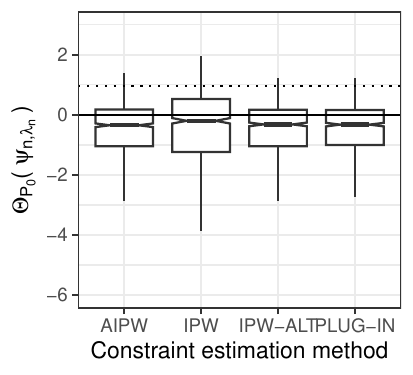}
\includegraphics[width=4cm, height=3.6cm]{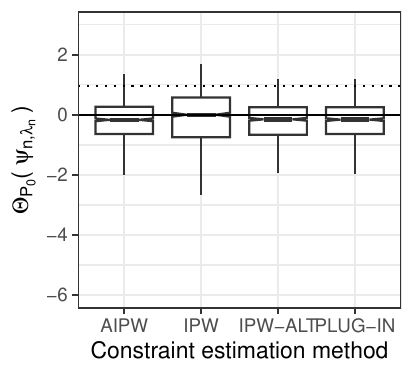}
\includegraphics[width=4cm, height=3.6cm]{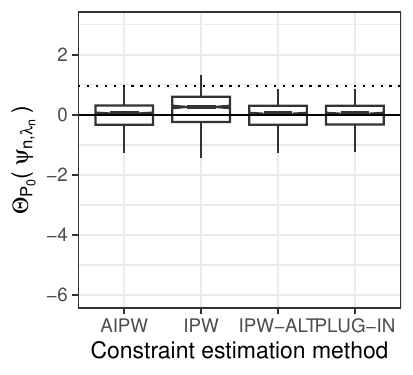}
\caption{\textbf{Comparison of pathway specific effect estimates used in construction of $\psi_n^*$ when $f_{0,M}$ is inconsistently estimated.} \underline{Top row:} The various estimators are shown in each figure for sample sizes (from left-to-right) of $n=200,400,800, 1600$. For each estimator, we show the distribution of mean squared error of $\psi_{n,\lambda_n}$ over 1000 realizations for each sample size for the equality constraint $\Theta_{P_0}(\psi) = 0$. The dashed line indicates the optimal risk $R_{P_0}(\psi_0^*)$. \underline{Bottom row:} Distribution of the true constraint over 1000 realizations for each sample size. The dashed line indicates the equality constraint value of zero. The dotted line indicates the true value of the constraint under $\psi_0$.
}
\label{fig:pse1_sim_by_theta_est_method_pMpL1}
\end{figure}

Figures \ref{fig:pse1_sim_pMpL1} and \ref{fig:pse1_sim_by_theta_est_method_pMpL1} illustrate results for when the conditional density for $M$ is inconsistently estimated. Somewhat surprisingly in this case, we saw reasonable performance of all of the estimators in terms of achieving near-optimal risk. Only the IPW estimator seemed to suffer in terms of control of the constraint. The good performance may be explained in part by the fact that the conditional density for $M$ appears in the gradient in a ratio form $f_{0,M}(M \mid S = 0, X) / f_{0,M}(M \mid S, X)$. Thus, even if $f_{0,M}$ is inconsistently estimated, this ratio will still be consistently estimated for $S = 0$. Thus, the $L^2$-norm of the gradient may still remain relatively small, even under misspecification of this nuisance parameter. We conjecture that more extreme patterns of misspecification and/or lower marginal probabilities of $S = 0$ in the data generating process may yield poorer performance.

\subsection{Additional simulation results for high-dimensional covariate simulation}

\begin{figure}
\centering
\includegraphics[width=4cm, height=3.6cm]{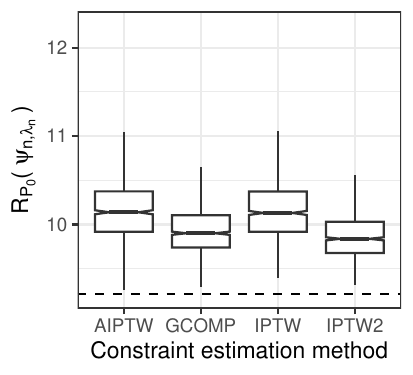}
\includegraphics[width=4cm, height=3.6cm]{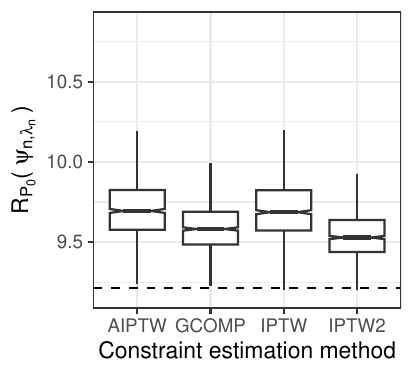}
\includegraphics[width=4cm, height=3.6cm]{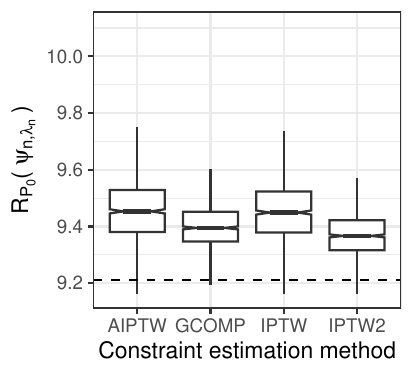}
\includegraphics[width=4cm, height=3.6cm]{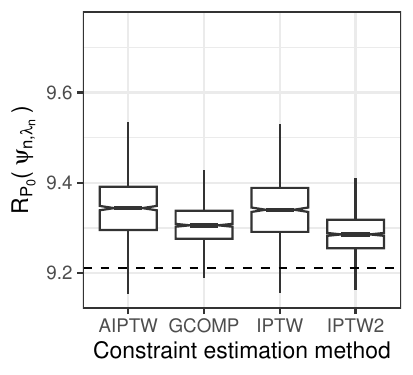}
\includegraphics[width=4cm, height=3.6cm]{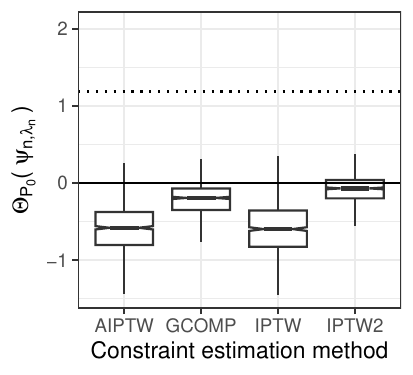}
\includegraphics[width=4cm, height=3.6cm]{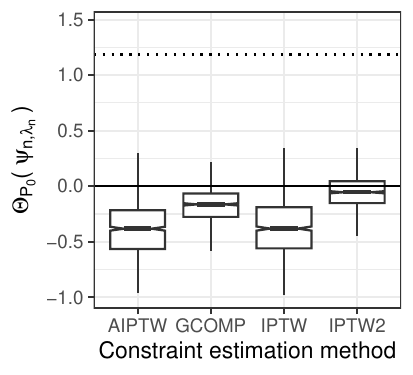}
\includegraphics[width=4cm, height=3.6cm]{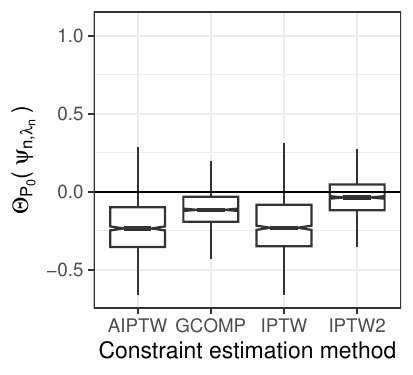}
\includegraphics[width=4cm, height=3.6cm]{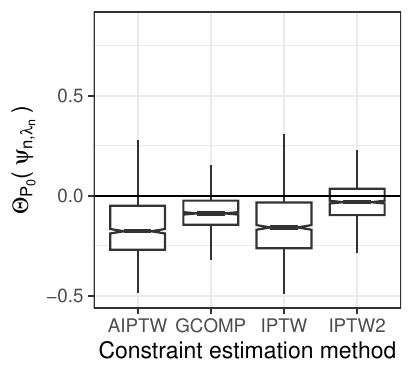}
\caption{\textbf{Comparison of pathway specific effect estimates used in construction of $\psi_n^*$ with high-dimensional covariates ($p = 10$).} \underline{Top row:} The various estimators are shown in each figure for sample sizes (from left-to-right) of $n=200,400,800, 1600$. For each estimator, we show the distribution of mean squared error of $\psi_{n,\lambda_n}$ over 1000 realizations for each sample size for the equality constraint $\Theta_{P_0}(\psi) = 0$. The dashed line indicates the optimal risk $R_{P_0}(\psi_0^*)$. \underline{Bottom row:} Distribution of the true constraint over 1000 realizations for each sample size. The dashed line indicates the equality constraint value of zero. The dotted line indicates the true value of the constraint under $\psi_0$.
}
\label{fig:pse1_sim_by_theta_est_method_glmnet_p10}
\end{figure}

\begin{figure}
\centering
\includegraphics[width=4cm, height=3.6cm]{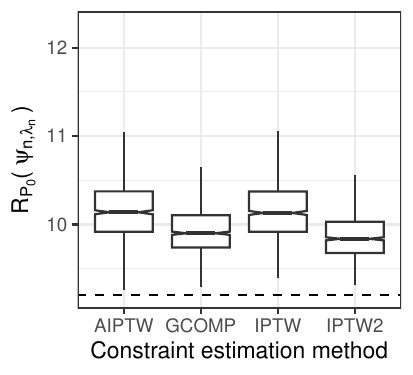}
\includegraphics[width=4cm, height=3.6cm]{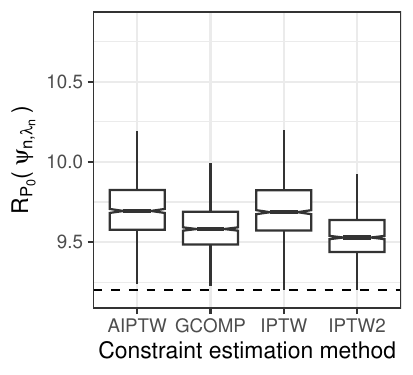}
\includegraphics[width=4cm, height=3.6cm]{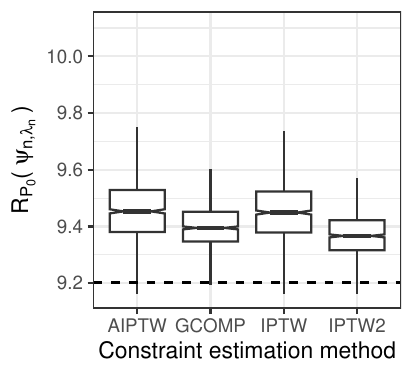}
\includegraphics[width=4cm, height=3.6cm]{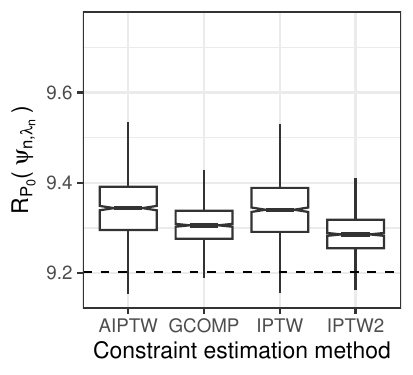}
\includegraphics[width=4cm, height=3.6cm]{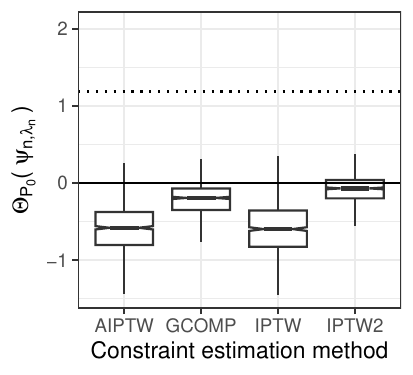}
\includegraphics[width=4cm, height=3.6cm]{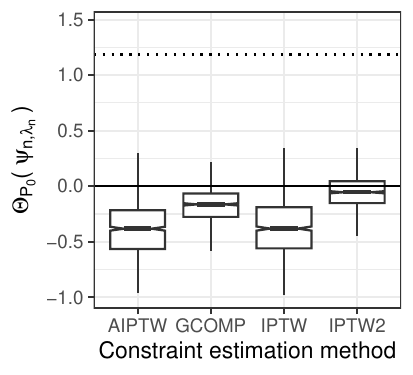}
\includegraphics[width=4cm, height=3.6cm]{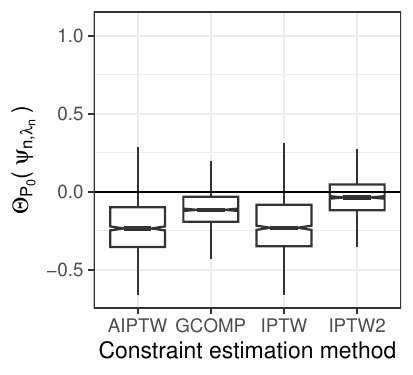}
\includegraphics[width=4cm, height=3.6cm]{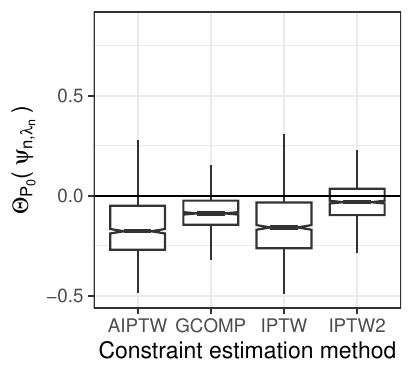}
\caption{\textbf{Comparison of pathway specific effect estimates used in construction of $\psi_n^*$ with high-dimensional covariates ($p = 50$).} \underline{Top row:} The various estimators are shown in each figure for sample sizes (from left-to-right) of $n=200,400,800, 1600$. For each estimator, we show the distribution of mean squared error of $\psi_{n,\lambda_n}$ over 1000 realizations for each sample size for the equality constraint $\Theta_{P_0}(\psi) = 0$. The dashed line indicates the optimal risk $R_{P_0}(\psi_0^*)$. \underline{Bottom row:} Distribution of the true constraint over 1000 realizations for each sample size. The dashed line indicates the equality constraint value of zero. The dotted line indicates the true value of the constraint under $\psi_0$.
}
\label{fig:pse1_sim_by_theta_est_method_glmnet_p50}
\end{figure}

The results of these settings were largely similar with the estimators constructed using the plug-in estimator and alternative IPW performing better in terms of both risk and constraint control when compared to IPW and AIPW (Figures \ref{fig:pse1_sim_by_theta_est_method_glmnet_p10} and \ref{fig:pse1_sim_by_theta_est_method_glmnet_p50}).

\end{appendix}

\end{document}